\documentclass{article}

\usepackage[nonatbib,preprint]{neurips_2023}

\usepackage[utf8]{inputenc} 
\usepackage[T1]{fontenc}    
\usepackage{hyperref}       
\usepackage{url}            
\usepackage{booktabs}       
\usepackage{amsfonts}       
\usepackage{nicefrac}       
\usepackage{microtype}      
\usepackage{xcolor}         

\usepackage{amsmath}
\usepackage{amssymb}
\usepackage{amsthm}
\usepackage{array}
\usepackage{epsfig}
\usepackage{graphicx}
\usepackage{xy}

\usepackage{times}
\usepackage{epsfig}
\usepackage{graphicx}
\usepackage{amsmath}
\usepackage{amssymb}
\usepackage{multirow}
\usepackage{subfigure}
\usepackage{comment}
\usepackage[export]{adjustbox}

\usepackage{algorithm}
\usepackage{algorithmic}
\usepackage{todonotes}

\newtheorem{theorem}{Theorem}
\newtheorem{definition}{Definition}
\newtheorem{lemma}[theorem]{Lemma}

\newtheorem*{theorem1}{Theorem~\ref{thm:metric_approx}}
\newtheorem*{theorem2}{Theorem~\ref{thm:hp_approx}}
\newtheorem*{lemma1}{Lemma~\ref{lemma:same_solution}}

\DeclareMathOperator*{\bbE}{\mathbb{E}}
\DeclareMathOperator*{\bbR}{\mathbb{R}}
\DeclareMathOperator*{\union}{\cup}

\DeclareMathOperator*{\argmin}{arg\,min}

\title{Robust Reinforcement Learning \\ through Efficient Adversarial Herding}

\author{%
  Juncheng Dong\thanks{equal contribution}\\
  Department of Electrical and Computer Engineering\\
  Duke University\\
  Durham, NC 27708 \\
  \texttt{juncheng.dong@duke.edu} \\
  \And
  Hao-Lun Hsu\footnotemark[1]\\
  Department of Computer Science\\
  Duke University\\
  Durham, NC 27708 \\
  \texttt{hao-lun.hsu@duke.edu} \\
  \AND
  Qitong Gao\\
  Department of Electrical and Computer Engineering\\
  Duke University\\
  Durham, NC 27708 \\
  \texttt{qitong.gao@duke.edu} \\
  \AND
  Vahid Tarokh\\
  Department of Electrical and Computer Engineering\\
  Duke University\\
  Durham, NC 27708 \\
  \texttt{vahid.tarokh@duke.edu} \\
  \AND
  Miroslav Pajic\thanks{corresponding author}\\
  Department of Electrical and Computer Engineering\\
  Duke University\\
  Durham, NC 27708 \\
  \texttt{miroslav.pajic@duke.edu} \\
}

\begin{document}

\maketitle

\begin{abstract}
Although reinforcement learning (RL) is considered the gold standard for policy design, it may not always provide a robust solution in various scenarios. This can result in severe performance degradation when the environment is exposed to potential disturbances. Adversarial training using a two-player max-min game has been proven effective in enhancing the robustness of RL agents. In this work, we extend the two-player game by introducing an adversarial herd, which involves a group of adversaries, in order to address (\textit{i}) the difficulty of the inner optimization problem, and (\textit{ii}) the potential over-pessimism caused by the selection of a candidate adversary set that may include unlikely scenarios. We first prove that adversarial herds can efficiently approximate the inner optimization problem. Then we address the second issue by replacing the worst-case performance in the inner optimization with the average performance over the worst-$k$ adversaries. We evaluate the proposed method on multiple MuJoCo environments. Experimental results demonstrate that our approach consistently generates more robust policies. 
\end{abstract}

\section{Introduction}
Deep reinforcement learning (RL) has shown its success toward synthesizing optimal strategies over environments with complex underlying dynamics, for example, in robotics~\cite{ibarz2021train, kalashnikov2018scalable}, healthcare~\cite{gao2022gradient, tang2021model} and games~\cite{silver2018general,vinyals2019grandmaster}. However, 
limited performance guarantees can be provided for the resulting policies
given the large parameter search space under the function approximation schema, and the limited scale of exploration over the state-action space during training due to sophisticated dynamics and environmental stochasticity~\cite{shen2020deep}. Consequently, there is often concern whether RL policies 
can perform consistently well upon deployment to the regions that are not well-explored during training, as well as against unforeseeable external disturbances applied to the agent,~\textit{i.e.}, the robustness of RL~\cite{RARL}. In this work, we introduce a robust RL framework that allows the resulting agents to generalize well toward environmental uncertainty and adversarial actions.

One framework that has been proven to effectively enhance the robustness of the RL agents is robustness through adversarial training~\cite{
gu2019adversary, kamalaruban2020robust, pattanaik2017robust, RARL, robust_population,zhang2021robust}. In this framework, the RL agent is assumed to share the environment with a hostile agent (adversary). The adversary takes actions to disturb the environment and/or the agent directly so that the cumulative reward received by the agent is minimized. Formulated as a max-min optimization problem, this framework optimizes the worst-case performance of RL agents under disturbance, and provides a lower bound for the performance of the agent, which is important for scenarios where safety and predictability are mostly~concerned.  

Despite these strengths, robustness through adversarial training has two major challenges: \textit{(i)} without a close-form solution, the inner minimization problem is difficult to solve and optimal solution is approximated by a first-order method such as gradient descent; \textit{(ii)} worst-case optimization can result in an over-conservative agent if the adversary is overly capable because the agent can be distracted by unlikely scenarios and attached to improve its performance in them. For instance, when training a controller for a helicopter, if the adversary is allowed to modify the environmental parameters to some physically unfeasible values, the agent must sacrifice its performance in real-world scenarios to improve its performance in these unlikely environments so that the worst-case performance~is~optimized. 

\begin{figure}
    \centering
    \includegraphics[width = 0.9\textwidth]{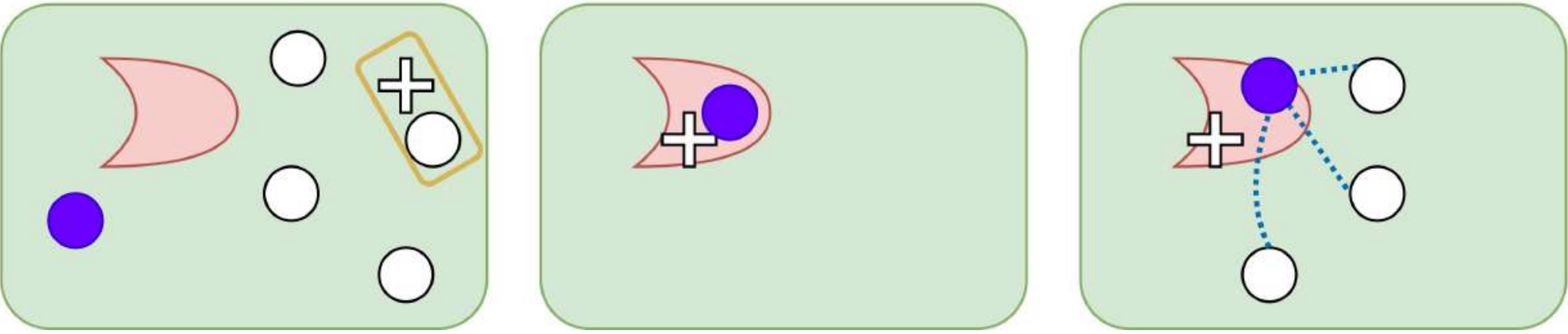}
    \caption{Benefits of introducing an adversarial herd. The green region is the space of all possible adversaries; the red  contains scenarios that are unlikely in real world and can result in over-conservative agents. The cross denotes optimal adversary; the circles denote adversaries; the circle with dark fill denotes the learning process of an agent with single adversary. (left) Compared with single adversary, an adversarial herd can efficiently approximate the optimal adversary, thus better approximating the inner optimization problem. (middle) The performance of the RL agent can be over-conservative if either the optimal adversarial is an unlikely scenario or the single adversary stays in "bad" region during the optimization. (right) An adversarial herd can help with over-pessimism by distributing the attention of the agent from the unlikely worst case to other cases.}
    \label{fig:intro_plot}
\end{figure}

To address these issues, we extend the two-layer max-min game by introducing an adversarial herd that involves a group of adversaries. Figure~\ref{fig:intro_plot} shows the advantages of an adversarial herd. In this work, we first theoretically establish that an adversary herd can relieve the first issue by proving that an adversarial herd can efficiently estimate the solution and the optimal value of the inner minimization problem. Specifically, we theoretically establish that \emph{the required number of adversaries to achieve high precision approximation is not demanding}. Next, we employ the proposed adversarial herd to mitigate the second issue by altering the objective of the RL agent from the worst-case performance to \emph{the average performance of the worst-$k$ adversaries}. 
The proposed method is evaluated on multiple MuJoCo environments with a variety of robust RL methods as baselines. It has been observed that our method consistently generate more robust policies than the baselines.  

\section{Preliminary}\label{sec:prelim}

\paragraph{Notations.} For any finite set $A$, we use $|A|$ to denote its cardinality. For any positive integer $m$, we use $[m]$ to represent the set of integers $\{1,\dots,m\}$. For any measurable set $\mathcal{M}$, we use $\Delta(\mathcal{M})$ to denote the set of all possible measures over $\mathcal{M}$. 

In this work, we consider a Markov Decision Process (MDP) with adversaries in the environment, defined by a tuple of 6 elements $(\mathcal{S}, \mathcal{A}^p, \mathcal{A}^a, \mathcal{P}, r, \gamma, p_0)$; 
here, $\mathcal{S}$ is the set of states of the environment, $\mathcal{A}^p$/$\mathcal{A}^a$ are the sets of actions that the agent (protagonist) or adversaries can take, $\mathcal{P}:\mathcal{S} \times \mathcal{A}^p \times \mathcal{A}^a \rightarrow \Delta(\mathcal{S})$ is the transition function that describes the distribution of the next state given the current state and actions taken by the agent and the adversaries, $r:\mathcal{S} \times \mathcal{A}^p \times \mathcal{A}^a \rightarrow \mathbb{R}$ is the reward function for the agent (we set the reward function for the adversary to $-r$ as we consider a zero-sum game framework in this work), $\gamma \in [0,1) $ is the discounting factor, and $p_0$ is the distribution of the initial state. 

We use $\pi_\theta: \mathcal{S} \rightarrow \Delta(\mathcal{A}^p)$ and $\pi_\phi: \mathcal{S} \rightarrow \Delta(\mathcal{A}^a)$ to respectively denote the polices of the agent and the adversaries, where $\theta$ and $\phi$ are their parameters. Specifically, we use $\pi_{\phi_i}$ and $\phi_i$ to denote the policy of the $i$-th adversary and its parameter. Let $s_t \in \mathcal{S}$ be the state of the environment at time $t$, $a^p_t \in \mathcal{A}^p$ (resp. $a^a_t \in \mathcal{A}^a$) the action of the agent (resp. adversary) at time $t$. We use  
\begin{equation}\label{eqn:def_R}
    R(\theta, \phi) \doteq \bbE_{s_0 \sim p_0}\big[\sum_{t=0}^\infty \gamma^t r(s_t, a^p_t, a^a_t ) | a^p_t \sim \pi_\theta(s_t), a^a_t \sim \pi_\phi(s_t)\big]
\end{equation}
to represent the cumulative discounted reward that the agent $\pi_\theta $ can receive under the disturbance of the adversary $\pi_\phi$. 
The objective of adversarial training (two-player max-min game) for robustness~\cite{RARL, robust_population} is defined as follows:
\begin{equation}\label{eqn:maxmin}
    \max_{\theta \in \Theta} \min_{\phi \in \Phi} R(\theta, \phi),
\end{equation}
where $\Theta$ and $\Phi$ are pre-defined parameter spaces for the agent and the adversaries. In this approach, the RL agent maximizes the worst-case performance under disturbance.

\section{Adversarial Herds}\label{sec:main}

Although adversarial training has achieved great empirical success, two major challenges persist. First, it is challenging to obtain a close approximation of the optimal solution $\phi^* \in \Phi$ to the inner minimization problem in~\eqref{eqn:maxmin}. The reason is that most commonly used 
techniques to solve~\eqref{eqn:maxmin} 
are based on iterative updating $\theta$ and $\phi$ with a first-order method, such as gradient descent, which is inclined to converge to a local-optimum as $R(\theta,\phi)$ is often highly non-convex over $\phi$. Second, if the candidate adversary set $\Phi$ is chosen without enough precision, it may lead to scenarios unlikely to be encountered in real-world applications. This will result in an over-conservative agent if it is distracted by these scenarios during policy optimization. 

To address these challenges, we propose a new approach that extends the two-player zero-sum game by employing an adversarial herd which involves a group of adversaries. In this section, our algorithm will be presented along with the theoretical results that illustrate the motivations and justify its effectiveness. Specifically, we first prove that an adversarial herd can help estimate the solutions to the inner optimization problem in Sec.~\ref{sec:efficient_est} efficiently,~\textit{i.e.}, the size of adversary herd is upper-bounded in order to obtain sufficient approximation precision. Next in Sec.~\ref{sec:worst-k}, we propose a new objective to replace the worst-case performance optimization in~\eqref{eqn:maxmin} to prevent the trained agents from being over-conservative. In Sec.~\ref{sec:algo}, we summarize and present detailed steps of the proposed algorithm.

\subsection{Efficient Approximation by Adversarial Herd}\label{sec:efficient_est}

The two-player zero-sum game approach has the objective $\max_{\theta \in \Theta} \min_{\phi \in \Phi} R(\theta, \phi)$, with $R(\theta, \phi)$ defined in~\eqref{eqn:def_R}, which is the expected cumulative reward of the agent $\pi_{\theta}$ under the attack from the adversary $\pi_{\phi}$. Due to the complexity of $R(\theta,\phi)$, the most popular approach to solve the inner optimization problem for a given $\theta$ is to use a single adversary and update the adversary with first-order optimization method such as gradient descent. However, this approach is likely to be stuck in the local optima, deviating from the global optimal solution and value of the inner problem. To address this issue, we first propose a variation of the above approach that employs multiple adversaries. 

Specifically, instead of a single adversary that updates itself, we employ a set of \emph{fixed} adversaries denoted by $\widehat \Phi \doteq \{\phi_i\}_{i=1}^m$, where $m$ is the total number of adversaries and for all $i\in[m], \; \phi_i \in \Phi$. Subsequently, we transform the original optimization problem in~\eqref{eqn:maxmin} into the following one
\begin{equation}\label{eqn:maxmin_new}
     \max_{\theta \in \Theta} \min_{\phi \in \widehat \Phi} R(\theta, \phi);
\end{equation}
in the new objective the agent $\pi_{\theta}$ still optimizes the worst-case performance but \emph{only over a finite set of adversaries}. A direct methodological advantage of this approach over the original one is that there is no need to use a first-order method. To find the optimal solution and value of the inner minimization problem in~\eqref{eqn:maxmin_new}, one only need to approximate $R(\theta,\phi_i)$ for all $\phi_i$ in $\widehat\Phi$, and to select the adversary $\phi$ that results in the minimum $R(\theta,\phi)$. This process takes linear time with respect to the number of~adversaries. 

Note in this approach, only the $1$-dimensional $R(\theta,\phi)$ needs to be approximated. However, in~the~original approach, to update the adversary, the gradient of $R(\theta,\phi)$ (with respect to $\phi$) must be estimated, which is a $d_{\phi}$-dimensional object where $d_{\phi}$ is the dimension of $\phi$ and often a large number. We next prove that the proposed approach can efficiently approximate the inner optimization problem in~\eqref{eqn:maxmin}.  

\begin{definition}[$L^{\infty}$ Norm and Norm Distance]
For a function $h: \mathcal{X} \rightarrow \mathbb{R}$, we define its $L^{\infty}$ norm as 
$$
||h||_{\infty} = \sup_{x \in \mathcal{X}}|h(x)|.
$$ 
\end{definition}
\begin{definition}[$\epsilon$-packing]
Let $(\mathcal{U}, d)$ be a metric space where $d: \mathcal{U}\times\mathcal{U} \rightarrow \mathbb{R}^+$ is the metric function. Then a finite set $\mathcal{X} \subset \mathcal{U}$ is an $\epsilon$-packing if no two distinct elements in $\mathcal{X}$ are $\epsilon$-close to each other, i.e.,
$$\inf_{x,x' \in \mathcal{X}: x\neq x'}d(x,x') > \epsilon.$$
\end{definition}
\begin{definition}[$\epsilon$-cover]
A set $\mathcal{X} \subset \mathcal{U}$ is an $\epsilon$-cover of $\mathcal{U}$ if for any $u \in \mathcal{U}$, there is an element in $\mathcal{X}$ that is $\epsilon$-close to $u$, or equivalently,~\textit{i.e.},
$$
\mathcal{U} \subseteq \union_{x \in \mathcal{X}}B_{\epsilon}(x) 
$$
where $B_{\epsilon}(x) \doteq\{u \in \mathcal{U}: d(u,x) \leq \epsilon \}$. 
\end{definition}
\begin{definition}[$\epsilon$-net]
A set $\mathcal{X} \subset \mathcal{U}$ is an $\epsilon$-net if $\mathcal{X}$ is both an $\epsilon$-packing and an $\epsilon$-cover. 
\end{definition}


Let $R_{\Phi}$ denote a function class defined as
$$R_{\Phi} \doteq \{ R_{\phi} \doteq R(\theta, \phi): \Theta \rightarrow \bbR | \phi \in \Phi \}.$$ 
Our first result illustrates that if one chooses a set of adversaries that are different enough, then the number of adversaries needed to approximate the inner optimization problem is in linear order of the desired precision. 

\begin{theorem}\label{thm:metric_approx}
Consider the metric space $(R_{\Phi}, ||\cdot||_{\infty})$ where for any two functions $R_{\phi}, R_{\phi'} \in R_{\Phi}$, the distance between them is defined as 
$$d(R_{\phi},R_{\phi'}) \doteq ||R_{\phi} - R_{\phi'}||_{\infty}.$$ 
Assume that $R_{\Phi}$ has finite radius under this metric, i.e., 
\begin{equation}\label{eqn:bouned_assumption}
    \sup_{\phi, \phi' \in \Phi}d(R_{\phi},R_{\phi'}) \leq r_{\max},
\end{equation}
where $r_{\max} < \infty$ is a finite number. Let $\widehat\Phi = \{\phi_i\}_{i=1}^m \subset \Phi$. If $R_{\widehat\Phi}$ is a \textit{maximal} $\epsilon$-packing then $|R_{\widehat \Phi}| \geq \lceil \frac{r_{\max}}{\epsilon} \rceil$, where $\lceil c \rceil$ is the smallest integer that is larger than or equal to $c$,  and $R_{\widehat \Phi}$ is also an $\epsilon$-net. Moreover, for any $\theta \in \Theta$, let $\hat \phi \doteq \argmin_{\phi \in \widehat\Phi}R(\theta,\phi)$ denote the approximated solution and $\phi^* \doteq \argmin_{\phi \in \Phi}R(\theta,\phi)$ denote the optimal solution. Then, the  approximation error of $\hat\phi$ on the inner optimization problem is upper bounded by $\epsilon$, i.e., 
$$
|R(\theta,\phi^*) - R(\theta,\hat\phi)| \leq \epsilon.
$$
\end{theorem}

The proof of the theorem is provided in Appendix~\ref{app:proofs}.
The assumption in~\eqref{eqn:bouned_assumption} is essentially requesting that for any policy $\pi_{\theta}$, its performance in two different environments cannot vary infinitely. From another perspective, this is equivalent to suggesting that the adversary cannot be omnipotent. Under this assumption, if we can construct a set of adversaries that are distinct from each other, then the number of adversaries one needs for approximation is about $O(\frac{1}{\epsilon})$, where $\epsilon$ can be interpreted as the desired level of accuracy towards the approximation. 
We next show that if one only want to use adversarial herd to approximate accurately with high probability, instead of an almost sure approximation as in Theorem~\ref{thm:metric_approx}, then the number of required adversaries can be reduced. 

\begin{theorem}\label{thm:hp_approx}
Assume that $\Phi$ is a metric space with a distance function $d: \Phi \times \Phi \mapsto \mathbb{R}$. Let $\sigma$ be any probability measure on $\Phi$. Let $\widehat\Phi = \{\phi_i\}_{i=1}^m$ be a set of independently sampled elements from $\Phi$ following identical measure $\sigma$. Consider a fixed $\theta \in \Theta$ and assume that $R(\theta,\phi)$ is an $L_{\phi}$-Lipschitz continuous function of $\phi$ with respect to the metric space $(\Phi, d)$. Let $\widehat\phi$ and $\phi^*$ be defined the same as in Theorem~\ref{thm:metric_approx}. For presentation simplicity, assume that $\sigma(\{\phi: d(\phi,\phi^*)\leq\epsilon\}) \geq L_{\sigma}\epsilon$. Let $0 < \delta < 1$ denote the probability of a bad event. Then with probability $1-\delta$, the  approximation error of $\hat\phi$ on the inner optimization problem is upper bounded by $\epsilon$ if 
$m \geq \log(\delta)\log^{-1}(1-\frac{L_{\sigma}}{L_{\phi}}\epsilon)$.
\end{theorem}

The proof is provided in Appendix~\ref{app:proofs}. In Theorem~\ref{thm:hp_approx}, one can replace $L_{\sigma}$ with other dense conditions about measure of $\Phi$ and reach similar results. Compared with Theorem~\ref{thm:metric_approx},  if one can sample from a measure that is dense around the optimal $\phi$, then the required number of adversaries can be decreased. 

While the above results shed some lights on how we should design the adversarial herd algorithm, one may still encounter a couple of challenges in practice. In Theorem~\ref{thm:metric_approx},we would like to construct an $\epsilon$-packing. However, as even verifying for two adversaries $\phi, \phi'$ that $d(R_{\phi},R_{\phi'}) = ||R_{\phi}-R_{\phi'}||_{\infty} \geq \epsilon$ is challenging, it makes construction of an $\epsilon$-packing to be intractable. In Theorem~\ref{thm:hp_approx}, it is often challenging to estimate $L_{\phi}$ as well as to construct a measure $\sigma$ that is dense near $\phi^*$. 

To address these problems,  we let $\phi_i \in \widehat \Phi$ be learners, instead of fixed adversaries. The objective then becomes 
\begin{equation}\label{eqn:rah}
    \max_{\theta \in \Theta} \min_{\phi_1,\dots,\phi_m \in \Phi} \min_{\phi \in \{\phi_i\}_{i=1}^m} R(\theta, \phi).
\end{equation}
It is important and interesting to observe that the solution set of~\eqref{eqn:rah} is identical to that of the maximin problem in the original approach.  

\begin{lemma}\label{lemma:same_solution}
The solution set to the optimization problem in~\eqref{eqn:maxmin} is identical to the solution set of the optimization problem in~\eqref{eqn:rah}.
That is, for any $\theta \in \Theta$ and integer $m \geq 1$,
$$
\min_{\phi \in \Phi} R(\theta,\phi) = \min_{\phi_1,\dots,\phi_m \in \Phi}\min_{\phi \in \{\phi_i\}_{i=1}^m} R(\theta,\phi). 
$$
\end{lemma}

The proof is presented in Appendix~\ref{app:proofs}. 
The above result implies that the true benefit brought by the proposed algorithm lies in the optimization process instead of the final optimal solution it offers. In other words, adversarial training with adversarial herd still optimizes the worst-case performance of an agent for a pre-defined candidate adversary set $\Phi$, but adversarial herd can help find the solution to this optimization problem by efficiently estimating the inner optimization problem.

\subsection{Resolving Potential Over-Pessimism}\label{sec:worst-k}
The max-min game in~\eqref{eqn:maxmin} can lead to a solution that is too conservative due to the worst case optimization. Specifically, if the range of the adversaries $\Phi$ is not chosen correctly, this can cause an agent to be distracted by scenarios that are unlikely to happen in the real world. To this end, we modify the objective of the agent $\pi_{\theta}$, from optimizing its worst-case performance, to optimizing its average performance over the worst-$k$ adversaries. 

We define the worst-$k$ adversaries in a set of adversaries $\{\phi_i\}_{i=1}^m$ for a fixed agent $\pi_\theta$ as follows. A group of $k$ adversaries is the worst-$k$ adversaries if the expected cumulative rewards received by the agent $\pi_\theta$ under their attack are smaller than that under the attack from the rest $m-k$ adversaries. Specifically, for a given set of adversaries $\widehat \Phi \doteq \{\phi_i\}_{i=1}^m$ and $\theta$, let $W_{\theta}(\phi)\doteq \{\phi' \in \widehat \Phi : R(\theta,\phi') \leq R(\theta,\phi) \}$. For an integer $k \geq 1$, let $I_{\theta,\widehat \Phi,k} \doteq \{i \in [m]:\phi_i \in \widehat \Phi, |W_{\theta}(\phi_i)| \leq k \}$ denote the set of indices of the worst-$k$ adversaries for a given policy $\pi_{\theta}$. The new objective is then defined as:
\begin{equation}
\label{eqn:our_obj}
     \max_{\theta \in \Theta} \min_{\phi_1,\dots,\phi_m \in \Phi} \frac{1}{|I_{\theta,\widehat \Phi,k}|}\sum_{i\in I_{\theta,\widehat \Phi,k}}R(\theta, \phi_i).
\end{equation}

Average over worst-$k$ performances can balance out the pessimism, preventing the agent from attaching to the scenarios that can potentially lead to over-conservative policies. We note that the algorithm proposed by~\cite{vinitsky2020robust} is a special case of our algorithm where $k$ is set to $m$. We believe that its empirical success is also contributed by its effective relieving of over-pessimism. However, their approach does not give the worst-case performance enough attention as they always average over all the adversaries. This can result in degraded robustness. This is corroborated by the superior performance of our algorithm to the one from~\cite{vinitsky2020robust}, as demonstrated in Section~\ref{sec:exp}.  

\subsection{Robust Reinforcement Learning with Adversarial Herd}\label{sec:algo}

We now introduce our algorithm, RObust reinforcement Learning with Adversarial Herds (ROLAH), with pesudo-code presented in Algorithm~\ref{alg:main}. ROLAH is an iterative algorithm that sequentially update the policy $\pi_{\theta}$ and the adversarial herd $\{\phi_i\}_{i=1}^m$ to solve 
$$
 \max_{\theta \in \Theta} \min_{\phi_1,\dots,\phi_m \in \Phi} \frac{1}{|I_{\theta,\widehat \Phi,k}|}\sum_{i\in I_{\theta,\widehat \Phi,k}}R(\theta, \phi_i),
$$
where $R(\theta, \phi) = \bbE\big[\sum_{t=0}^{\infty}\gamma^t r_t | \pi_{\theta}, \pi_{\phi}\big]$ is the expected (discounted) cumulative rewards that the agent $\pi_{\theta}$ can receive under the disturbance of the adversary $\pi_{\phi}$. For ease of presentation, we assume that all the rollout trajectories have length $H$. We will use superscript to denote the index of iteration number. For instance, $\phi^t_i$ denotes the parameter of the $i$-th adversary in the $t$-th iteration of the algorithm. 

ROLAH first randomly initialize the agent policy and the adversarial herds. In each iteration, we first update the adversary herds and then update the agent policy with the updated adversaries. Specifically, in the $t$-th iteration, for $i \in [m]$, we collect a batch of trajectories $\rho^t_i =\{\tau^{t,j}_i\}_{j=1}^{b_a}$ where $b_a$ is the batch size for training the adversarial herds. The trajectories are collected by rolling out the agent $\pi_\theta$ and the $i$-adversary in the environment. Each trajectory in $\rho^t_i$ consists of $H$ transition tuples $\{(s_0,a_0,-r_0, s_1)\times\dots\times(s_H,a_H,-r_H,s_{H+1})\}$, where for $0 \leq h \leq H$, $a_h$ is the action by the $i$-th adversary and $r_h$ is the reward received by the agent. After collecting the trajectories for all the adversaries, we use these trajectories to estimate $R(\theta,\phi_i)$ for all $i \in [m]$, and select the worst-$k$ adversaries. Then we update these $k$ selected adversaries with the corresponding trajectories. The rest $m-k$ adversaries remain unchanged. Note that any RL algorithms can be used in the update. 

After the adversarial herd has been updated, we update the agent policy $\pi_{\theta}$. To identify the worst-$k$ adversaries, i.e., the elements in $I_{\theta,\widehat \Phi,k}$, we first estimate $R(\theta,\phi_i)$ for $i \in [m]$ by rolling out the agent $\pi_{\theta}$ with the $i$-th adversary in the environment to have an estimation $\widehat R_i$. Then we set $I_{\theta,\widehat \Phi,k}$ to contain all the indices $i$ such that $\widehat R_i$ is no greater than the $k$-th smallest element of the set $\{\widehat  R_j\}_{j=1}^m$. For each adversary $i$ in $I_{\theta,\widehat \Phi,k}$, we roll out the agent $\pi_{\theta}$ with $\pi_{\phi_i}$ to collect $b_p$ trajectories, each trajectory consisting of $\{(s_0,a_0,r_0, s_1)\times\dots\times(s_H,a_H,r_H,s_{H+1})\}$, where for $0 \leq h \leq H$, $a_h$ is the action by the agent $\pi_\theta$ and $r_h$ is the reward received by the agent. Then we pull all the collected trajectories together as the training dataset $\rho_p^t$ with $k\cdot b_p$ trajectories in total. Finally we use TRPO\footnote{This can be generalized to any RL policy optimization method.} to update $\theta$, i.e., the parameter of the agent, with $\rho_p^t$. The proposed algorithm is executed until the parameter of the agent policy $\theta$ converges or for a maximum of $T$ iteration, whichever happens first. 
 
\renewcommand{\algorithmicrequire}{\textbf{Input:}}
\renewcommand{\algorithmicensure}{\textbf{Output:}}
\begin{algorithm}[h]
\caption{RObust reinforcement Learning with Adversarial Herds (ROLAH)}
\begin{algorithmic} 
\REQUIRE $m$: size of the adversarial herd ; $k$: the number of the worst adversaries to use; $\lambda_p$: step size for updating the agent policy; $\lambda_a$: step size for updating the adversary herd; 
\ENSURE $\widehat \theta$: parameter for the agent policy. 
\STATE Randomly initialize $\theta$ and $\{\phi_i\}_{i=1}^m$
\STATE $t \leftarrow 0$, $\theta^t \leftarrow \theta$, $\phi^t_{i} \leftarrow \phi_i \;\; \forall i \in [m]$
\FOR{$t=0:T-1$}
\STATE \COMMENT{Update the adversarial herd.}
    \FOR{$i=1:m$}
        \STATE Estimate $R(\theta^t,\phi^t_i)$ by rolling out the agent $\pi_{\theta^t}$ with the adversary $\pi_{\phi^t_i}$
    \ENDFOR
\STATE Construct $I_{\theta,\widehat \Phi,k}$ with the estimations. 
\STATE $\phi^{t+1}_j \leftarrow \phi^t_j - \lambda_a \nabla_{\phi} R(\theta^t, \phi^t_j) \quad \forall j \in I_{\theta,\widehat \Phi,k} $
\STATE \COMMENT{Update the agent policy.}
\FOR{$i=1:m$}
        \STATE Estimate $R(\theta^t,\phi^{t+1}_i)$ by rolling out the agent $\pi_{\theta^t}$ with the adversary $\pi_{\phi^{t+1}_i}$
    \ENDFOR
\STATE Construct $I_{\theta,\widehat \Phi,k}$ with the estimations. 
\STATE $\theta^{t+1} \leftarrow \theta^t - \lambda_p \sum_{j \in I_{\theta,\widehat \Phi,k}}\nabla_{\theta} R(\theta^t, \phi^{t+1}_j)$

\ENDFOR

\STATE $\widehat \theta \leftarrow \theta^T$
\end{algorithmic}
\label{alg:main}
\end{algorithm}

    

\begin{figure}
     \centering
     \subfigure[Hopper]{
         \includegraphics[width=0.2\linewidth]{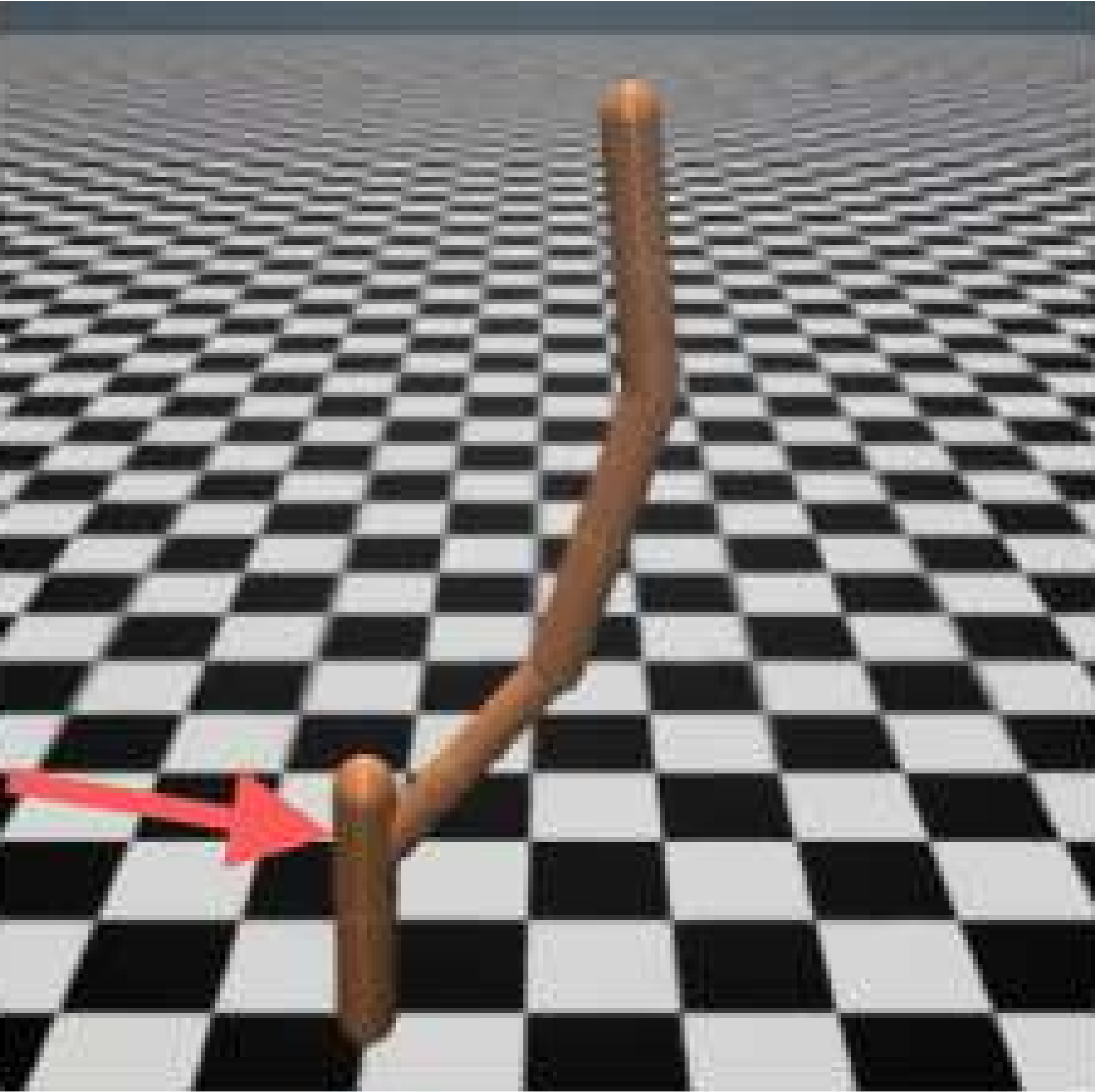}
         \label{fig:hopper_visual}
     }
     \subfigure[Walker2d]{       
         \includegraphics[width=0.2\linewidth]{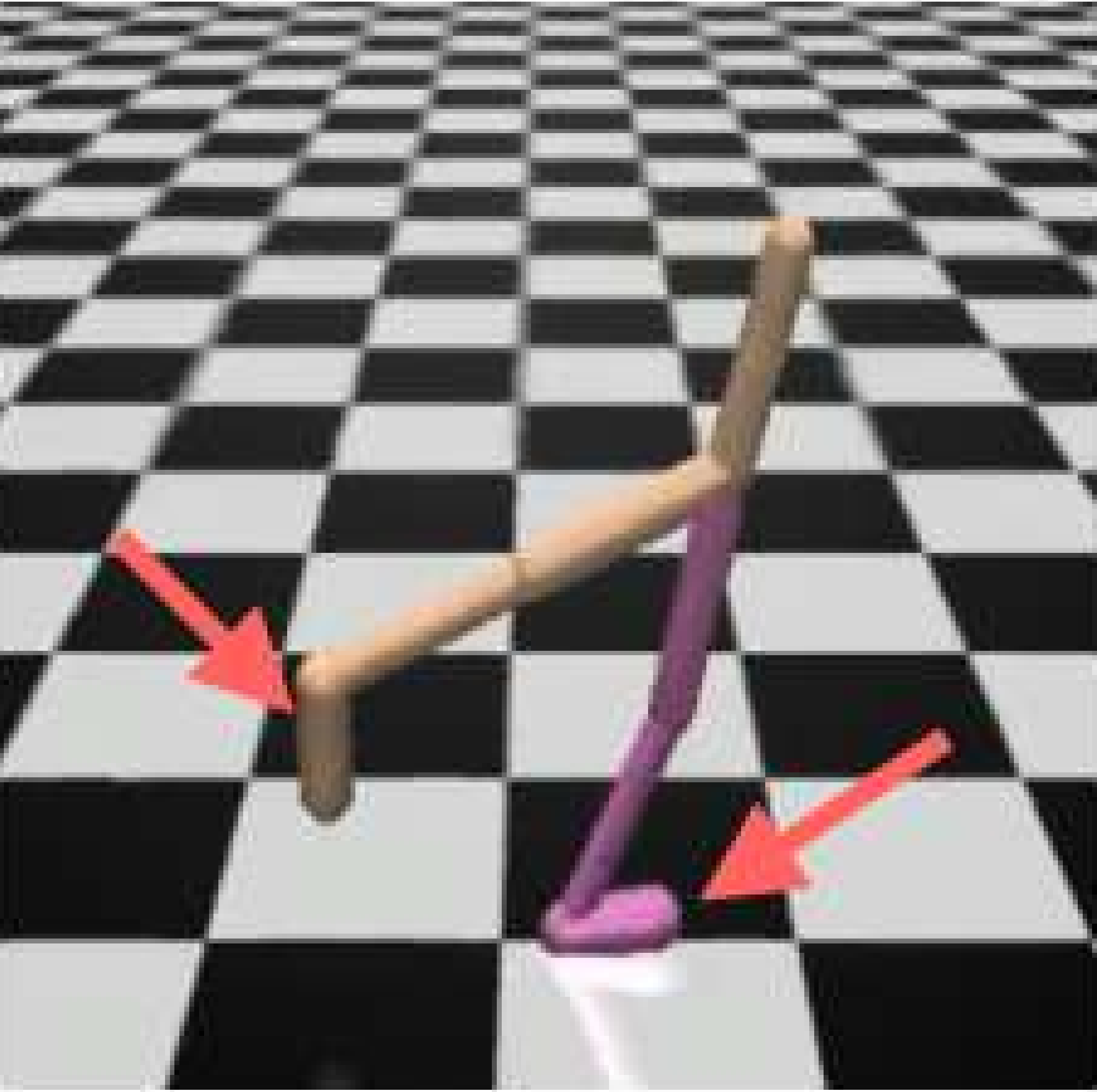}
         \label{fig:walker2d_visual}
     }
     \subfigure[Half-Cheetah]{
         \includegraphics[width=0.2\linewidth]{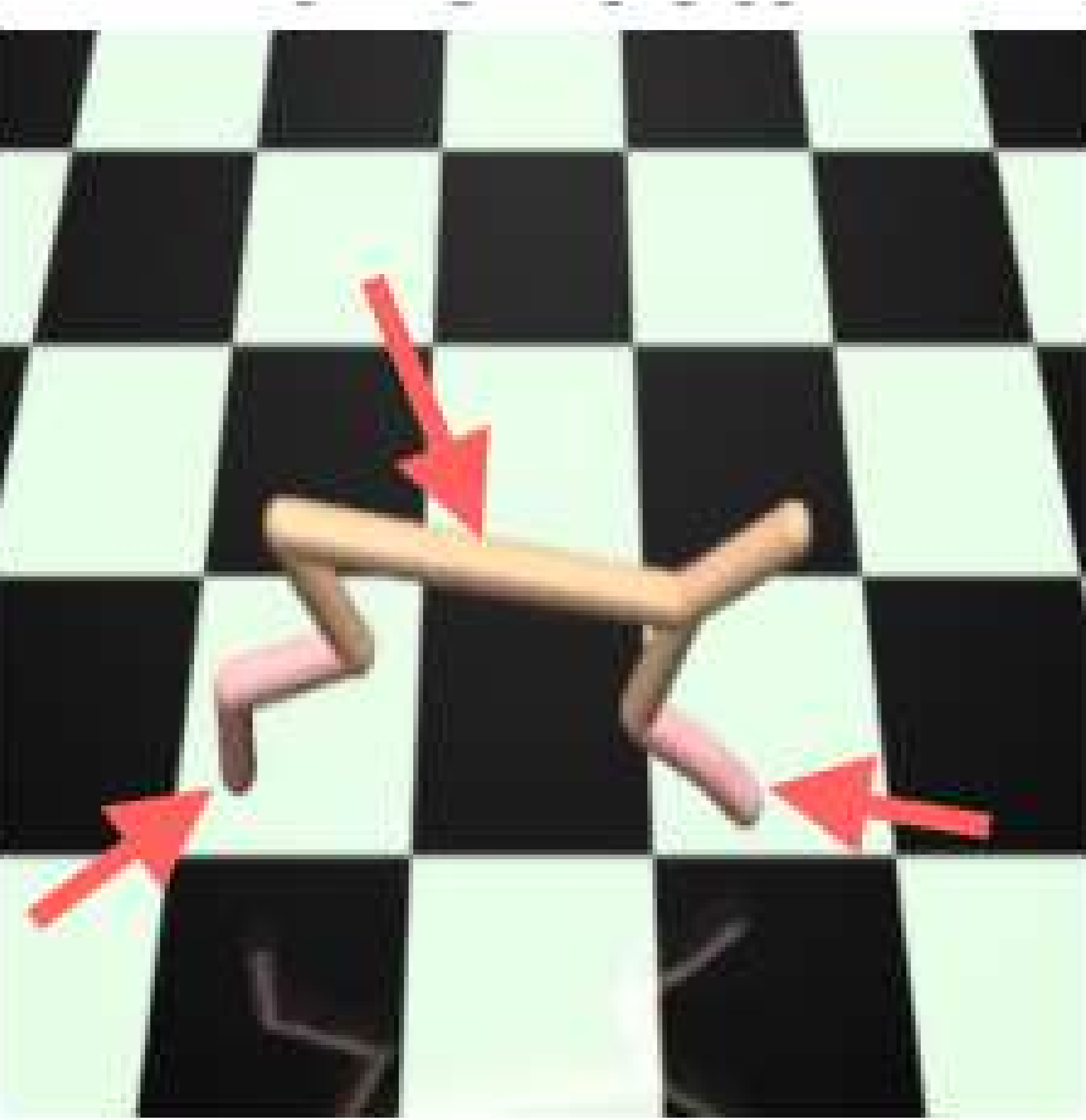}
         \label{fig:halfcheetah_visual}
     }
        \caption{Illustrations of the environments evaluated in our experiments.}
        \label{fig:task_visual}
\end{figure}

\section{Experiments}\label{sec:exp}
 
In this section, we empirically evaluate ROLAH with the following baselines: (\textit{i}) RL agents trained without adversarial training, (\textit{ii}) RARL (Robust Adversarial Reinforcement Learning): RL agent trained against a single adversary in a zero-sum game~\cite{RARL}, and (\textit{iii}) RAP (Robustness via Adversary Populations): agent trained with a uniform sampling from a population of adversaries \cite{robust_population}. We investigate $2$ types of  robustness: (\textit{a}) robustness to environmental change (e.g., mass and friction) and (\textit{b}) robustness to disturbance on the agent (e.g., action noise and adversarial policies). We also conducted experiments to test robustness to random initialization. However, due to space limitation, we defer these results to the Appendix. For fairness and consistency of the performance, we use Trust Region Policy Optimization (TRPO) \cite{pmlr-v37-schulman15} to update policies for all baselines as well as ROLAH. 

We conduct experiments on the Hopper, Walker2d, and Half-Cheetah continuous control tasks in MuJoCo environments. Our adversarial setting follows~\cite{RARL}, where the adversary learns to destabilize the protagonist by applying forces on specific points, which is denoted by red arrows in Figure \ref{fig:task_visual}. All our experiments are run on Nvidia RTX A5000 with 24GB RAM and our implementation are partly based on the codes published by \emph{rllab}~\cite{rllab}. Please refer to the appendix for additional comparison and the exact hyper-parameters we have used during training and evaluation.

 \begin{figure*}
        \centering
        \subfigure[Baseline (0 adv)]{
            \centering
    \includegraphics[width=0.23\textwidth]{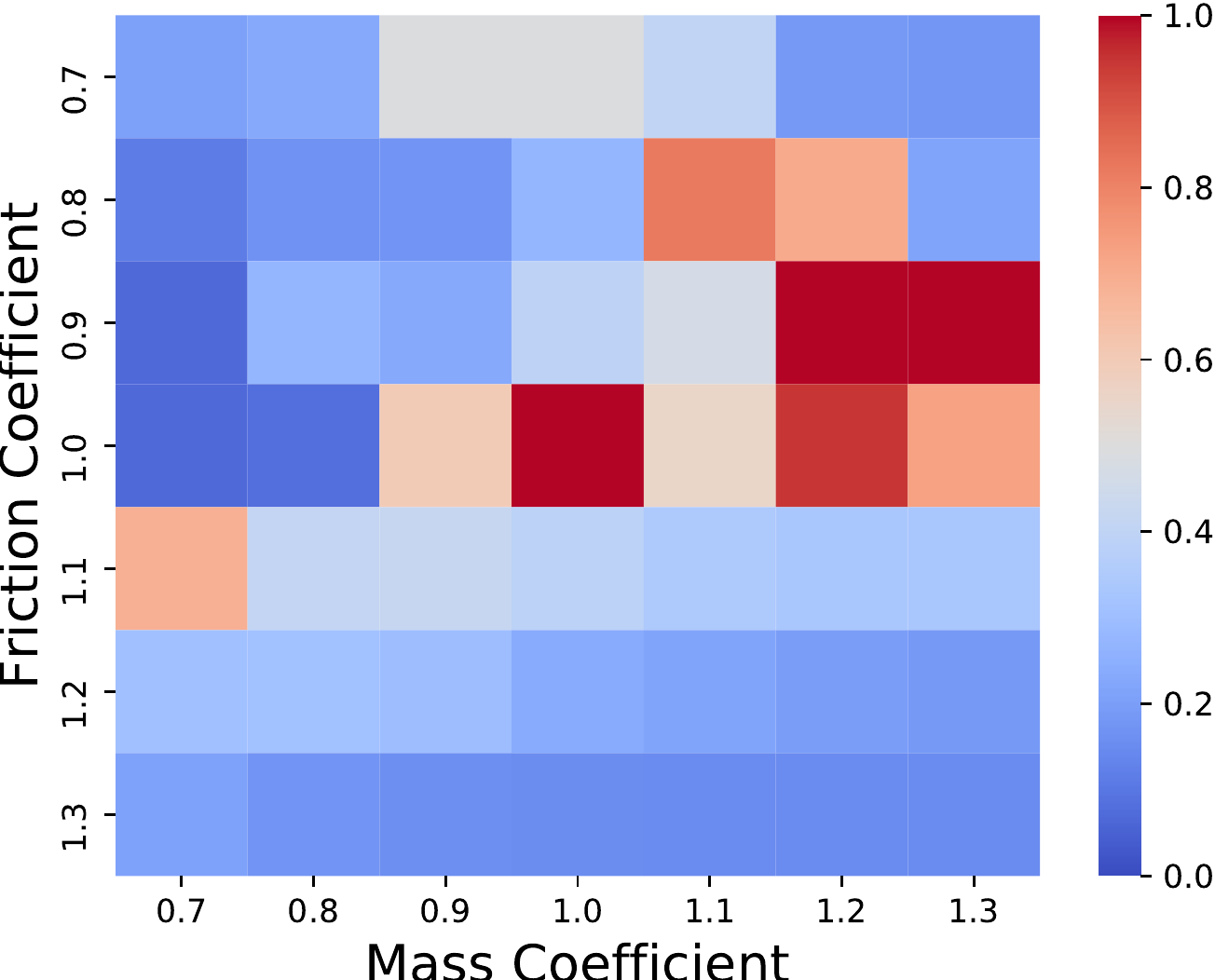}   
        \label{fig:hopper_heatmap_zero}}
        \vspace{-16pt}
        \subfigure[RARL (1 adv)]{  
            \centering \includegraphics[width=0.23\textwidth]{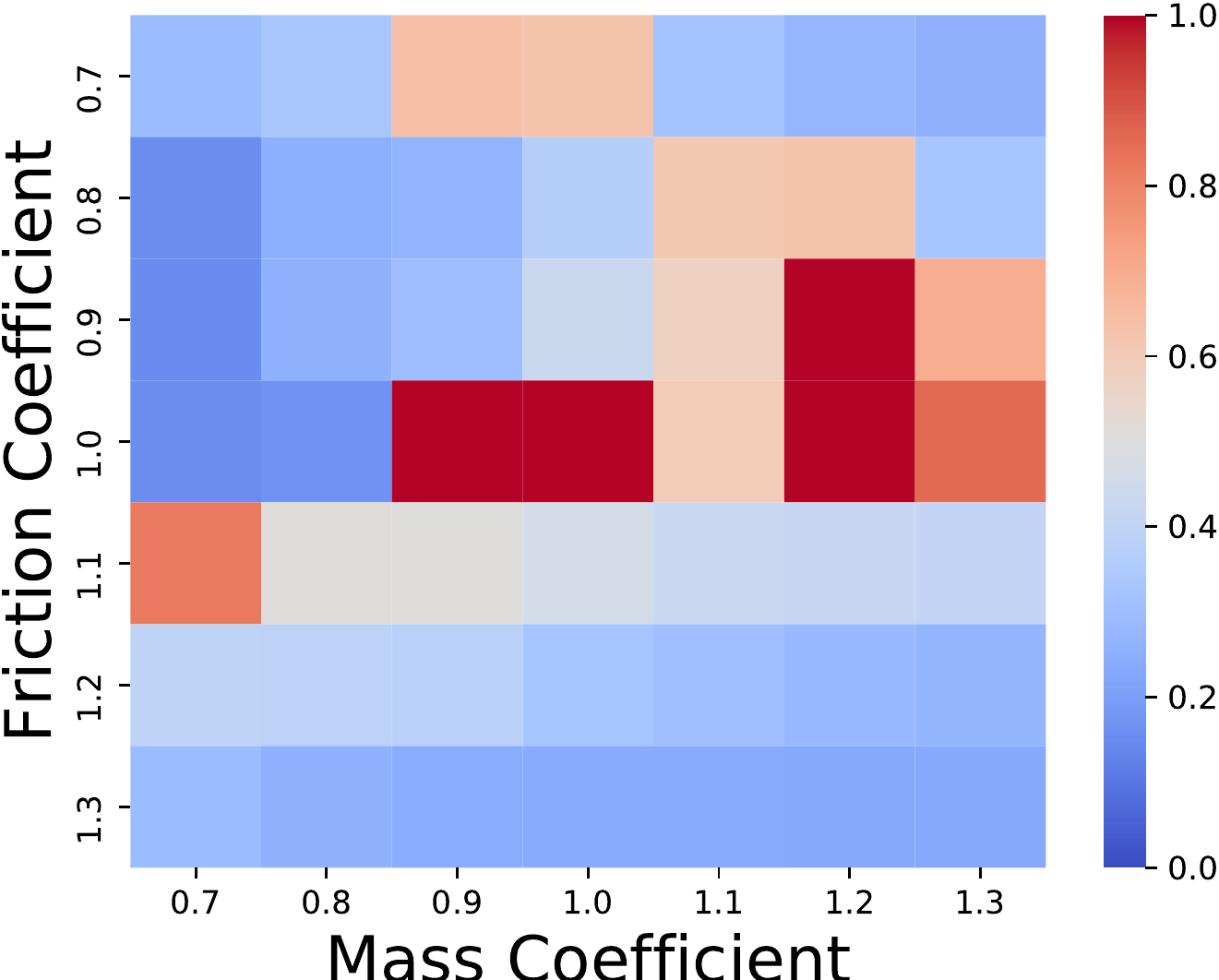}  
    \label{fig:hopper_heatmap_single}}
        \subfigure[RAP (population)]{  
            \centering 
    \includegraphics[width=0.23\textwidth]{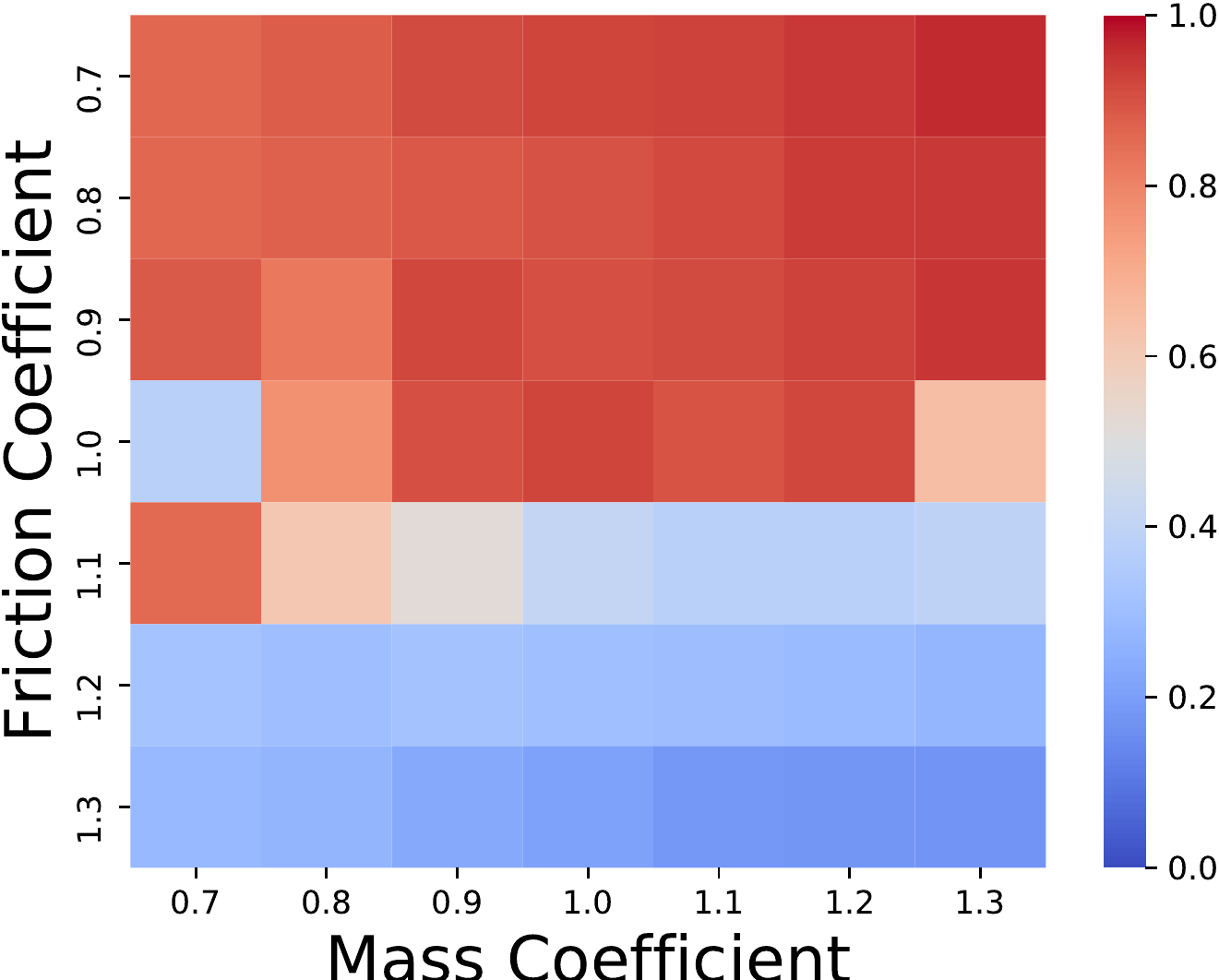}\label{fig:hopper_heatmap_popu}
        }
        \subfigure[ROLAH]{ 
            \centering 
       \includegraphics[width=0.23\textwidth]{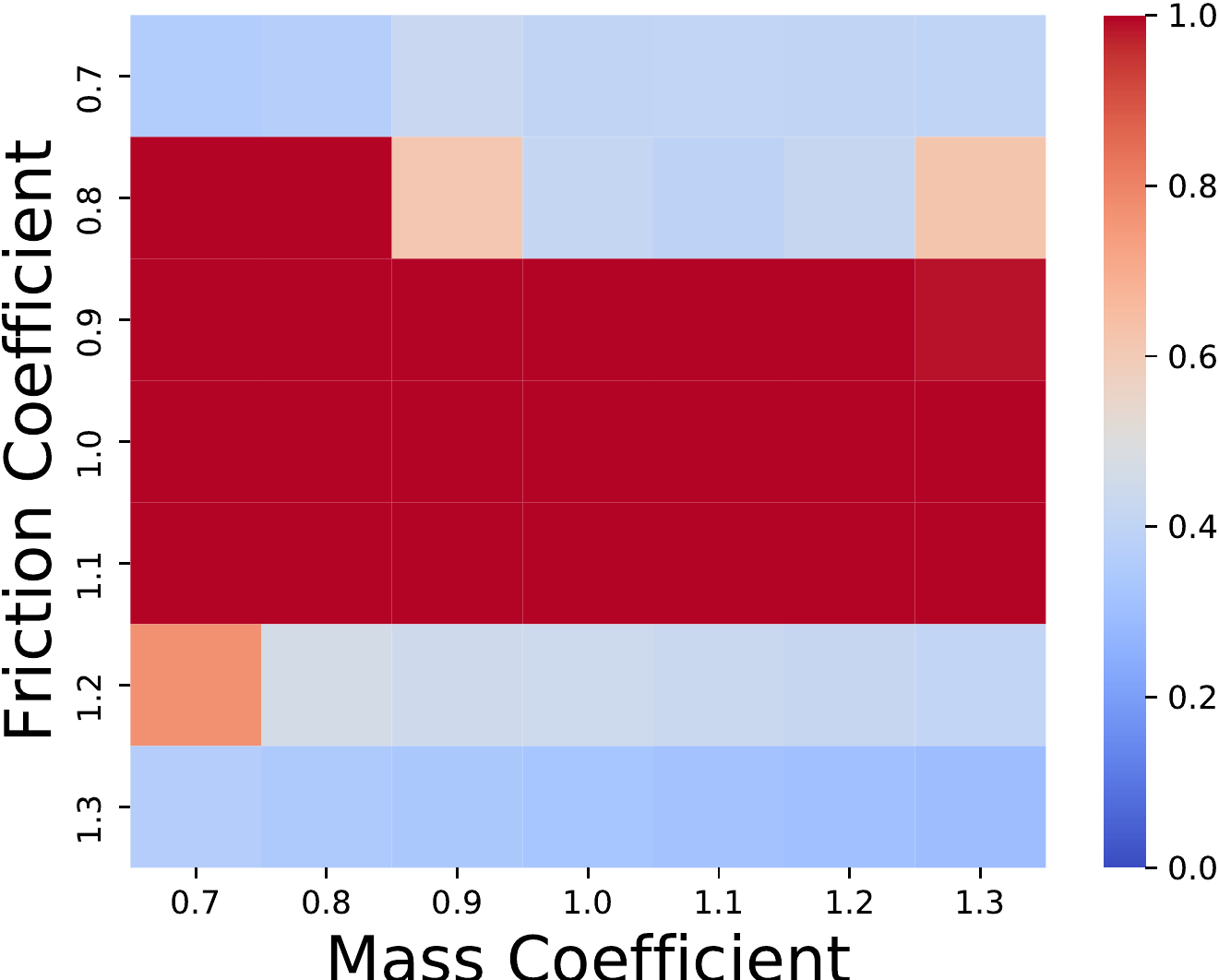}       \label{fig:hopper_heatmap_percent_worst}
        }
    \vskip\baselineskip
     \centering
        \subfigure[Baseline (0 adv)]{
            \centering
    \includegraphics[width=0.23\textwidth]{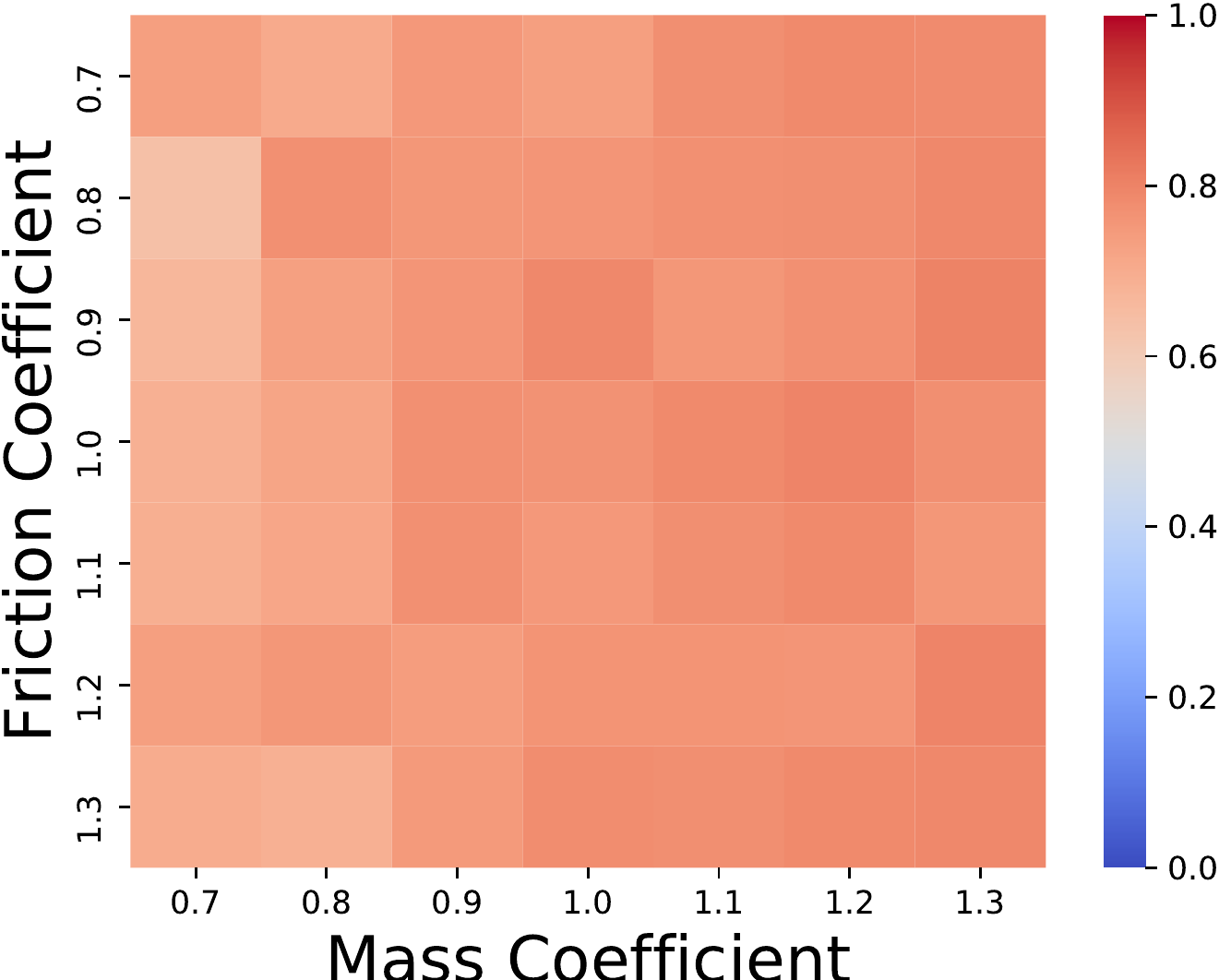}   
        \label{fig:halfcheetah_heatmap_zero}}
        \vspace{-16pt}
        \subfigure[RARL (1 adv)]{  
            \centering \includegraphics[width=0.23\textwidth]{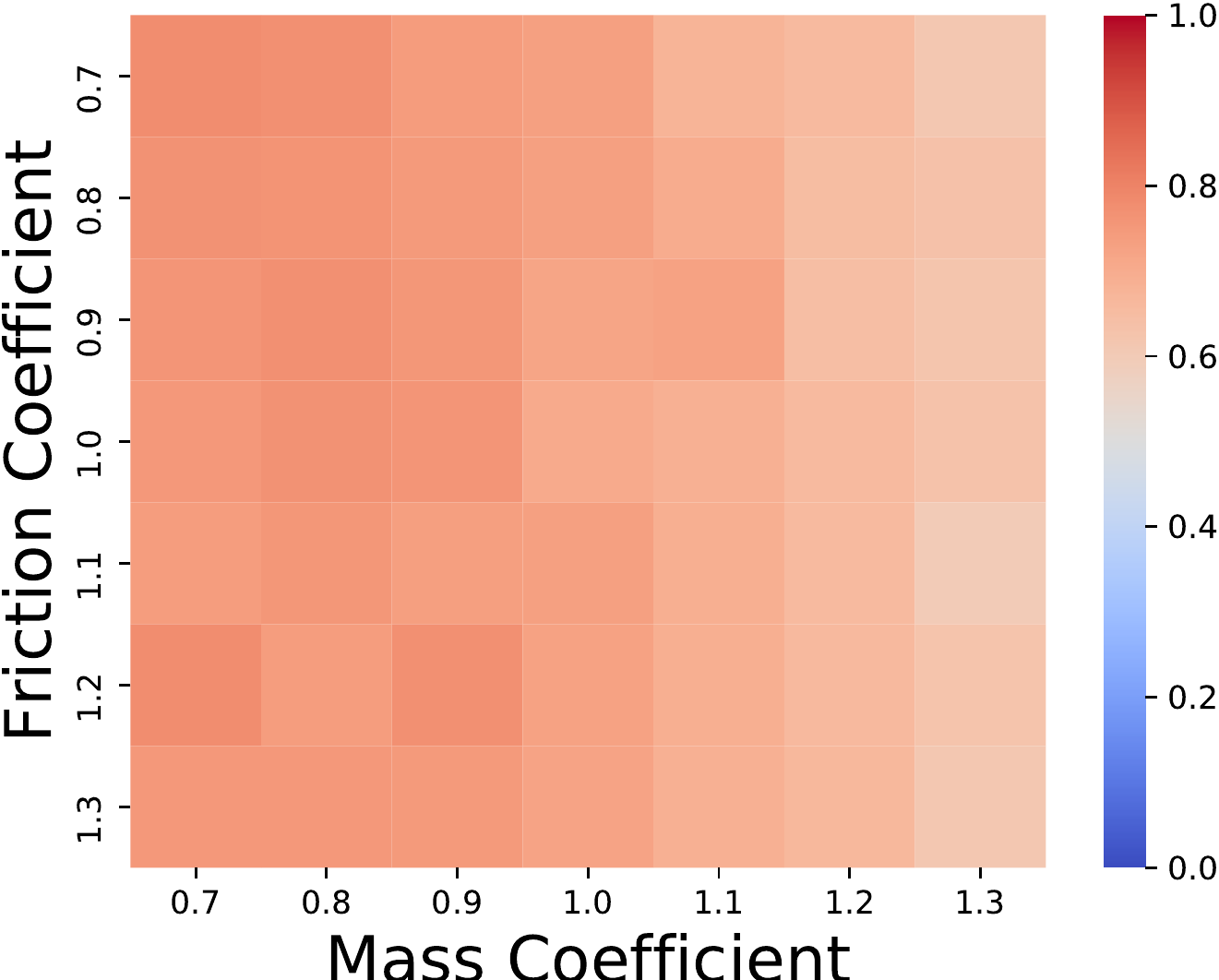}  
    \label{fig:halfcheetah_heatmap_single}}
        \subfigure[RAP (population)]{  
            \centering 
    \includegraphics[width=0.23\textwidth]{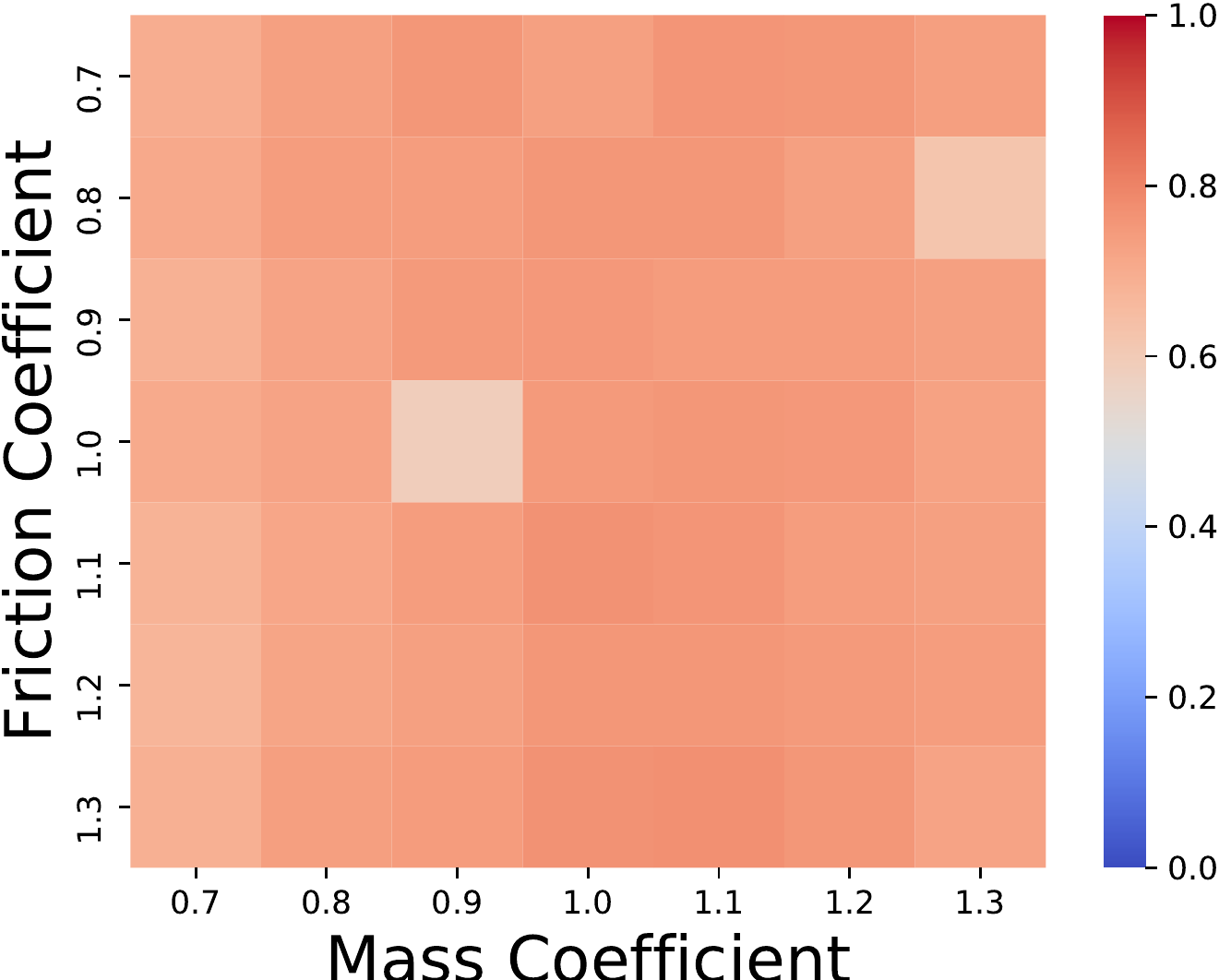}\label{fig:halfcheetah_heatmap_popu}
        }
        \subfigure[ROLAH]{ 
            \centering 
       \includegraphics[width=0.23\textwidth]{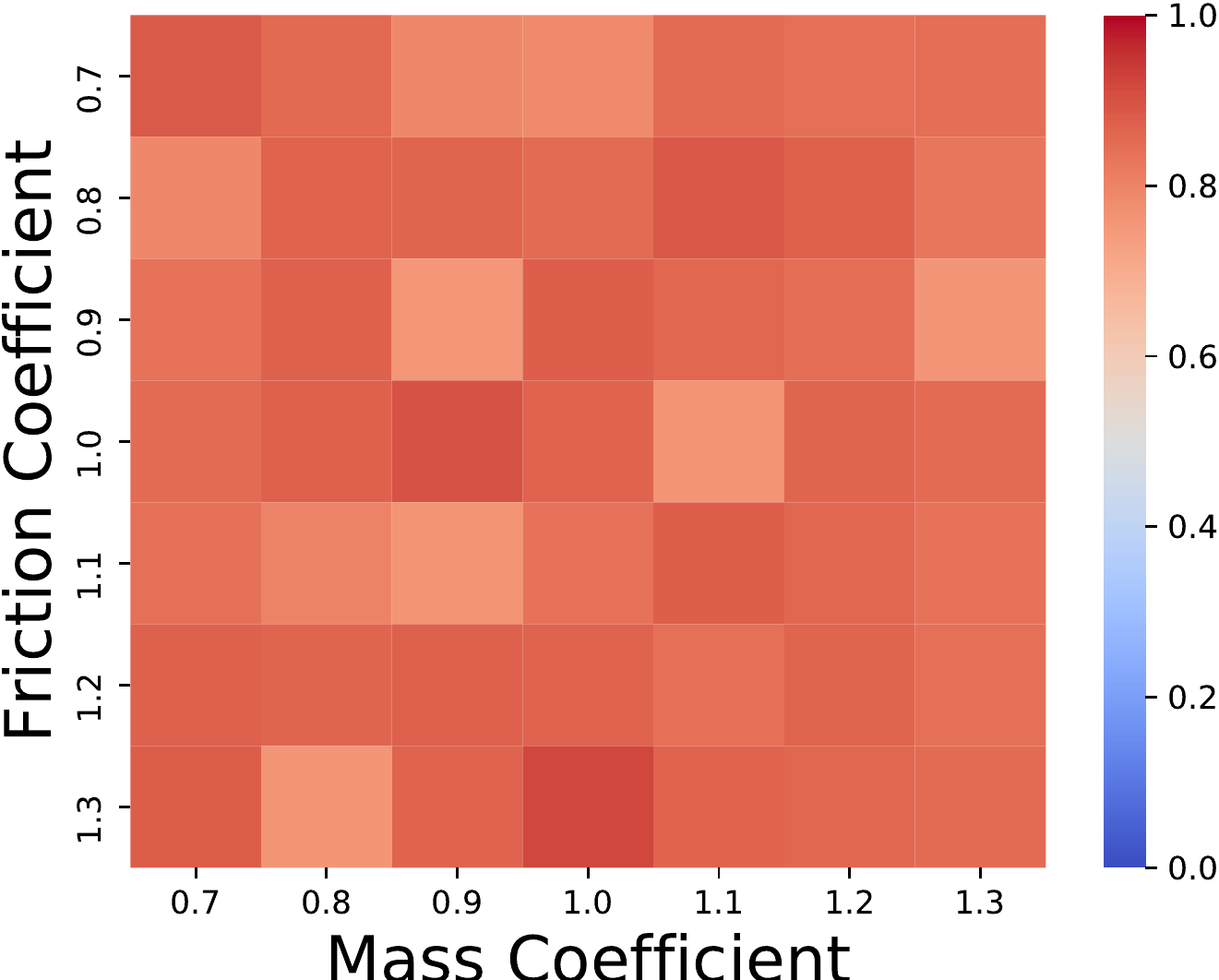}       \label{fig:halfcheetah_heatmap_percent_worst}
        }
         \vskip\baselineskip
     \centering
        \subfigure[Baseline (0 adv)]{
            \centering
    \includegraphics[width=0.23\textwidth]{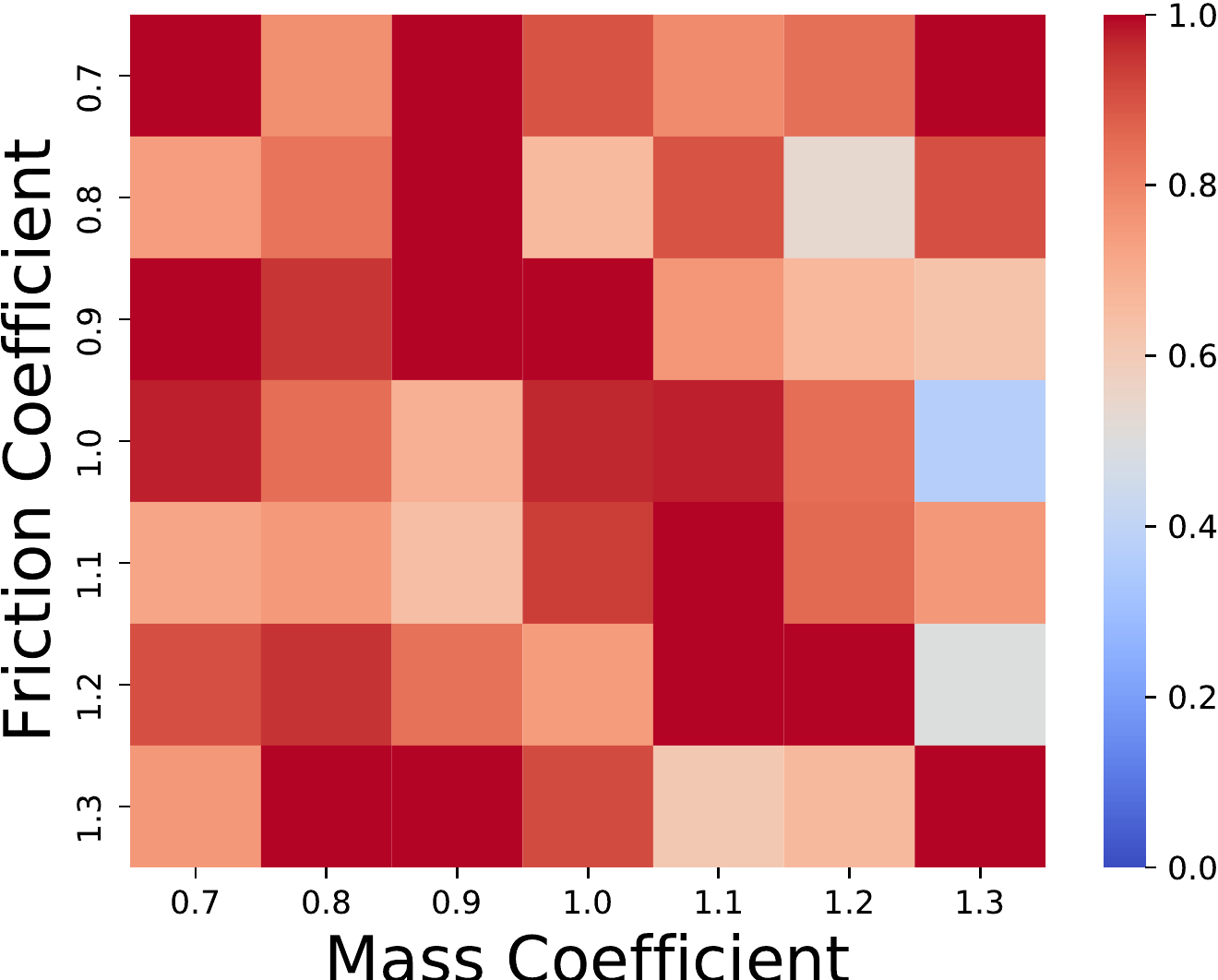}   
        \label{fig:walker2d_heatmap_zero}}
        \subfigure[RARL (1 adv)]{  
            \centering \includegraphics[width=0.23\textwidth]{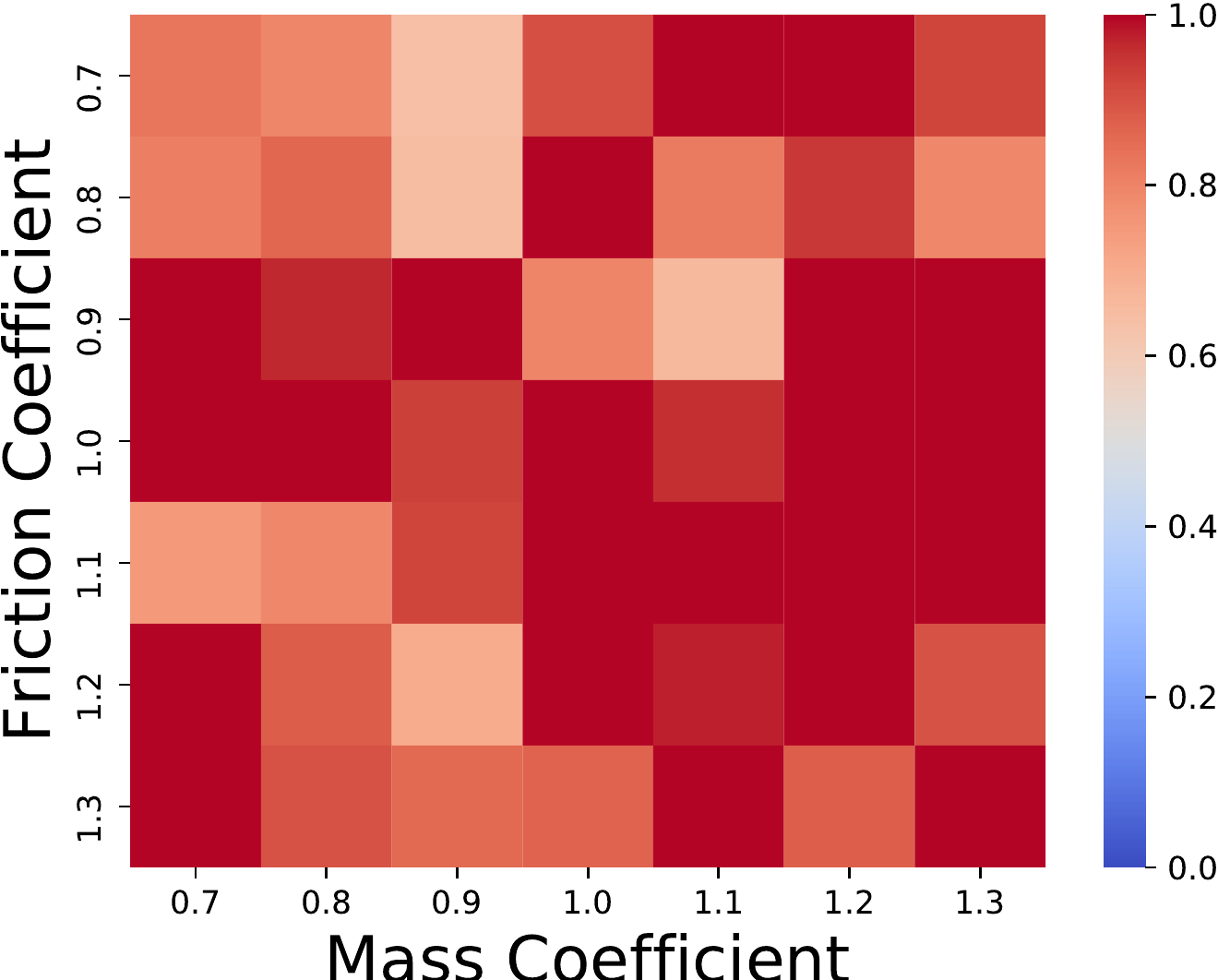}  
    \label{fig:walker2d_heatmap_single}}
        \subfigure[RAP (population)]{  
            \centering 
    \includegraphics[width=0.23\textwidth]{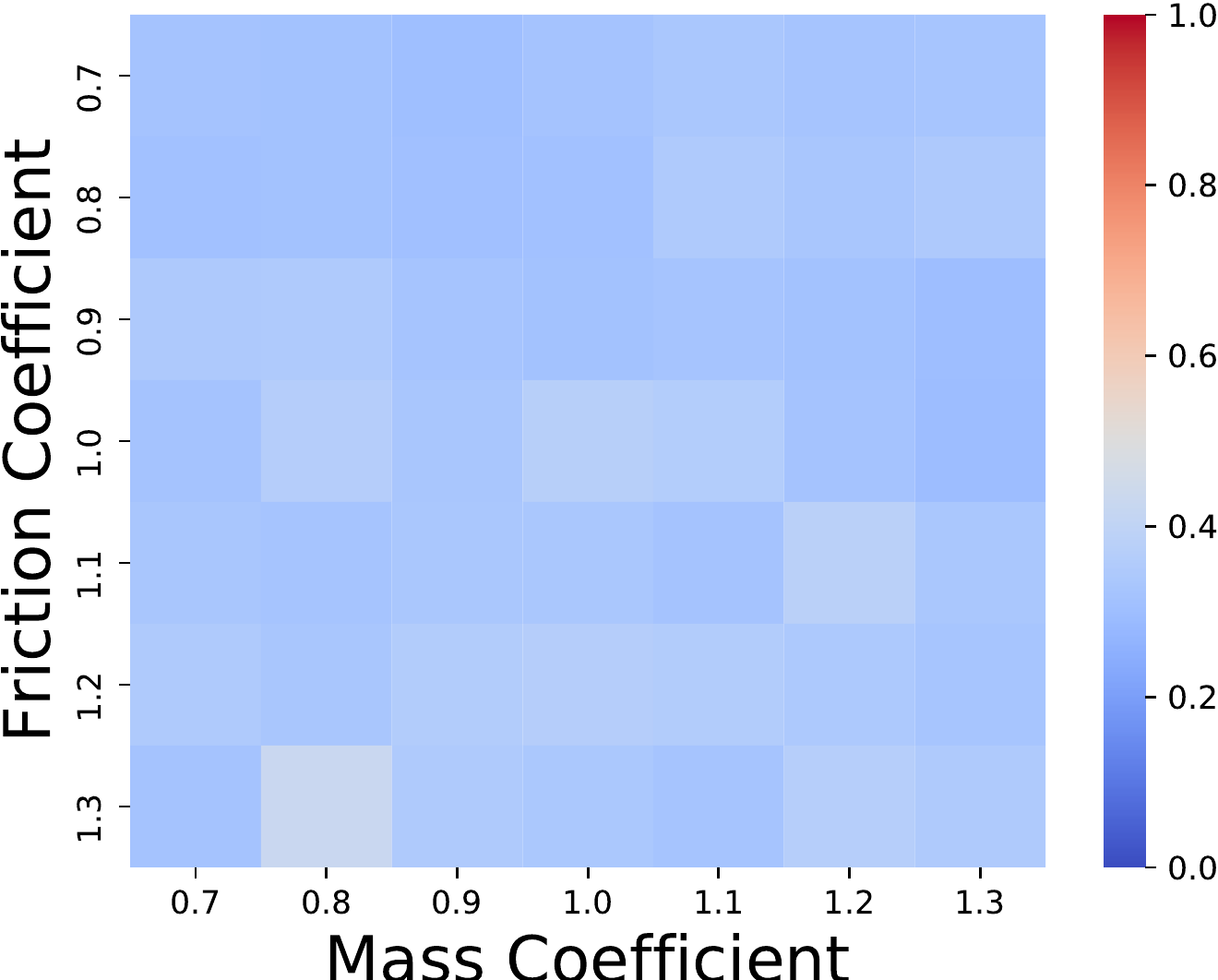}\label{fig:walker2d_heatmap_popu}
        }
        \subfigure[ROLAH]{ 
            \centering 
       \includegraphics[width=0.23\textwidth]{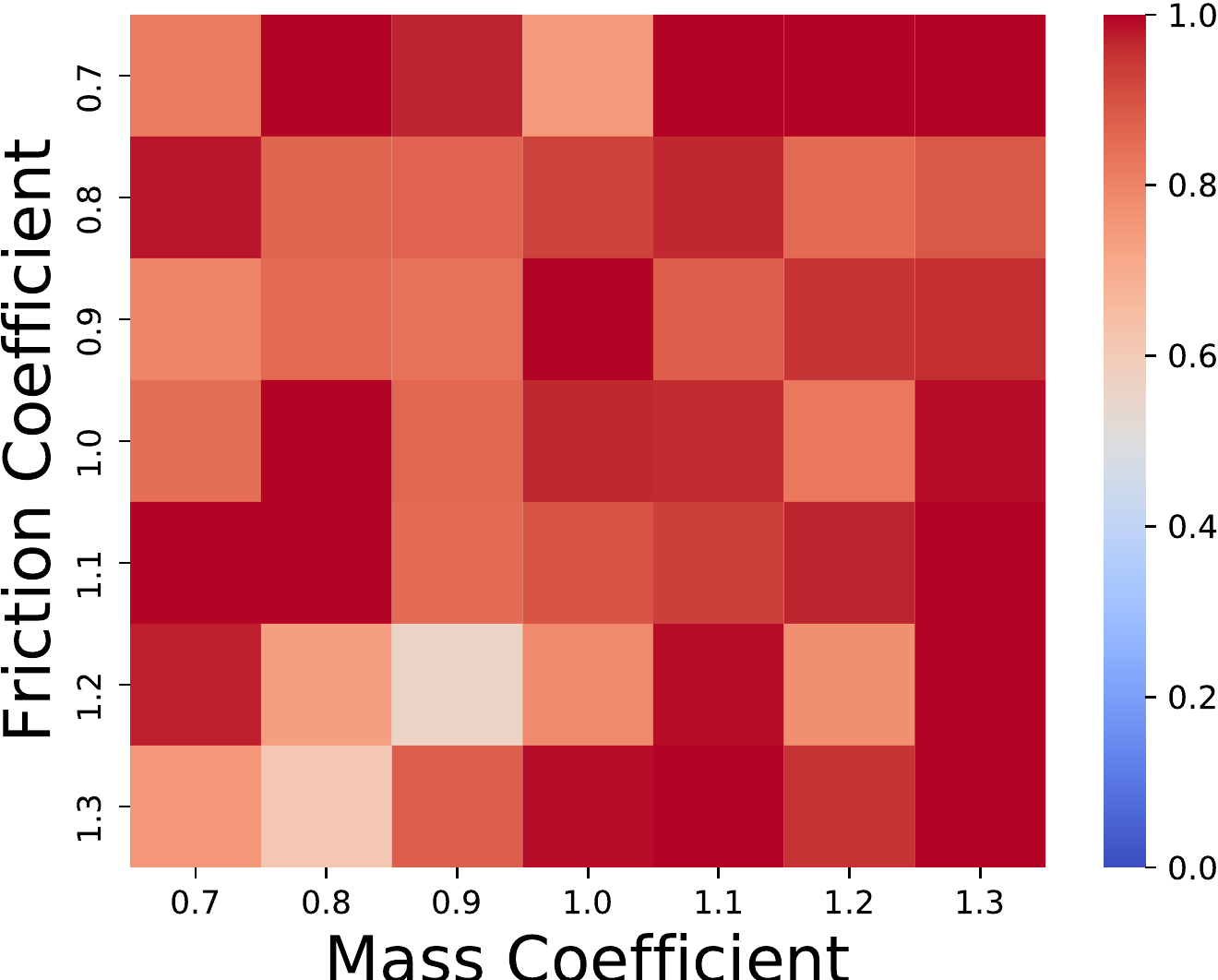}       \label{fig:walker2d_heatmap_percent_worst}
        }

         \caption {Average normalized return across 10 seeds tested via different mass coefficients on the x-axis and friction coefficients on the y-axis. High reward has red color; low reward has blue color. $1^{st}$ row: Hopper, $2^{nd}$ row: Half-Cheetah, $3^{rd}$: Walker2d} 
        \label{fig:heatmap}
    \end{figure*}

     \begin{figure*}
        \centering
        \subfigure[Baseline (0 adv)]{
            \centering
    \includegraphics[width=0.23\textwidth]{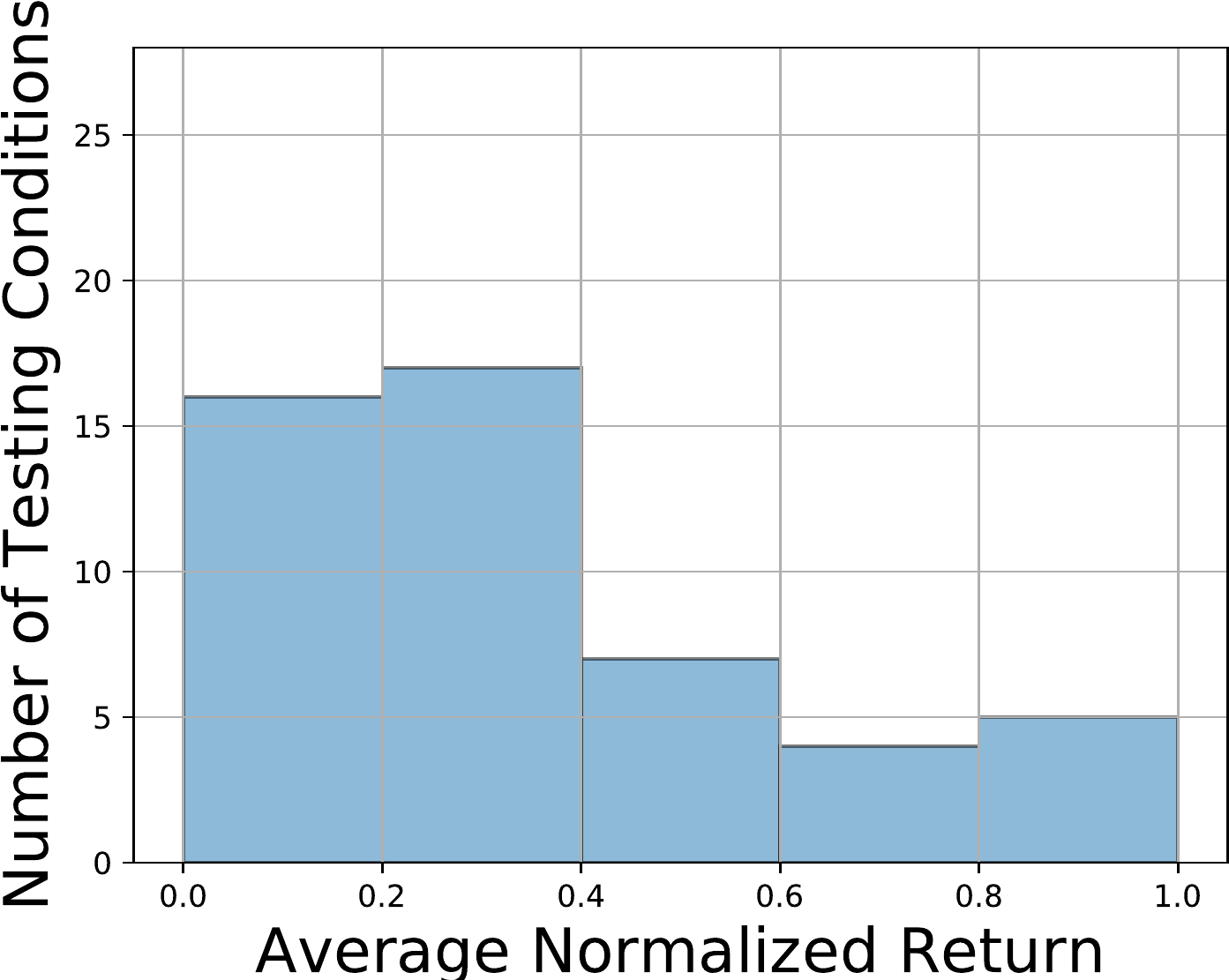}   
        \label{fig:hopper_hist_zero}}
        \vspace{-16pt}
        \subfigure[RARL (1 adv)]{  
            \centering \includegraphics[width=0.23\textwidth]{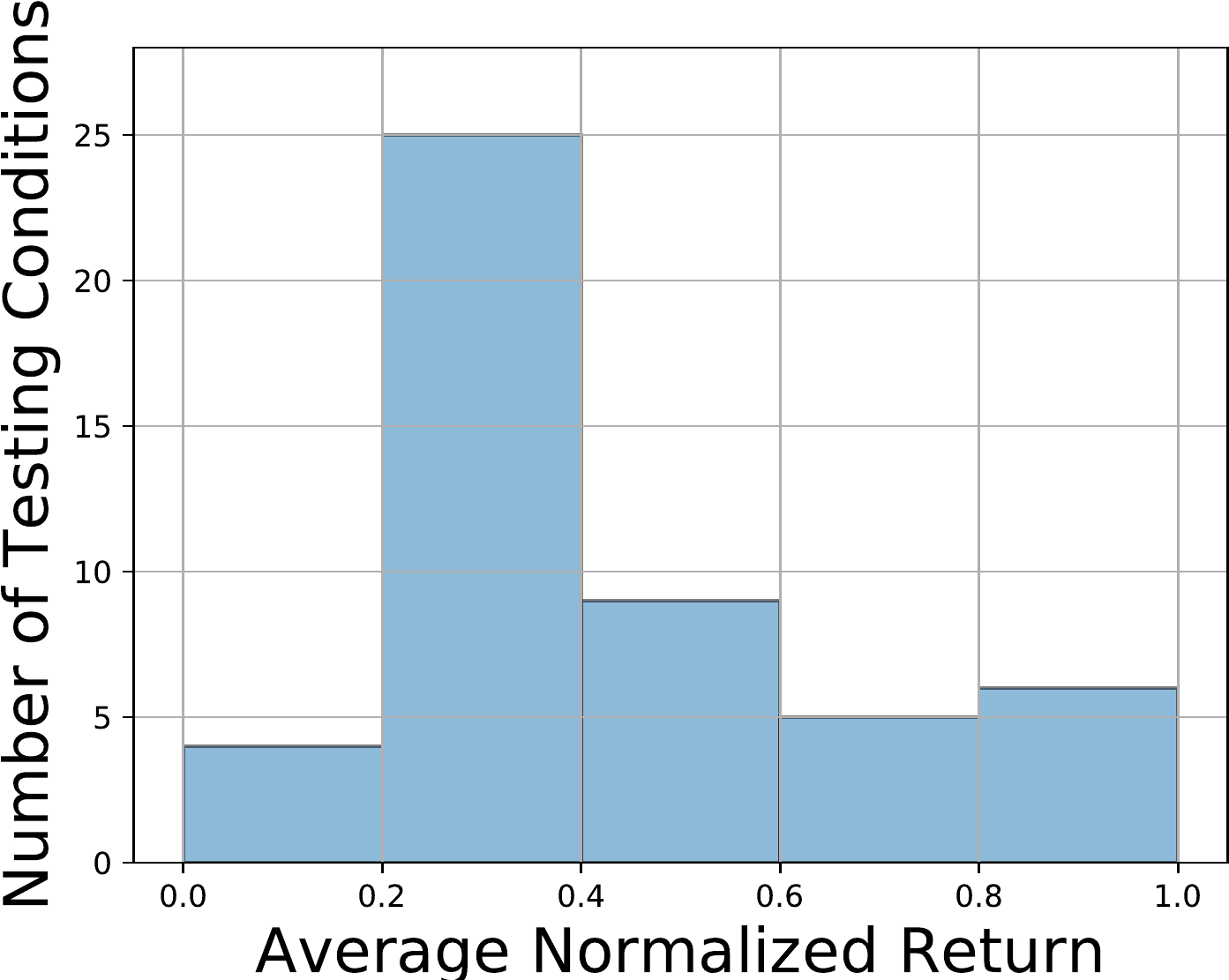}  
    \label{fig:hopper_hist_single}}
        \subfigure[RAP (population)]{  
            \centering 
    \includegraphics[width=0.23\textwidth]{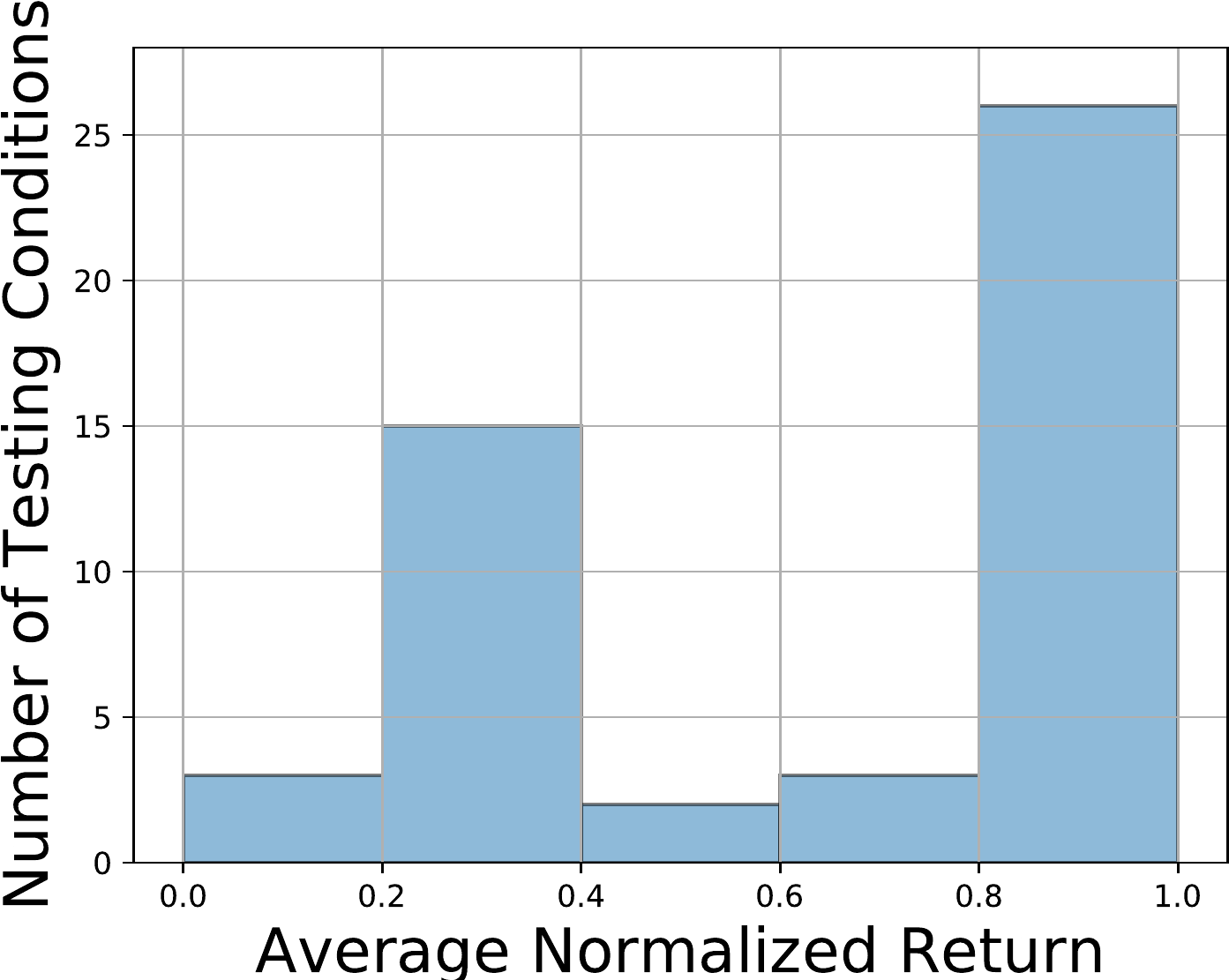}\label{fig:hopper_hist_popu}
        }
        \subfigure[ROLAH]{ 
            \centering 
       \includegraphics[width=0.23\textwidth]{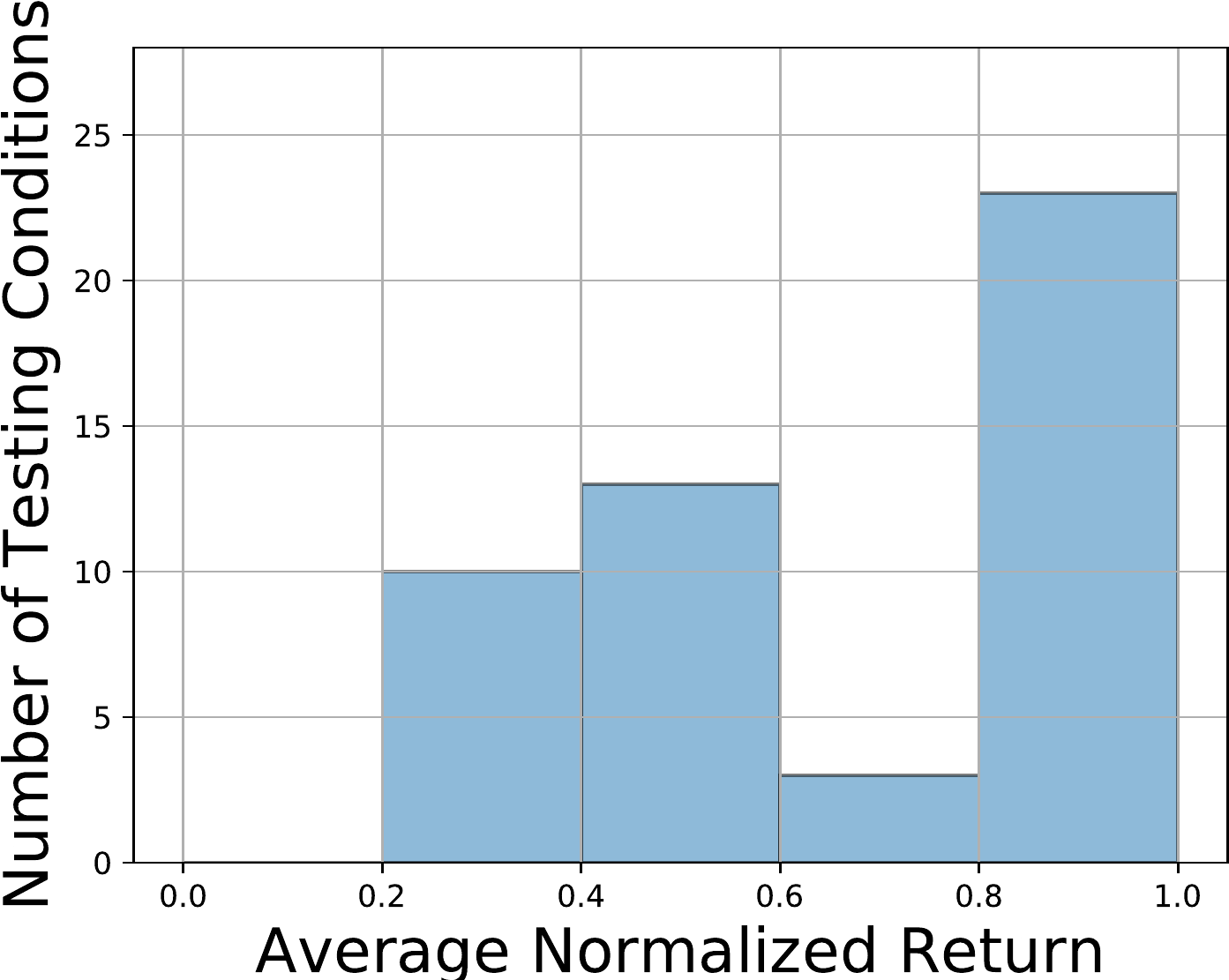}       \label{fig:hopper_hist_worst}
        }
    \vskip\baselineskip
     \centering
        \subfigure[Baseline (0 adv)]{
            \centering
    \includegraphics[width=0.23\textwidth]{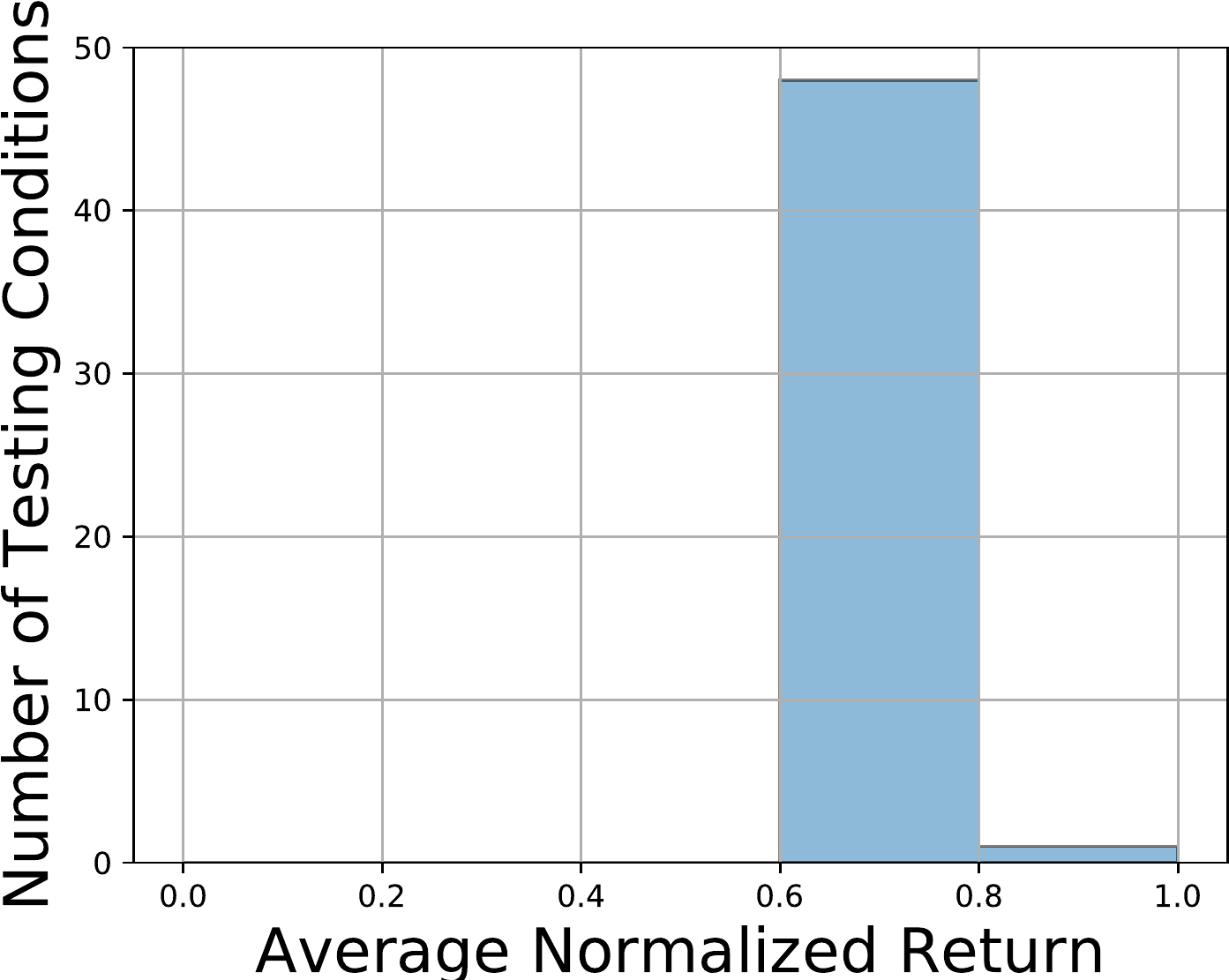}   
        \label{fig:halfcheetah_hist_zero}}
        \vspace{-16pt} 
        \subfigure[RARL (1 adv)]{  
            \centering \includegraphics[width=0.23\textwidth]{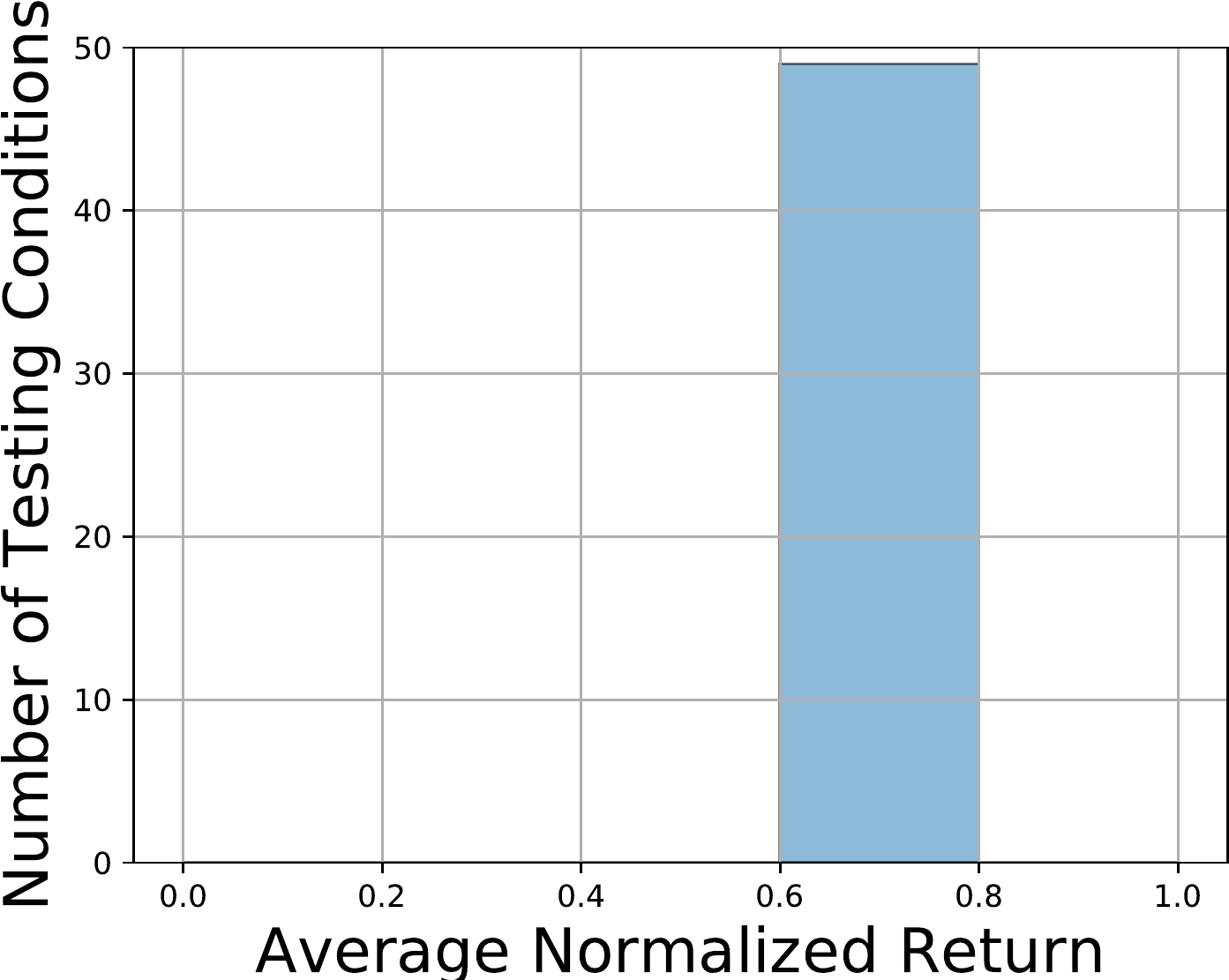}  
    \label{fig:halfcheetah_hist_single}}
        \subfigure[RAP (population)]{  
            \centering 
    \includegraphics[width=0.23\textwidth]{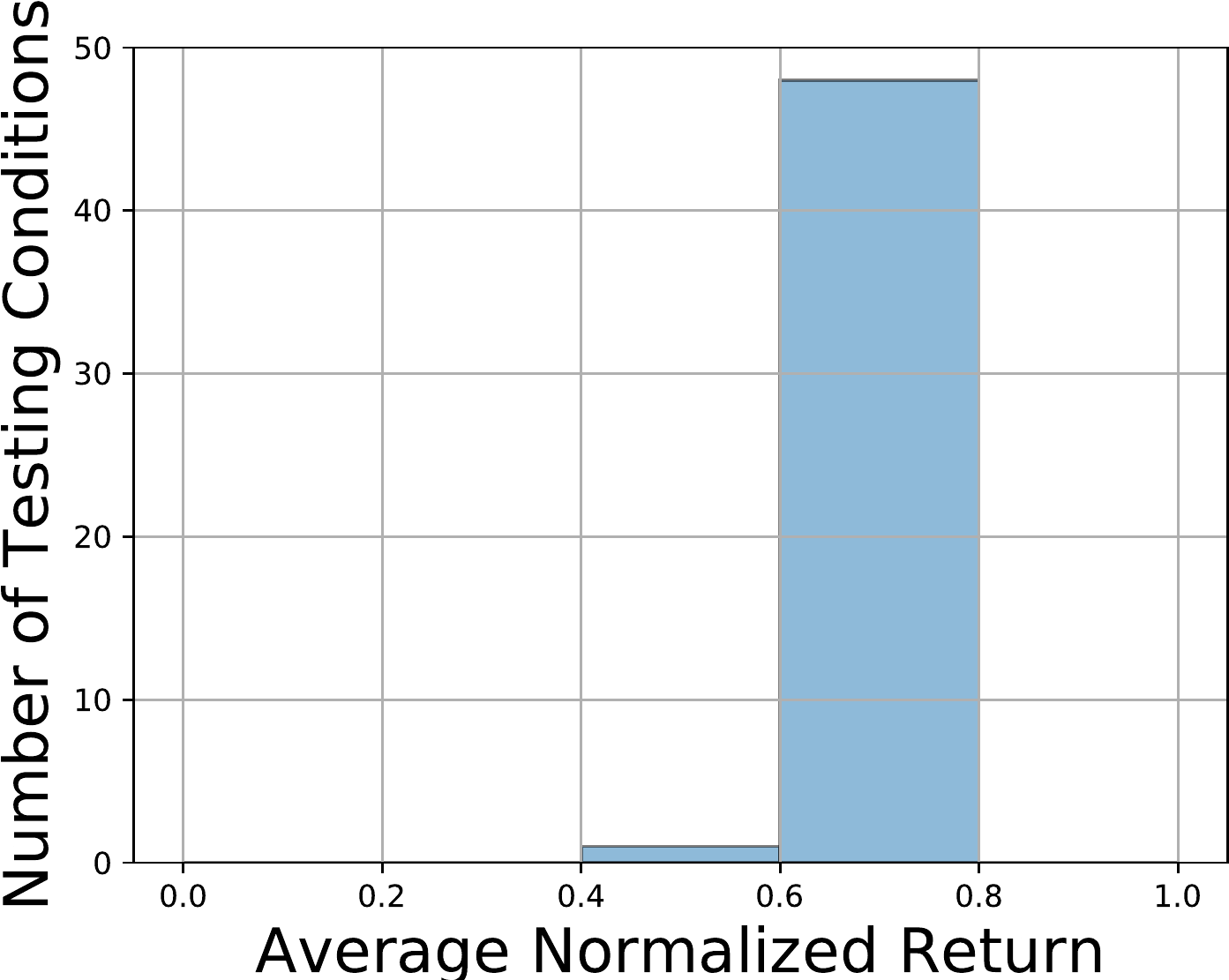}\label{fig:halfcheetah_hist_popu}
        }
        \subfigure[ROLAH]{ 
            \centering 
       \includegraphics[width=0.23\textwidth]{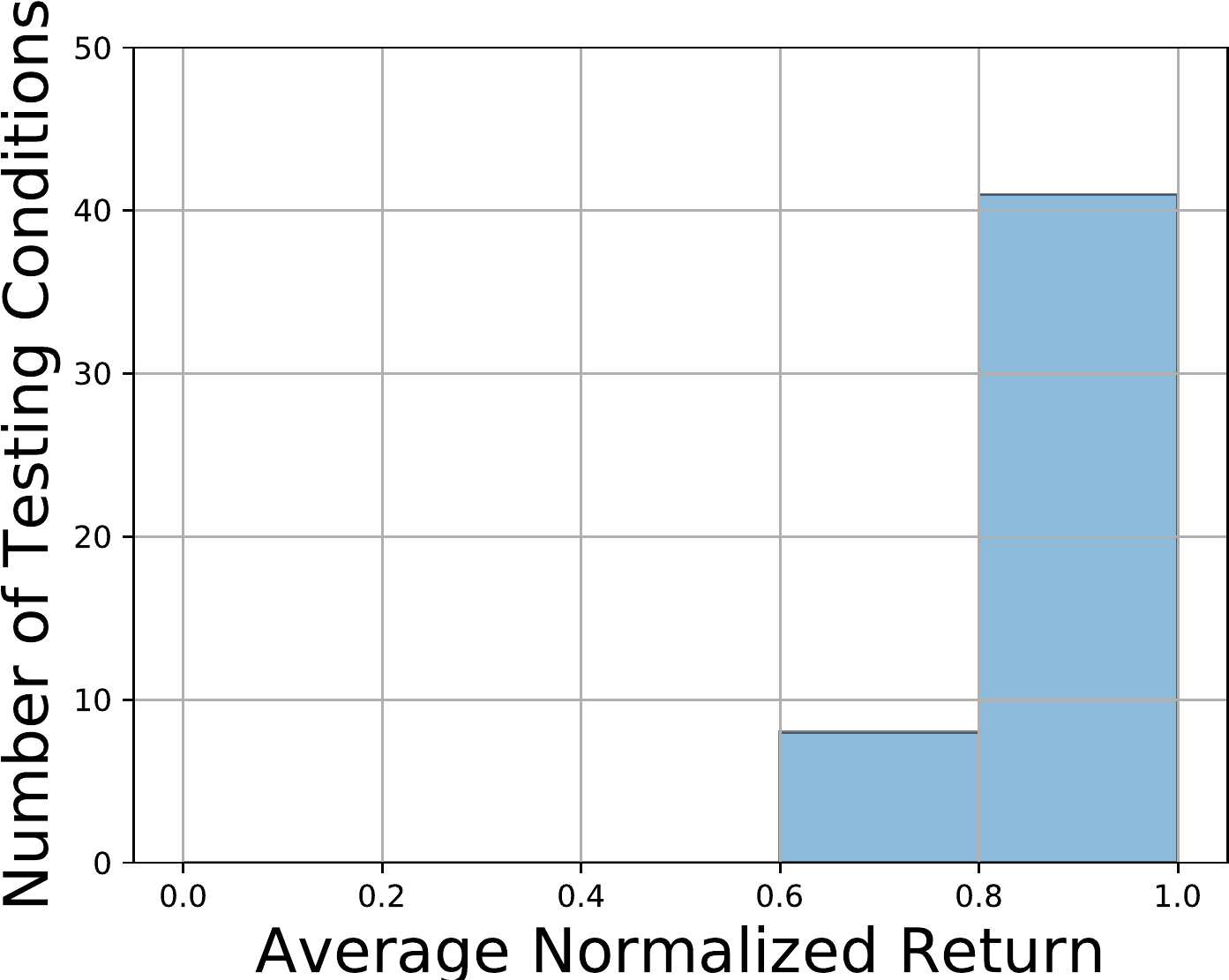}       \label{fig:halfcheetah_hist_worst}
        }
         \vskip\baselineskip
     \centering
        \subfigure[Baseline (0 adv)]{
            \centering
    \includegraphics[width=0.23\textwidth]{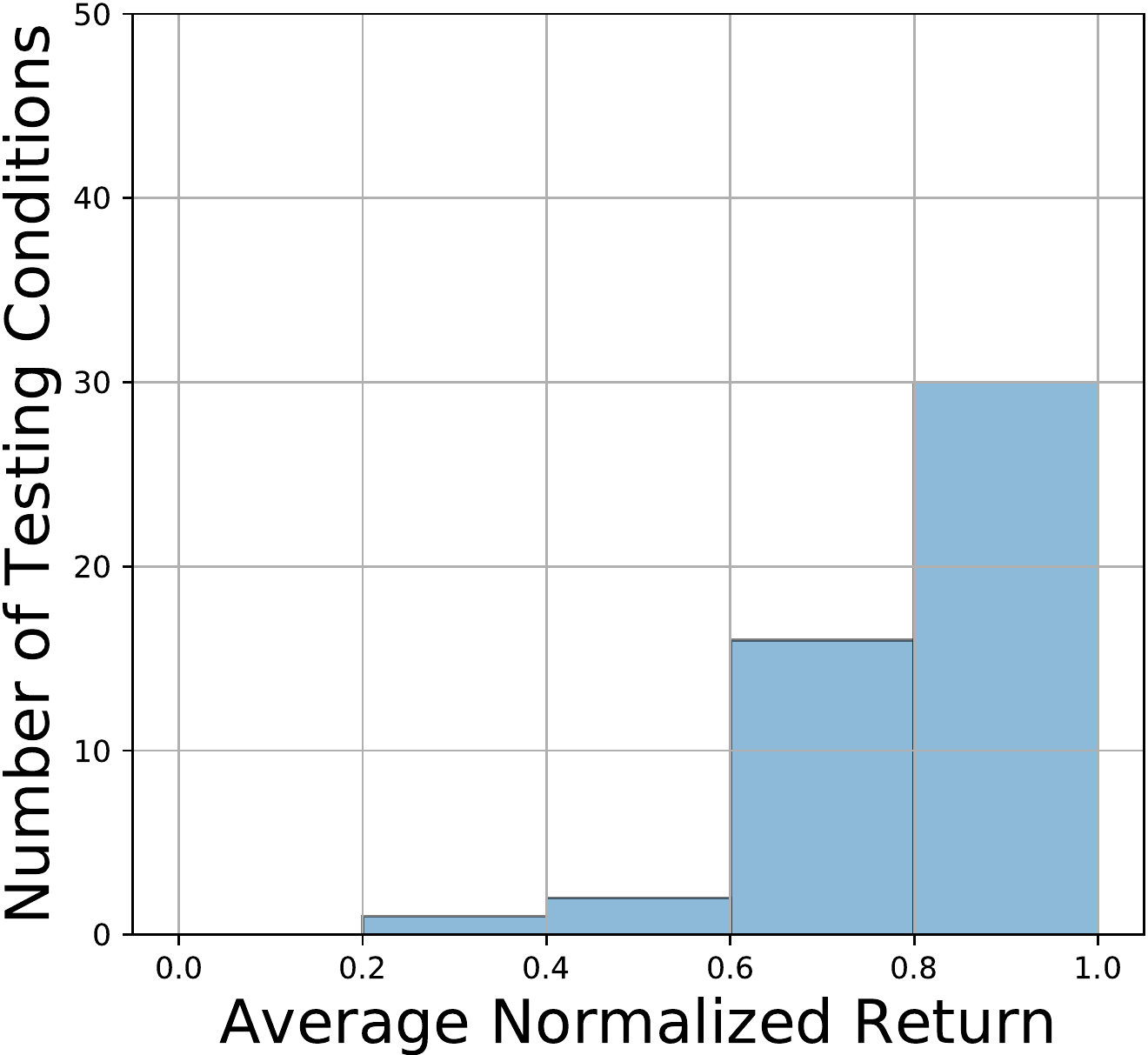}   
        \label{fig:walker2d_hist_zero}}
        \subfigure[RARL (1 adv)]{  
            \centering \includegraphics[width=0.23\textwidth]{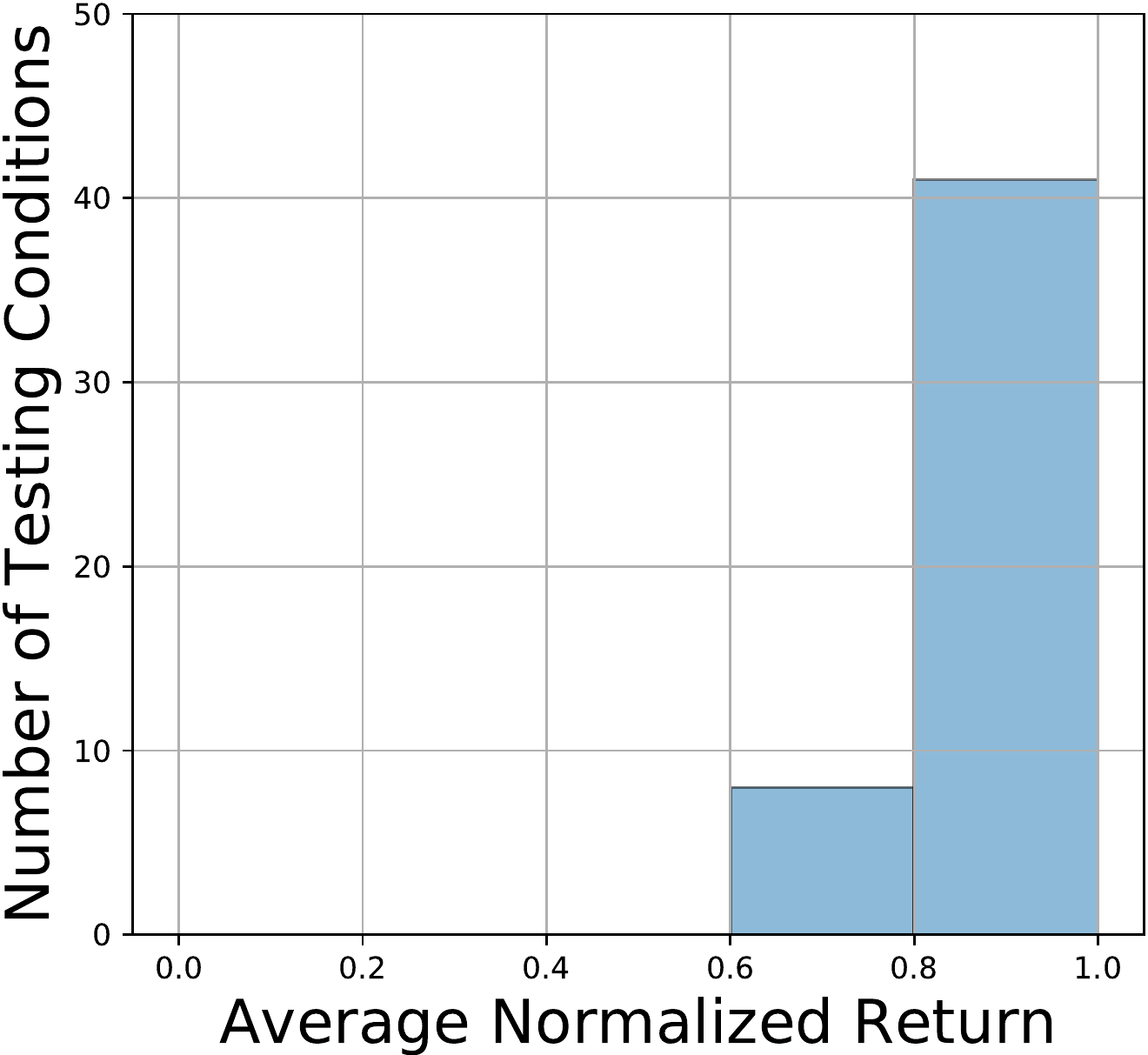}  
    \label{fig:walker2d_heatmap_single_hist}}
        \subfigure[RAP (population)]{  
            \centering 
    \includegraphics[width=0.23\textwidth]{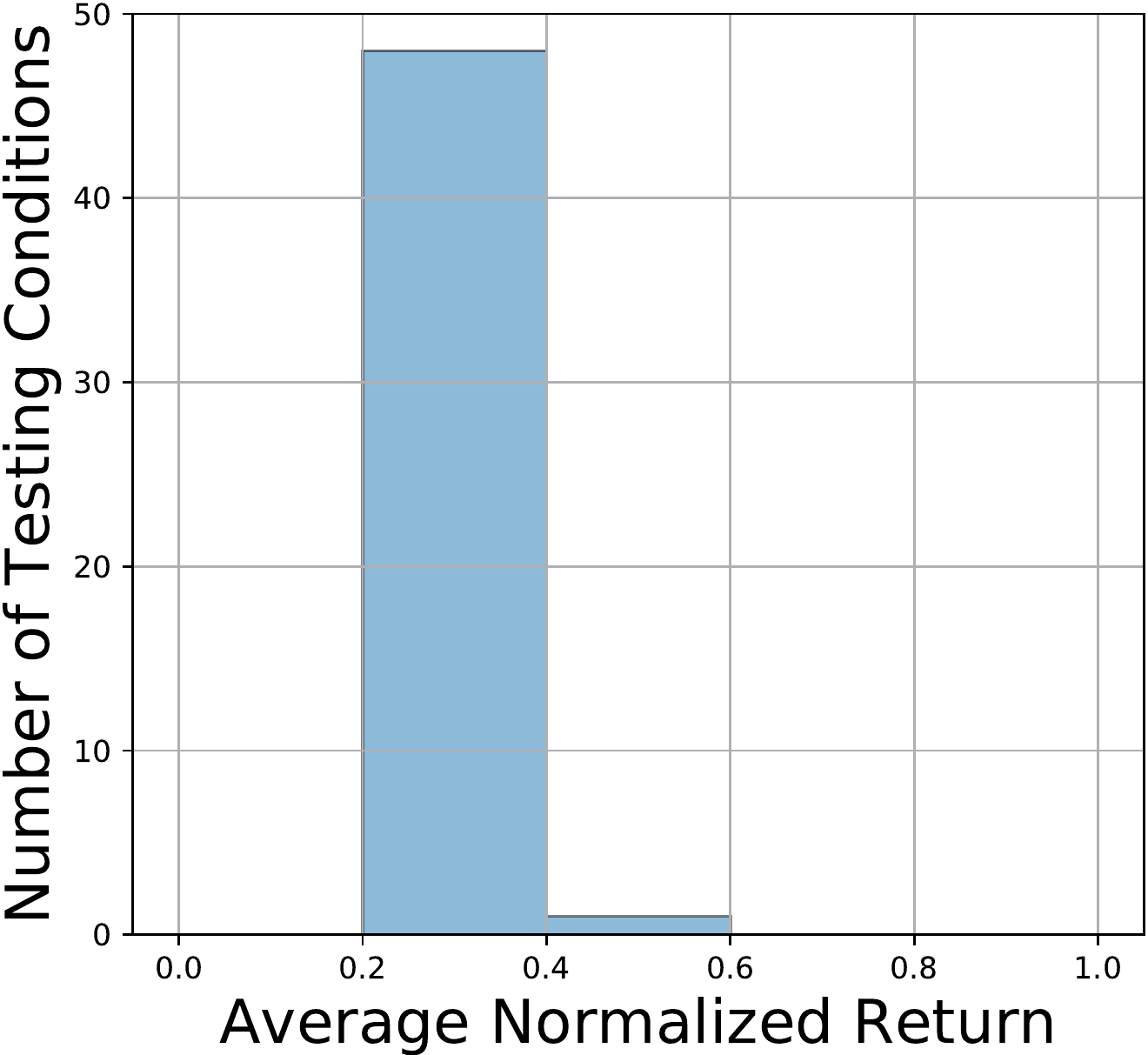}\label{fig:walker2d_hist_popu}
        }
        \subfigure[ROLAH]{ 
            \centering 
       \includegraphics[width=0.23\textwidth]{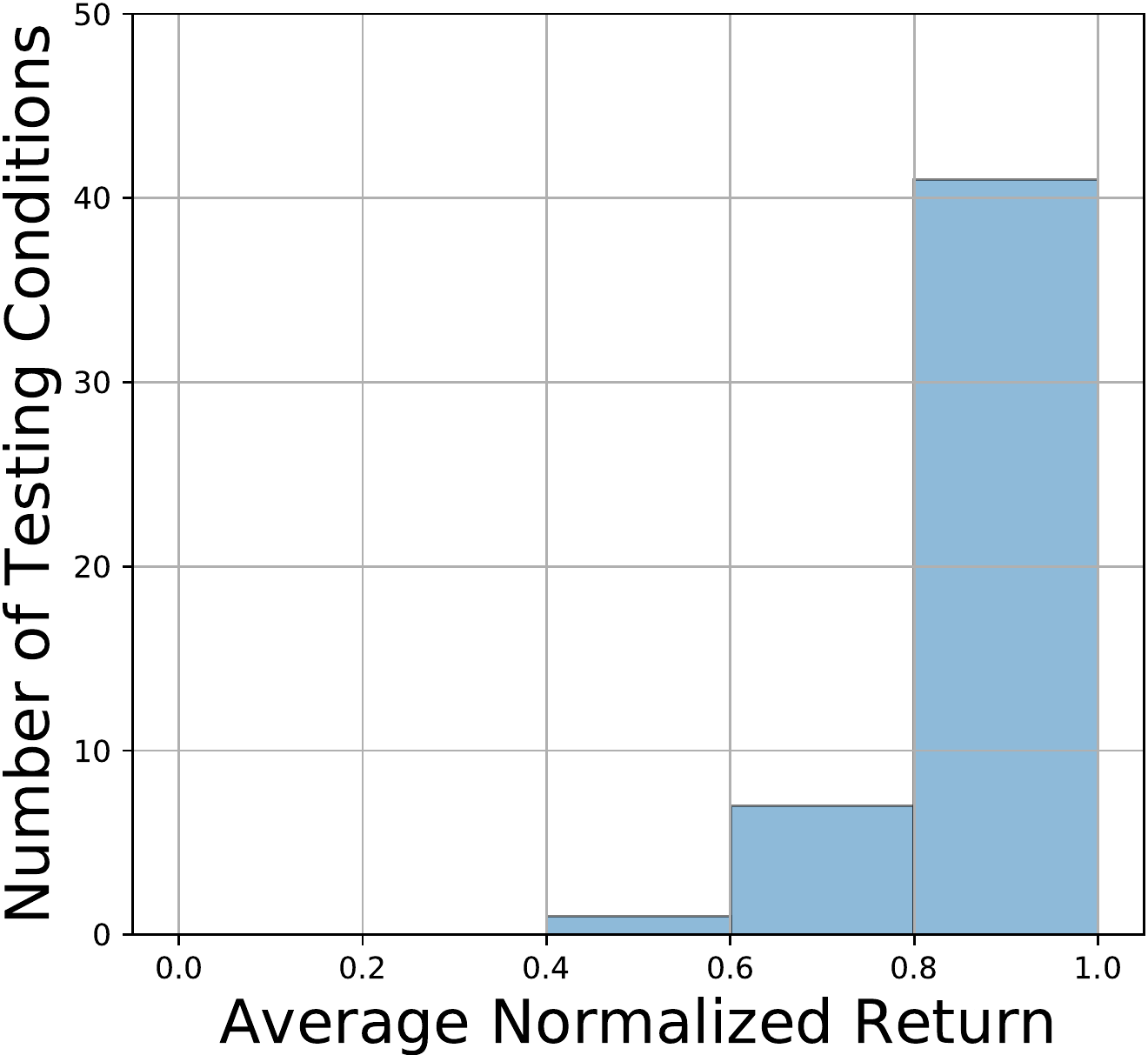}       \label{fig:walker2d_hist_percent_worst}
        }

         \caption {Distribution of average normalized return across 10 seeds with jointly varying test conditions. High reward on the right; low reward on the left. $1^{st}$ row: Hopper, $2^{nd}$ row: Half-Cheetah, $3^{rd}$: Walker2d.} 
        \label{fig:histogram}
    \end{figure*}

\subsection{Robustness to Test Conditions (Environmental Change)}\label{section:test_condition}

We first consider robustness to the conditions of the test environments, such as mass and friction, which are critical parameters for locomotion tasks in the MuJoCo environment. 
During training, all the policies across different methods are trained in the environment with a specific pair of mass and friction values. To evaluate the robustness and generalization of the learned policies, we test the policies in distinct environments with jointly varying mass and friction coefficients. 

As shown in Figure~\ref{fig:heatmap}, our method (ROLAH) has competitive performance (significantly improved performance in Hopper and Half-Cheetah) under varying test conditions. Notably, ROLAH has demonstrated symmetric robustness with respect to varying mass and friction in Hopper task ($1^{st}$ row (a)-(d) in Figure \ref{fig:heatmap}) where we set both the friction and mass coefficients equal to $1.0$ during training. It can observed that the performance of ROLAH is symmetric under decrease/increase of the coefficients centered at $1.0$, the training coefficients. The performance of RAP and other baselines does not demonstrate this trend. Moreover, when tested in environments that gradually shift away from the training environments, the performance drop of ROLAH is less rapid compared to other baselines. This demonstrates the stability and the predictability of ROLAH. 

In Figure~\ref{fig:histogram}, we visualize the distribution of the rewards of various methods in distinct environments. Compared with other baselines, the rewards of ROLAH are more centered in the high reward region and there is no extremely low rewards,  further demonstrating the efficacy of our approach. 

\begin{table*}[th!]
\caption{Performance of ROLAH and baselines under various disturbances.}
\begin{center}
\vspace{-10pt}
        \label{table:task}
    
        \resizebox{14cm}{!}{
        \begin{tabular}{l | l l l l }
            \toprule
            \textbf{Method} & \textbf{Baseline (0 adv)} & \textbf{RARL (1 adv)} & \textbf{RAP (population adv)} & \textbf{ROLAH} \\
            \midrule
            Hopper (No disturbance)        & 0.78$\pm$0.003          & 0.79$\pm$0.02               & 0.84$\pm$0 & \textbf{0.95$\pm$0.01}  \\
            Hopper(Action noise)         & 0.71$\pm$0.001         & 0.74$\pm$0.004      & 0.80$\pm$0     & \textbf{0.91$\pm$0.006}      \\
            Hopper (Worst Adversary)       & 0.42$\pm$0.03           & 0.54$\pm$0.04          & 0.70$\pm$0.007  & \textbf{0.84$\pm$0.14}  \\
            \midrule
            Half-Cheetah (No disturbance)        & 0.77$\pm$0.05          & 0.72$\pm$0.03               & 0.76$\pm$0.02 & \textbf{0.87$\pm$0.05}  \\
            Half-Cheetah(Action noise)         & 0.59$\pm$0.2         & \textbf{0.76$\pm$0.04}      & 0.67$\pm$0.1     & \textbf{0.76$\pm$0.16}      \\
            Half-Cheetah (Worst Adversary)       & 0.16$\pm$0.1            & 0.19$\pm$0.05          & 0.24$\pm$0.36  & \textbf{0.52$\pm$0.21}  \\
            \midrule
            Walker2d (No disturbance)        & \textbf{0.85$\pm$0.27}          & 0.84$\pm$0.43              & 0.43$\pm$0.02 & 0.84$\pm$0.44  \\
            Walker2d (Action noise)         & 0.78$\pm$0.31         & 0.80$\pm$0.28      & 0.36$\pm$0.04     & \textbf{0.83$\pm$0.37}      \\
            Walker2d  (Worst Adversary)       & 0.36$\pm$0.26           & 0.34$\pm$0.12          & 0.34$\pm$0.22  & \textbf{0.68$\pm$0.23}  \\
            \bottomrule
        \end{tabular}}
        
   \end{center}
   
    \end{table*}
    
\subsection{Robustness to Agent Disturbance}
\label{section:disturbance}

In addition to being robust to the internal conditions of the system (e.g., mass and friction), robustness should also be reflected on the external disturbance. To measure robustness to such effect, we report the normalized return of the learned policies in Table \ref{table:task} for 3 types of disturbances during evaluation: (\textit{i}) no disturbance, (\textit{ii}) random adversary that adds noise to the actions of the agents, and (\textit{iii}) the worst adversary that represents the worst case performance of a given policy. To provide such an extreme disturbance in (\textit{iii}), for each policy trained either by a baseline method or ROLAH, we train an adversary to minimize its reward while holding the parameters of that policy as constant, and this process is repeated with $10$ random seeds. In other words, the trained policies undergo disturbances from distinct adversaries, specifically trained to minimize their rewards. 

In Table \ref{table:task}, we first show that learning with adversaries improves the performance compared with the baseline ($1^{st}$ column in Table \ref{table:task}) even though there is no change between training and testing conditions for the baseline, an observation also reported by~\cite{RARL}. We also emphasize that ROLAH outperforms RAP under disturbance, which supports our argument that simply averaging over all the adversaries may decrease robustness. We observe that RARL training with a single adversary is relatively robust to simple action noise while our ROLAH demonstrates its strength in robustness to the learned adversarial policy.

\textbf{Ablation study.} To further understand the strength of ROLAH, we investigate a variation of ROLAH (referred to as \emph{ROLAH-all}) where instead of updating the worst-$k$ adversaries we update all the adversarial policies in the adversary herd. To validate our analysis in Section~\ref{sec:main}, we conduct experiments in Hopper environment and cross-validate the robustness. Empirical evidence demonstrates that ROLAH significantly outperforms ROLAH-all, a result consistent with the less competitive result of RAP in Figures~\ref{fig:heatmap} and~\ref{fig:histogram}. We also found that quite surprisingly ROLAH-all cannot obtain good performance during evaluation even under disturbance from the worst adversary in \emph{its own} group of adversaries during training. Due to space limitation, please refer to Appendix for details.

\section{Related Works}
Recent deep RL advancements, over TD learning~\cite{kostrikov2021offline,kumar2020conservative}, actor-critic~\cite{haarnoja2018soft, lee2020stochastic}, model-based~\cite{hafner2019learning, kaisermodel} and RvS~\cite{chen2021decision,scott2021rvs} methods, have significantly impacted how autonomous agents can facilitate efficient decision making in real-world applications, including healthcare~\cite{gao2022gradient, tang2021model}, robotics~\cite{ibarz2021train, kalashnikov2018scalable}, natural language processing~\cite{ziegler2019fine}, etc. However, the large parameter search space and sample efficiency leave the robustness of RL policies unjustified. Consequentially, there exists a long line of research investigating robust RL~\cite{robust_rl_survey}. 

The robustness problem was initially studied by robust MDPs~\cite{bagnell2001solving, iyengar2005robust, nilim2003robustness} through a model-based manner by assuming the uncertainty set of environmental transitions is known, which can be solved by dynamic programming. Subsequent works generalize the objective to unknown uncertainty sets, and formulate the uncertainty as perturbations/disturbance introduced into, or intrinsically inherited from, the environments, including perturbations in the constants which are used to define environmental dynamics (\textit{e.g.}, gravity, friction, mass)~\cite{
abraham2020model, mankowitz2019robust, mehta2020active,RARL, robust_population, tessler2019action, robust_population, vinitsky2020robust}, disturbance introduced to the observations~\cite{zhang2020robust} and actions~\cite{li2021safe,tessler2019action}. There are meta-RL works that tackle distributional shift across tasks~\cite{lin2020model,zahavy2021discovering}, which are orthogonal to the type of robustness we consider. 

This work is closely related to RARL~\cite{RARL} and RAP~\cite{robust_population}. Specifically, RARL solves the two-player max-min game by leveraging a single adversary,~\textit{i.e.}, following~\eqref{eqn:maxmin}. However, it requires the use of first-order optimization methods and the performance may be degraded due to local optimality. In contrast, our work reformulates the objective such that a fixed set of adversaries considered,~\textit{i.e.},~\eqref{eqn:our_obj}; on the theory side, Theorems~\ref{thm:metric_approx} and~\ref{thm:hp_approx} imply that multiple adversaries can, with high probability, lead to an $\epsilon$-precise approximation under mild assumptions. Moreover, our objective can potentially overcome the over-pessimism of the agent trained by RARL. Different from RAP which also employs a group of adversaries but optimizes the average performance over the adversaries, our method optimizes the average performance over the worst-$k$ adversaries to further improve the robustness. More importantly, we provide theoretical insights that justify the usage of multiple adversaries. 


\section{Conclusion}\label{sec:conclusion}
In this work, we have focused on the task of designing a robust reinforcement learning agent by optimizing its worst-case performance. We have proposed a new algorithm ROLAH that employs an adversarial herd to address two important challenges in solving the two-player max-min game. Our method can efficiently solve the inner optimization problem and is robust to the potential over-pessimism caused by the selection of the candidate adversary set that may include unlikely scenarios. Experimental results on diverse RL environments corroborate that ROLAH can generate policies robust to a variety of environmental disturbance. 

\section*{Acknowledgements}
This work is sponsored in part by the AFOSR under award number FA9550-19-1-0169 and by the NSF National AI Institute for Edge Computing Leveraging Next Generation Wireless Networks, Grant CNS-2112562.

\bibliographystyle{plain}

\newpage
\appendix
\textbf{\huge{Appendix}}

\section{Proofs of Theoretical Results}
\label{app:proofs}

\begin{theorem1}
Consider the metric space $(R_{\Phi}, ||\cdot||_{\infty})$ where for any two functions $R_{\phi}, R_{\phi'} \in R_{\Phi}$, the distance between them is defined as 
$$d(R_{\phi},R_{\phi'}) \doteq ||R_{\phi} - R_{\phi'}||_{\infty}.$$ 
Assume that $R_{\Phi}$ has finite radius under this metric, i.e., 
\begin{equation}
    \sup_{\phi, \phi' \in \Phi}d(R_{\phi},R_{\phi'}) \leq r_{\max},
\end{equation}
where $r_{\max} < \infty$ is a finite number. Let $\widehat\Phi = \{\phi_i\}_{i=1}^m \subset \Phi$. If $R_{\widehat\Phi}$ is a \textbf{maximal} $\epsilon$-packing then $|R_{\widehat \Phi}| \geq \lceil \frac{r_{\max}}{\epsilon} \rceil$, where $\lceil c \rceil$ is the smallest integer that is larger than or equal to $c$,  and $R_{\widehat \Phi}$ is also an $\epsilon$-net. Moreover, for any $\theta \in \Theta$, let $\hat \phi \doteq \argmin_{\phi \in \widehat\Phi}R(\theta,\phi)$ denote the approximated solution and $\phi^* \doteq \argmin_{\phi \in \Phi}R(\theta,\phi)$ denote the optimal solution. Then, the  approximation error of $\hat\phi$ on the inner optimization problem is upper bounded by $\epsilon$, i.e.,
$$
|R(\theta,\phi^*) - R(\theta,\hat\phi)| \leq \epsilon.
$$
\end{theorem1}
\begin{proof}
Since $R_{\widehat\Phi}$ is an $\epsilon$-packing, balls of radius $\frac{\epsilon}{2}$ do not overlap. Consider $\mathcal{U}$ the union of the balls. Any point in $\mathcal{U}$ is clearly within distance $\frac{\epsilon}{2} < \epsilon$ from $R_{\widehat\Phi}$. Consider a point $\phi_* \not \in \mathcal{U}$. If the ball of radius $\frac{\epsilon}{2}$ around $\phi_*$ is disjoint from $\mathcal{U}$, then $R_{\widehat\Phi} \cup \phi_*$ is an $\epsilon$ packing that strictly contains $R_{\widehat\Phi}$. This violates the maximality assumption on  $R_{\widehat\Phi}$. 
Since $R_{\widehat\Phi}$ is an $\epsilon$-packing,then balls of radius $\frac{\epsilon}{2}$ do not overlap. Consider $\mathcal{U}$ the union of the balls. Any point in $\mathcal{U}$ is clearly within distance $\frac{\epsilon}{2} < \epsilon$ from $R_{\widehat\Phi}$. 

Now, consider a point $\phi_* \not \in \mathcal{U}$. If the ball $B(\phi_*, \frac{\epsilon}{2})$ of radius $\frac{\epsilon}{2}$ around $\phi_*$ is disjoint from $\mathcal{U}$ then $R_{\widehat\Phi} \cup \phi_*$ is an $\epsilon$-packing that strictly contains $R_{\widehat\Phi}$. This violates the maximality of $R_{\widehat\Phi}$.  Thus $B(\phi_*, \frac{\epsilon}{2})$ has an intersection with at least a ball of radius $\frac{\epsilon}{2}$ around some point of $R_{\widehat\Phi}$. It follows from triangle inequality that $\phi_*$ is within distance $\epsilon$ of this point. Since $\phi_*$ was arbitrary, then $R_{\widehat\Phi}$ is an $\epsilon$-covering and an $\epsilon$-net. The fact that $|R_{\widehat \Phi}| \geq \lceil \frac{r_{\max}}{\epsilon} \rceil$ follows trivially from the fact that balls of radius $\frac{\epsilon}{2}$ around the points of $R_{\widehat\Phi}$ do not intersect and the triangle inequality.

Since $R_{\widehat\Phi}$ is an $\epsilon$-net of $R_\Phi$,  for any $\phi^*$ there exists $\phi \in \widehat\Phi$ such that $||R_\phi-R_{\phi^*}||_{\infty} \leq \epsilon$. By definition of the $L_{\infty}$ norm, this implies that for any $\theta \in \Theta$,  $|R(\theta,\phi^*) - R(\theta,\phi)| \leq \epsilon$. Also because $\hat \phi \doteq \argmin_{\phi \in \widehat\Phi}R(\theta,\phi)$, we have $R(\theta,\hat\phi) \leq R(\theta,\phi)$. Since $\phi^*$ is defined as $\argmin_{\phi \in \Phi}R(\theta,\phi)$, it holds that 
\begin{align*}
    |R(\theta,\phi^*) - R(\theta,\hat\phi)| &= R(\theta,\hat\phi) - R(\theta,\phi^*) \\
    &\leq R(\theta,\phi) - R(\theta,\phi^*) \\
    &= |R(\theta,\phi^*)-R(\theta,\phi)| \leq \epsilon,
\end{align*}
completing the proof.
\end{proof}

\begin{theorem2}
Assume that $\Phi$ is a metric space with a distance function $d: \Phi \times \Phi \mapsto \mathbb{R}$. Let $\sigma$ be any probability measure on $\Phi$. Let $\widehat\Phi = \{\phi_i\}_{i=1}^m$ be a set of independently sampled elements from $\Phi$ following identical measure $\sigma$. Consider a fixed $\theta \in \Theta$ and assume that $R(\theta,\phi)$ is an $L_{\phi}$-Lipschitz continuous function of $\phi$ with respect to the metric space $(\Phi, d)$. Let $\widehat\phi$ and $\phi^*$ be defined the same as in Theorem~\ref{thm:metric_approx}. For presentation simplicity, assume that $\sigma(\{\phi: d(\phi,\phi^*)\leq\epsilon\}) \geq L_{\sigma}\epsilon$. Let $0 < \delta < 1$ denote the probability of a bad event. Then with probability $1-\delta$, the  approximation error of $\hat\phi$ on the inner optimization problem is upper bounded by $\epsilon$ if 
$m \geq \log(\delta)\log^{-1}(1-\frac{L_{\sigma}}{L_{\phi}}\epsilon)$.
\end{theorem2}

\begin{proof}
Assume that we have $\widehat\Phi = \{\phi_i\}_{i=1}^m$ as a batch of independently sampled elements from $\Phi$, all following the measure of $\sigma$ during sampling. For any $c > 0$, we have that
\begin{align}
    &\mathbb{P}(\exists \phi \in \widehat\Phi \quad s.t. \quad d(\phi,\phi^*) \leq c) \nonumber \\ 
    &= 1- \mathbb{P}(\forall \phi \in \widehat\Phi: d(\phi,\phi^*)>c) \nonumber \\
    &= 1- \mathbb{P}^m(\phi: d(\phi,\phi^*)>c) \nonumber \\
    &= 1- (1-\sigma(\{\phi: d(\phi,\phi^*)\leq c\}))^m. \label{eqn:prob_1}
\end{align}

On the other hand, if there exists $\phi \in \widehat\Phi$ such that $d(\phi,\phi^*) \leq c$, then by the assumption that $R_\phi$ is $L_\phi$-Lipschitz continuous, $|R(\theta, \phi)-R(\theta,\phi^*)| \leq L_\phi\cdot c$. By definition of $\widehat\Phi$, it holds that
\begin{align*}
    & |R(\theta, \widehat\phi)-R(\theta,\phi^*)| = R(\theta, \widehat\phi) - R(\theta,\phi^*) \\
    & \leq R(\theta, \phi) - R(\theta,\phi^*) = |R(\theta, \phi) - R(\theta,\phi^*)| \\
    & \leq L_\phi\cdot c.
\end{align*}

To prove the theorem, let $c=\frac{\epsilon}{L_\phi}$, and we want 
\begin{align}
&1-\delta \leq \mathbb{P}(\exists \phi \in \widehat\Phi \quad s.t. \quad d(\phi,\phi^*) \leq c) \nonumber \\
&1-\delta \leq 1- (1-\sigma(\{\phi: d(\phi,\phi^*)\leq c\}))^m \label{eqn:prob_2} \\
&(1-\sigma(\{\phi: d(\phi,\phi^*)\leq c\}))^m \leq \delta \nonumber \\
&m \leq \frac{\log(\delta)}{\log(1-\sigma(\{\phi: d(\phi,\phi^*)\leq c)} \nonumber \\
&m \leq \frac{\log(\delta)}{\log(1-\frac{L_\sigma}{L_\phi}\epsilon)}
\label{eqn:prob_3} \\
&m \leq \log(\delta)\log^{-1}(1-\frac{L_\sigma}{L_\phi}\epsilon) \nonumber
\end{align}
where in Eq.~\eqref{eqn:prob_2} we use Eq.~\eqref{eqn:prob_1} and in~\eqref{eqn:prob_3} we use the fact that $c=\frac{\epsilon}{L_\phi}$ and the density assumption that $\sigma(\{\phi: d(\phi,\phi^*)\leq\epsilon\}) \geq L_{\sigma}\epsilon$. This concludes the proof. 
\end{proof}

\begin{lemma1}
The solution set to the optimization problem in~\eqref{eqn:maxmin} is identical to the solution set of the optimization problem in~\eqref{eqn:rah}.
That is, for any $\theta \in \Theta$ and integer $m \geq 1$,
$$
\min_{\phi \in \Phi} R(\theta,\phi) = \min_{\phi_1,\dots,\phi_m \in \Phi}\min_{\phi \in \{\phi_i\}_{i=1}^m} R(\theta,\phi). 
$$
\end{lemma1}
\begin{proof}
There are only $3$ possibilities regarding the order of $\min_{\phi \in \Phi} R(\theta,\phi)$ and $\min_{\phi_1,\dots,\phi_m \in \Phi}\min_{\phi \in \{\phi_i\}_{i=1}^m} R(\theta,\phi)$: 
\begin{itemize}
    \item \textit{(i)} $\min_{\phi \in \Phi} R(\theta,\phi)=\min_{\phi_1,\dots,\phi_m \in \Phi}\min_{\phi \in \{\phi_i\}_{i=1}^m} R(\theta,\phi)$;
    \item  \textit{(ii)} $\min_{\phi \in \Phi} R(\theta,\phi)>\min_{\phi_1,\dots,\phi_m \in \Phi}\min_{\phi \in \{\phi_i\}_{i=1}^m} R(\theta,\phi)$;
    \item  \textit{(iii)} $\min_{\phi \in \Phi} R(\theta,\phi)<\min_{\phi_1,\dots,\phi_m \in \Phi}\min_{\phi \in \{\phi_i\}_{i=1}^m} R(\theta,\phi)$. 
\end{itemize}
We prove by contradiction that \textit{(ii)} and \textit{(iii)} are impossible to happen. 

If \textit{(ii)} holds, let $\widehat\Phi^*$ denote the optimal solution to the right hand side (RHS) and let $\hat\phi \doteq \min_{\phi \in \widehat\Phi^*}R(\theta, \phi)$. Because \textit{(ii)} holds, we have that $\min_{\phi \in \Phi}R(\theta,\phi) > R(\theta,\hat\phi)$. However, this is impossible because $\hat\phi \in \Phi$ by definition of $\widehat\Phi$. 

If \textit{(iii)} holds, let $\phi^* \doteq \min_{\phi \in \Phi} R(\theta,\phi)$ be the optimal solution of the left hand side (LHS). Consider any $\widehat\Phi$ that includes $\phi^*$, then $\min_{\phi \in \Phi} R(\theta,\phi) = R(\theta,\phi^*) \geq \min_{\phi \in \widehat\Phi}R(\theta,\phi) \geq \min_{\phi_1,\dots,\phi_m \in \Phi}\min_{\phi \in \{\phi_i\}_{i=1}^m} R(\theta,\phi)$. This is contradicting to the fact that \textit{(iii)} holds. 
Hence, the Lemma is proved. 
\end{proof}

\section{Ablation Studies for ROLAH with Different Update Methods }
To validate our theory in Section~\ref{sec:main}, we conduct extra experiments in the Hopper environment. We investigate two versions of ROLAH that updates the adversarial head with different schemes: in each iteration during training, (\textit{i}) ROLAH-worst: only update the worst-$k$ adversaries, where the worst-$k$ adversaries are defined as in Section \ref{sec:worst-k}, and (\textit{ii}) ROLAH-all: update the whole population of adversaries.

 \begin{figure*}
        \centering
        
        \subfigure[ROLAH-all]{  
            \centering 
    \includegraphics[width=0.28\textwidth]{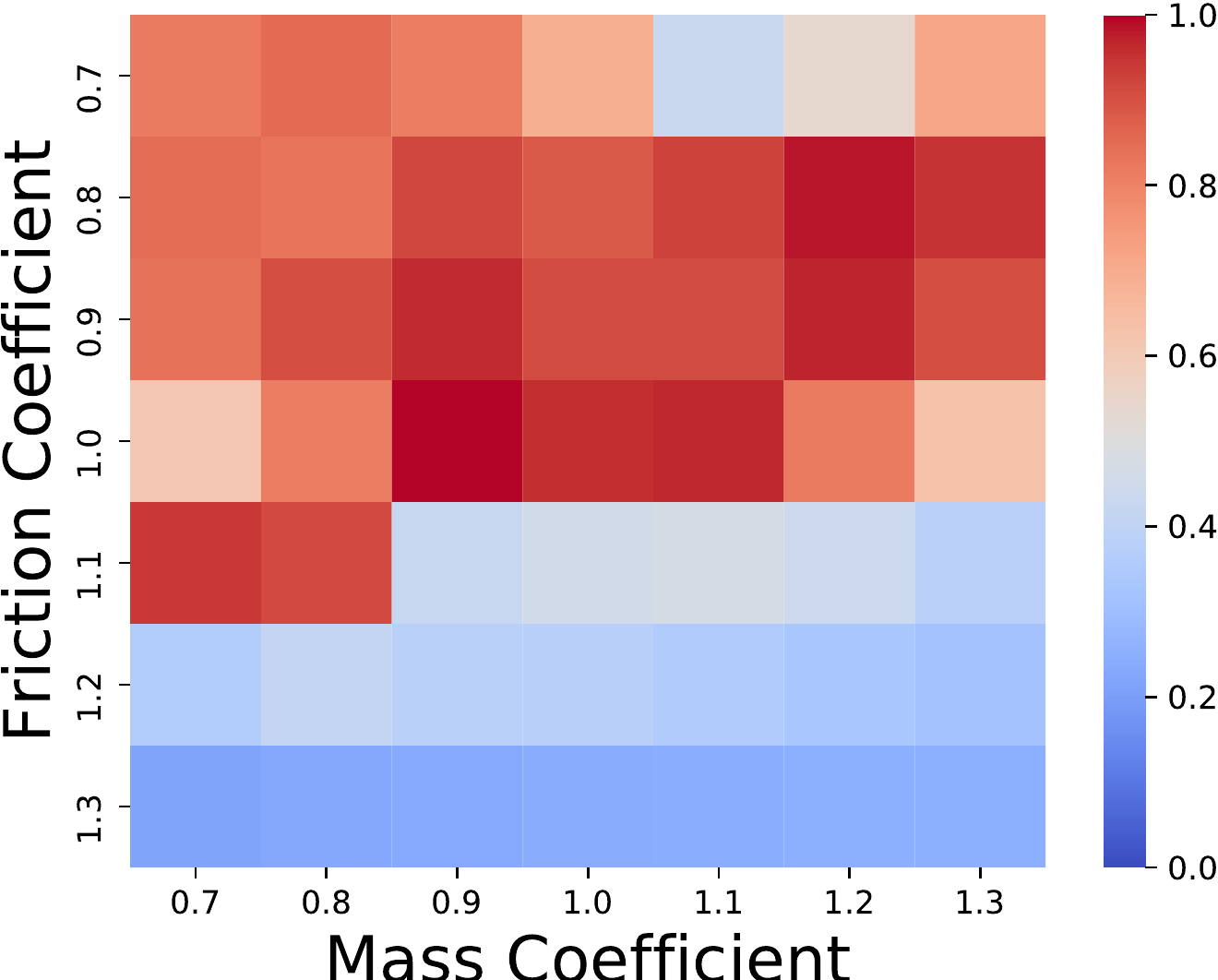}\label{fig:hopper_heatmap_all}
        }
        \subfigure[ROLAH-worst]{ 
            \centering 
       \includegraphics[width=0.28\textwidth]{figure/hopper_heatmap_percent_worst.pdf}       \label{fig:hopper_heatmap_percent_worst_appendix}
        }

         \caption {Average normalized return across 10 seeds tested via different mass coefficients on the x-axis and friction coefficients on the y-axis for two variations of ROLAH in the Hopper environment. High reward has red color; low reward has blue color.} 
        \label{fig:heatmap_different_update}
    \end{figure*}

\begin{table*}[th!]
\caption{Performance of ROLAH and baselines under various disturbances in \textbf{Hopper environment.}}
\begin{center}

\vspace{0.1cm}
        \label{table:task_ROLAH_update}
    
        \resizebox{10cm}{!}{
        \begin{tabular}{l | l l l l }
            \toprule
            \textbf{Method} & \textbf{ROLAH-all} & \textbf{ROLAH-worst} \\
            \midrule
            Hopper (No disturbance)        &0.86$\pm$0.07 & \textbf{0.95$\pm$0.01}  \\
            Hopper(Action noise)         & 0.81$\pm$0.01     & \textbf{0.91$\pm$0.006}      \\
            Hopper (Worst Adversary)       & 0.63$\pm$0.22  & \textbf{0.84$\pm$0.14}  \\
        
            \bottomrule
        \end{tabular}}
        
   \end{center}
   
    \end{table*}

\begin{figure}
     \subfigure[Adversarial policy from ROLAH-all]{       
         \includegraphics[width=0.475\linewidth]{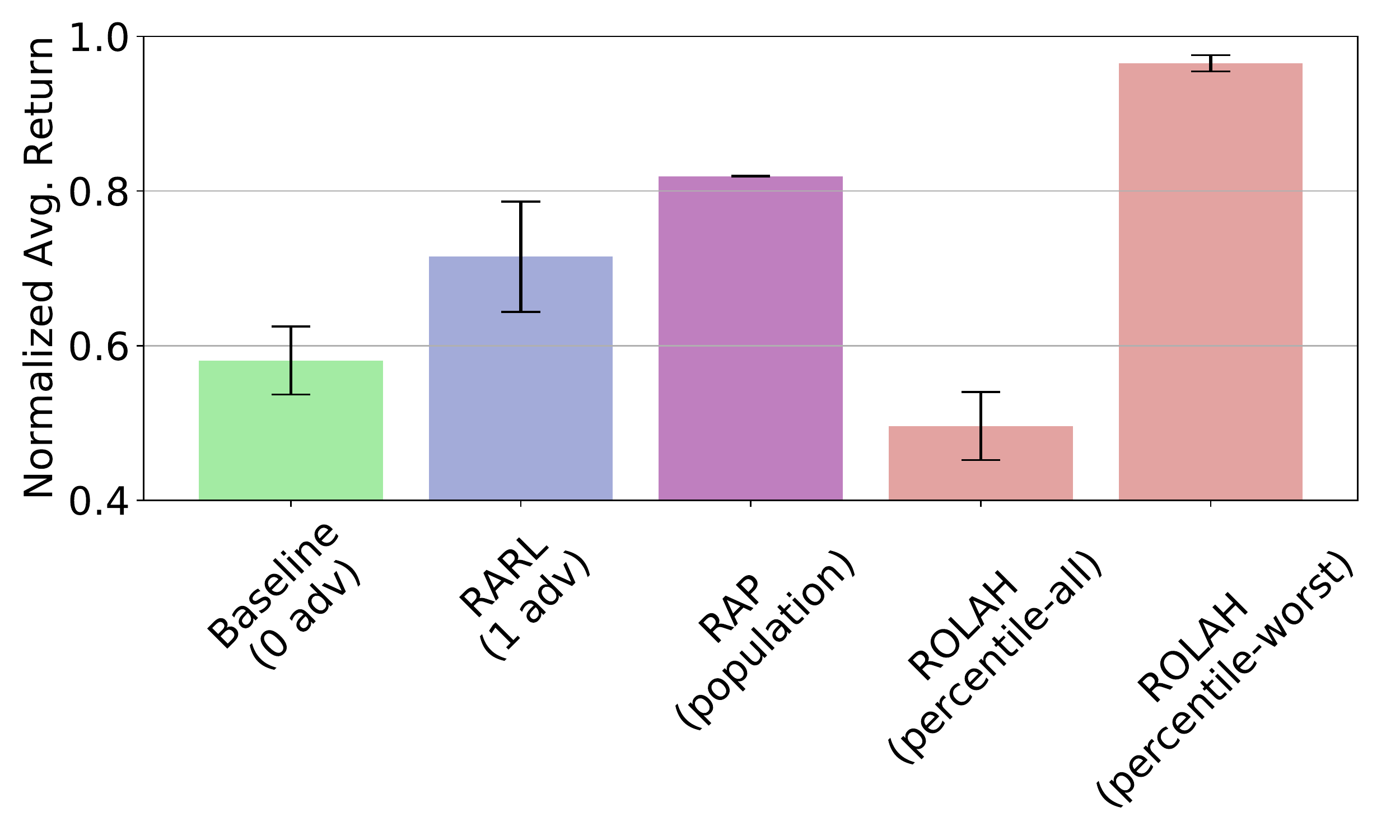}
         \label{fig:hopper_adversary_all}
     }
     \subfigure[Adversarial policy from ROLAH-worst]{
         \includegraphics[width=0.475\linewidth]{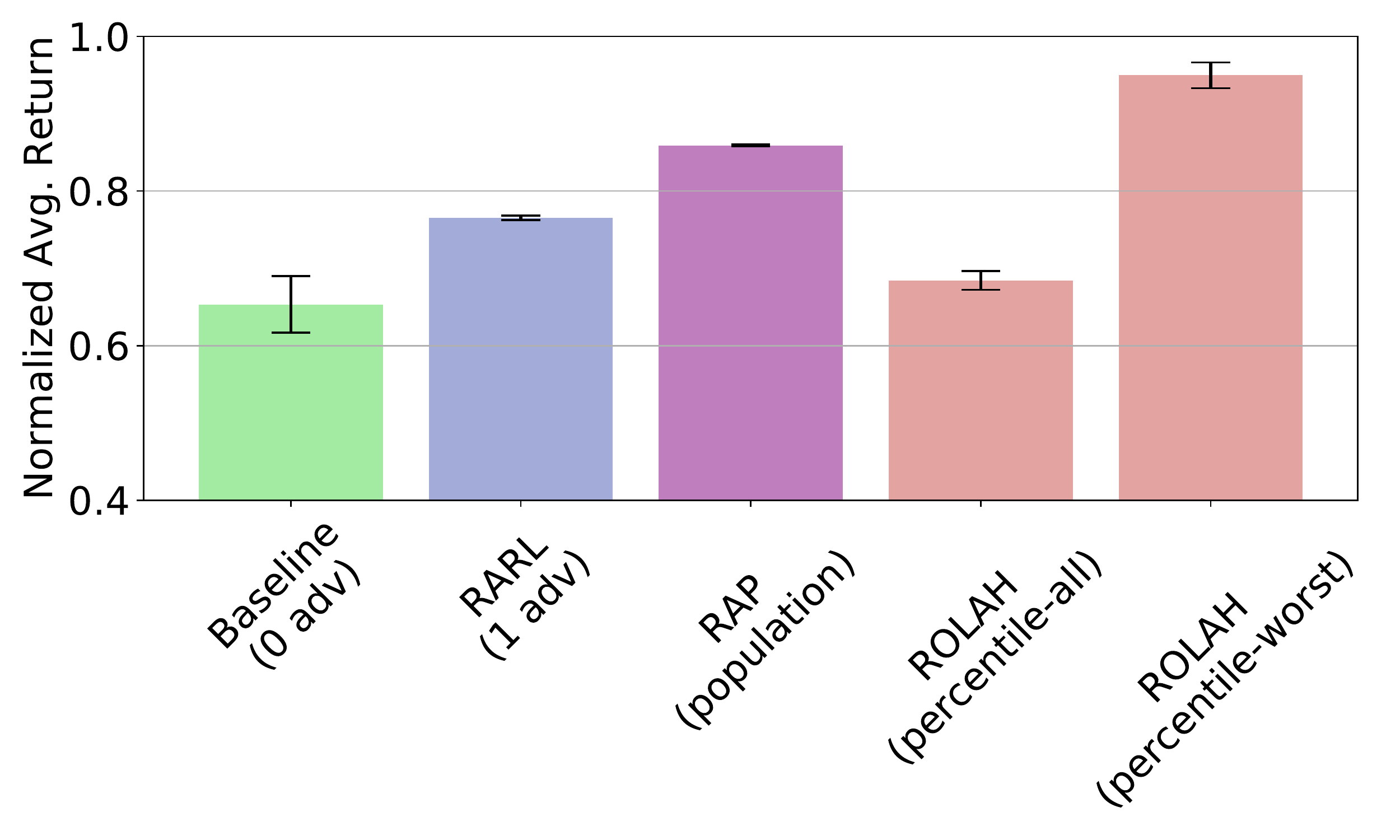}
         \label{fig:hopper_adversary_worst}
     }
     
     \caption {Average normalized return for Hopper task cross-tested with the worst adversary from (1) ROLAH-all and (2) ROLAH-worst.} 
        \label{fig:hopper_ablation}
 
\end{figure}

\subsection{Robustness to Test Conditions (Environmental Changes)} \label{section:appendix_test_conditions}
We follow the same evaluation metrics as we demonstrate in Section \ref{section:test_condition}, considering training with a fixed pair of mass and friction values while evaluating the trained policies with varying mass and friction coefficients. We show that the ability to generalization is better with only updating the worst$-k$ adversaries during training in Figure~\ref{fig:heatmap_different_update}.

\subsection{Robustness to Disturbance on the Agent}
We report the normalized return of ROLAH with different update methods as the discussion in Section \ref{section:disturbance} in Table \ref{table:task_ROLAH_update} for 3 types of disturbances during evaluation: (\textit{i}) no disturbance, (\textit{ii}) random adversary that adds noise to the actions of the agents, and (\textit{iii}) the worst adversary that represents the worst case performance of a given policy. The relative performance is also consistent with Appendix \ref{section:appendix_test_conditions} that only updates the worst$-k$ adversaries leading to more robustness to disturbance.

\subsection{Cross-Validation of ROLAH}
After the training process of ROLAH is finished, we have access to a trained agent and a group of trained adversarial policies. To evaluate the effectiveness of training, we evaluate all the baseline methods and ROLAH-worst/all under the disturbance from two adversaries: \textit{(i)} the worst adversary in the trained adversarial herd of ROLAH-worst and \textit{(ii)} the worst adversary in the trained adversarial herd of ROLAH-all. The selection of the worst adversary follows the the same process as described in Section~\ref{section:disturbance}. As can be seen in Figure~\ref{fig:hopper_ablation},  ROLAH-all cannot survive from its own adversary, i.e., the adversary that it has encountered during training. 

 \begin{figure*}
        \centering
        \subfigure[Baseline (0 adv)]{
            \centering
    \includegraphics[width=0.23\textwidth]{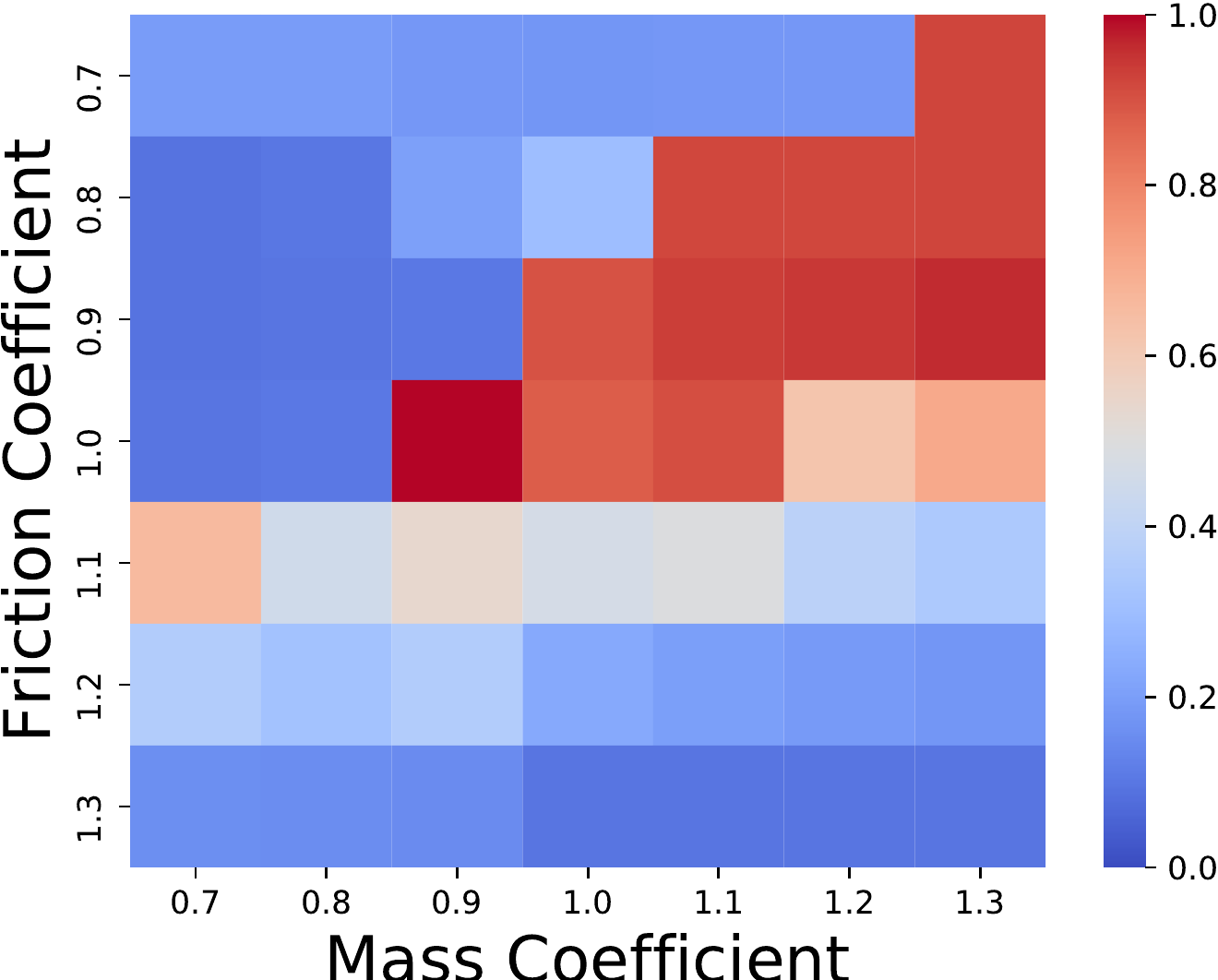}   
        \label{fig:ppo_hopper_heatmap_zero}}
        \vspace{-16pt}
        \subfigure[RARL (1 adv)]{  
            \centering \includegraphics[width=0.23\textwidth]{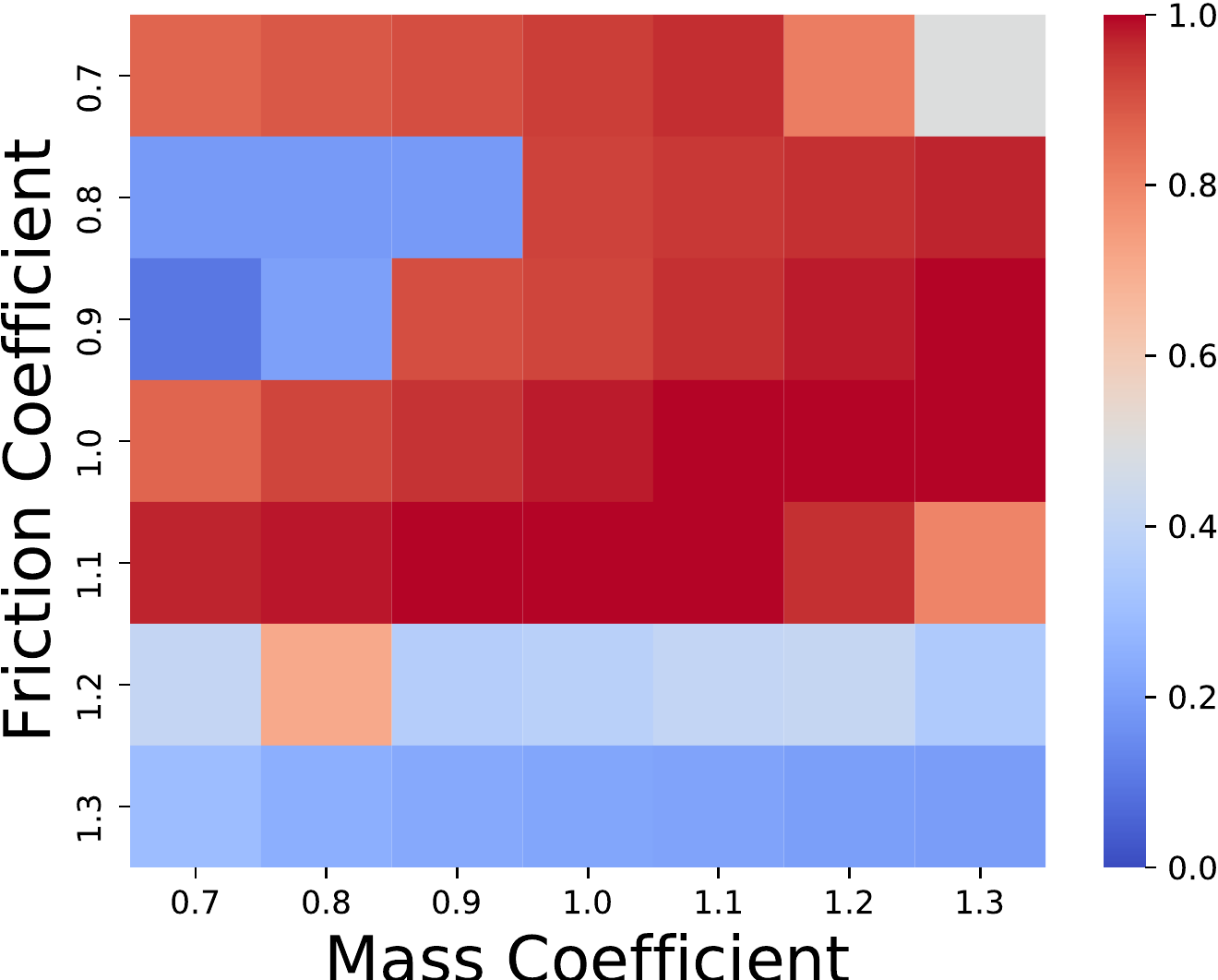}  
    \label{fig:ppo_hopper_heatmap_single}}
        \subfigure[RAP (population)]{  
            \centering 
    \includegraphics[width=0.23\textwidth]{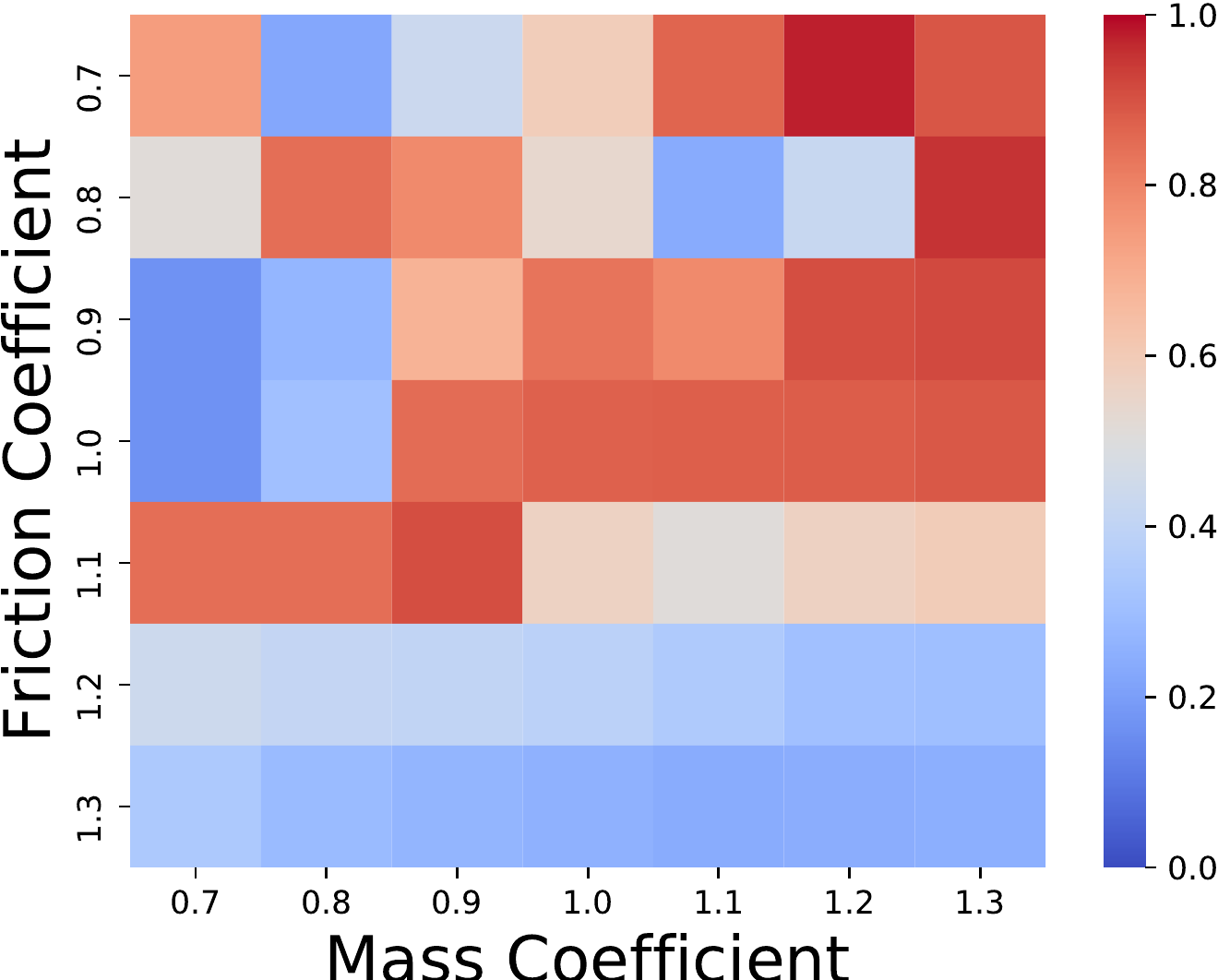}\label{fig:ppo_hopper_heatmap_popu}
        }
        \subfigure[ROLAH]{ 
            \centering 
       \includegraphics[width=0.23\textwidth]{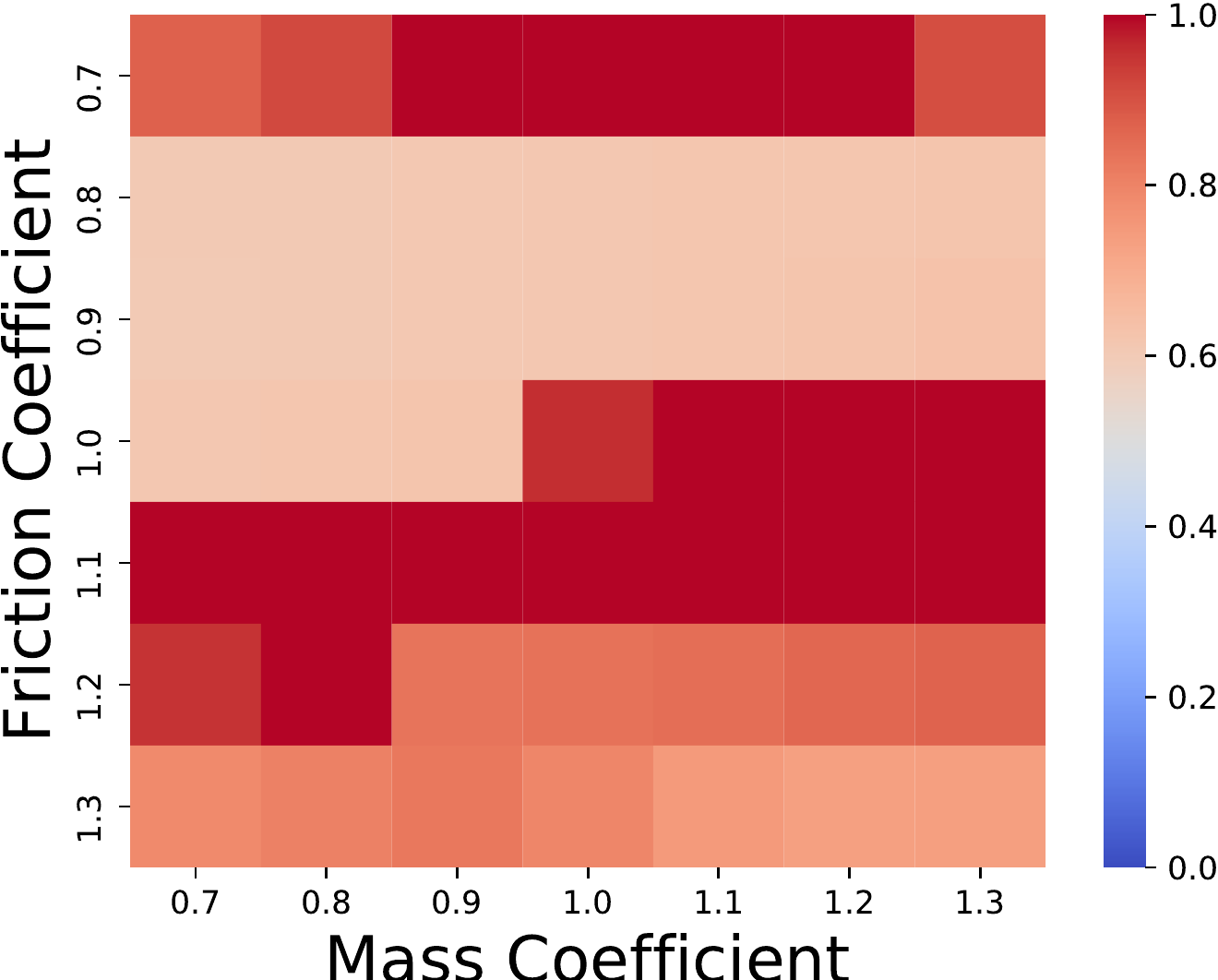}       \label{fig:ppo_hopper_heatmap_percent_worst}
        }
    \vskip\baselineskip
     \centering
        \subfigure[Baseline (0 adv)]{
            \centering
    \includegraphics[width=0.23\textwidth]{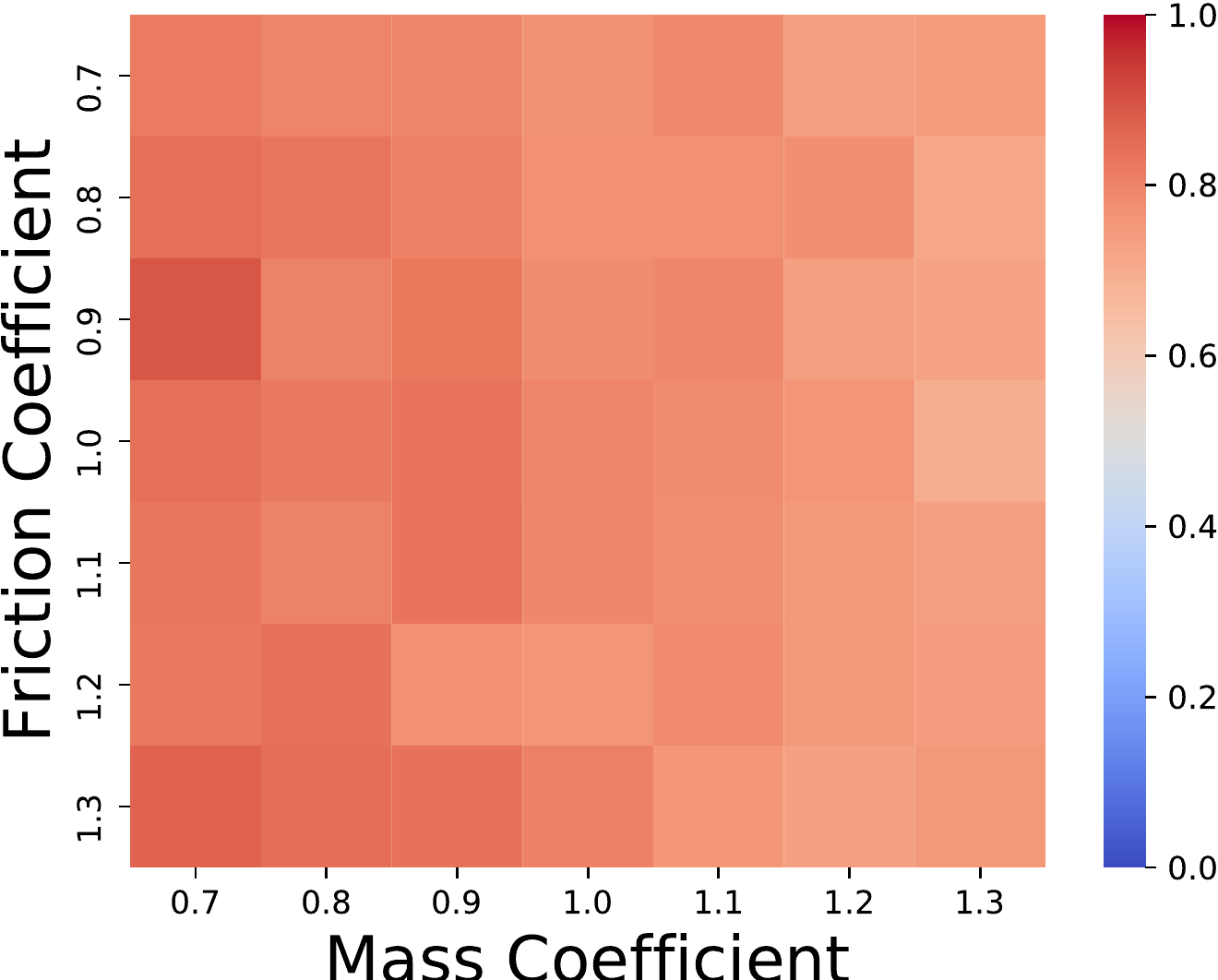}   
        \label{fig:ppo_halfcheetah_heatmap_zero}}
        \vspace{-16pt}
        \subfigure[RARL (1 adv)]{  
            \centering \includegraphics[width=0.23\textwidth]{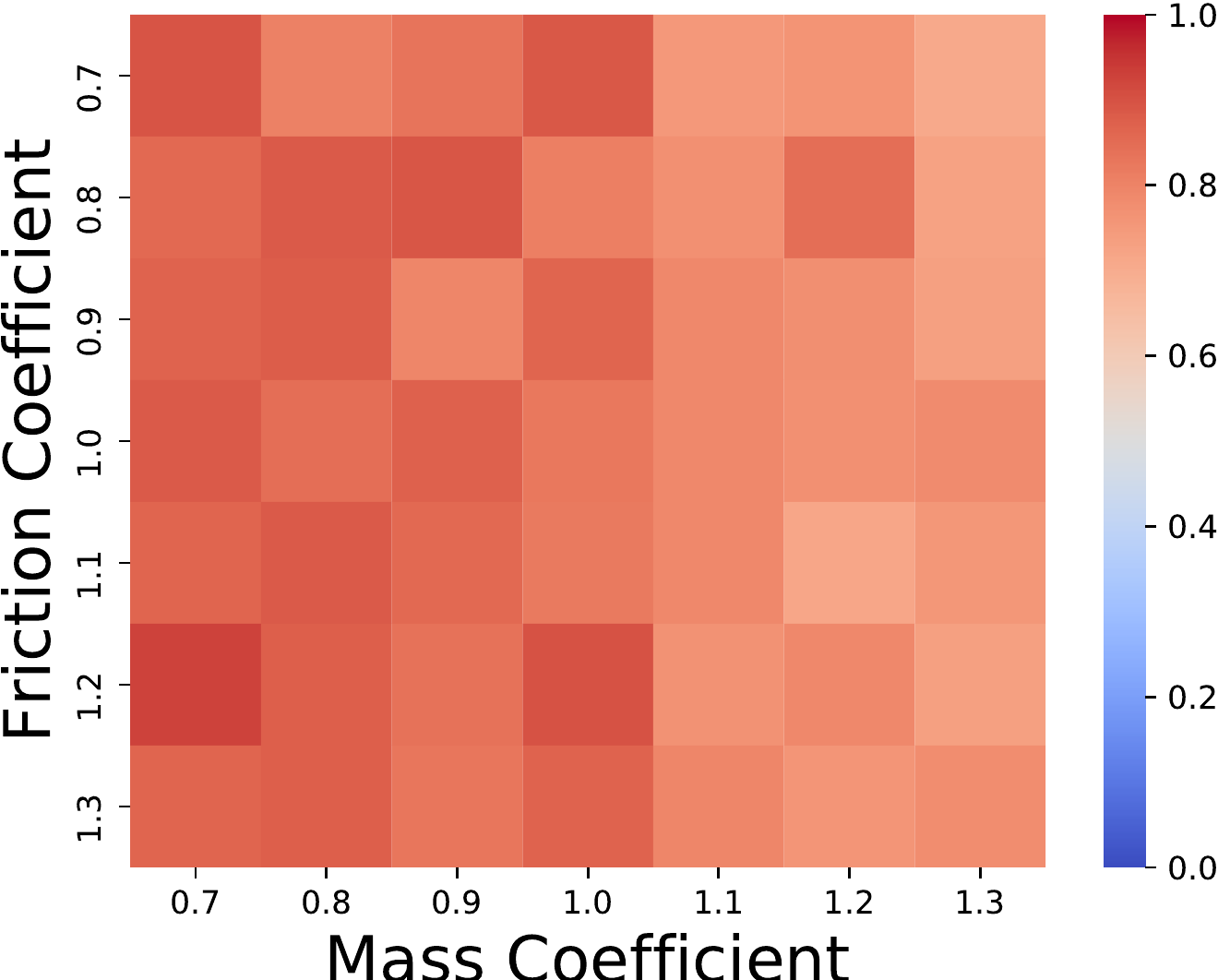}  
    \label{fig:ppo_halfcheetah_heatmap_single}}
        \subfigure[RAP (population)]{  
            \centering 
    \includegraphics[width=0.23\textwidth]{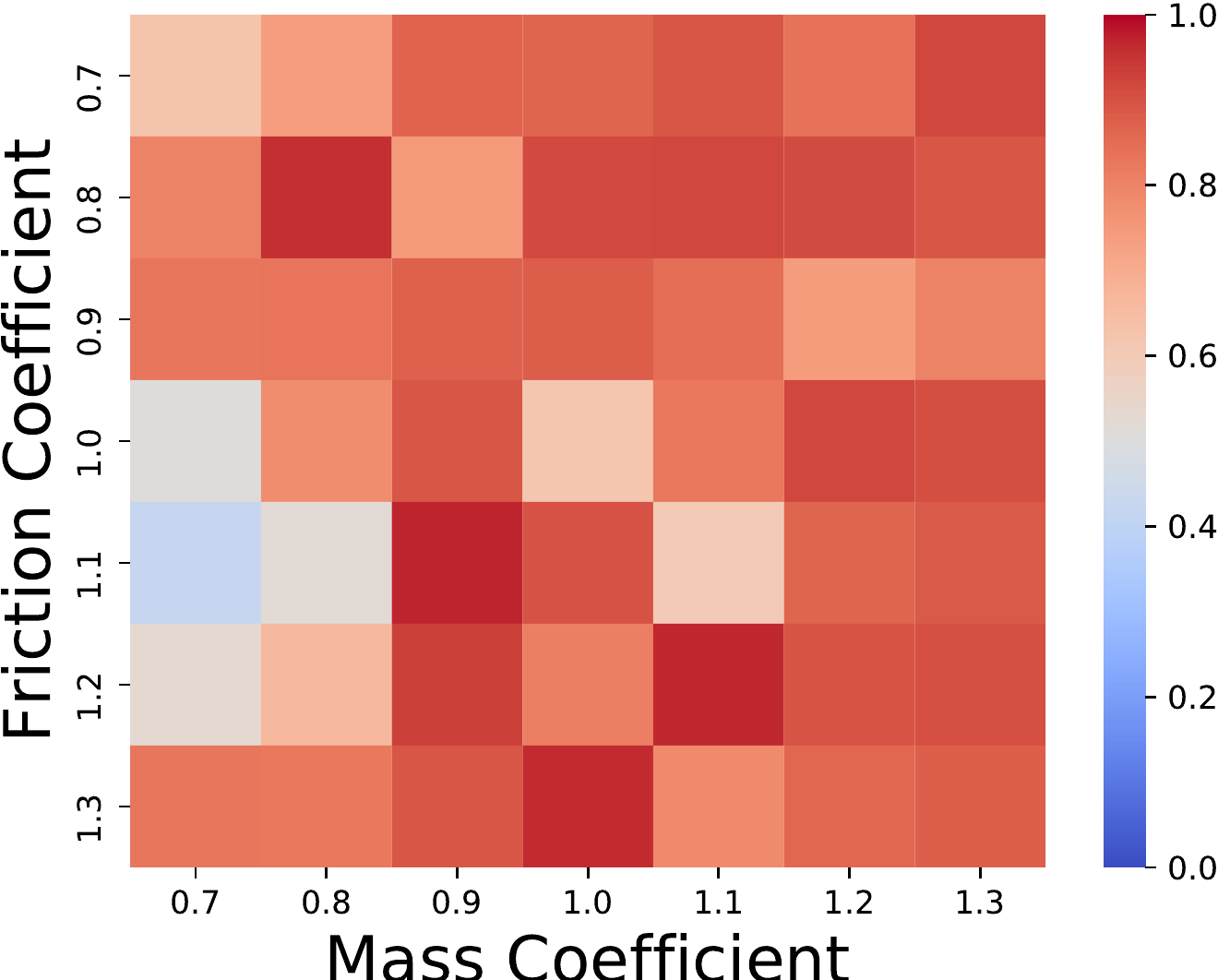}\label{fig:ppo_halfcheetah_heatmap_popu}
        }
        \subfigure[ROLAH]{ 
            \centering 
       \includegraphics[width=0.23\textwidth]{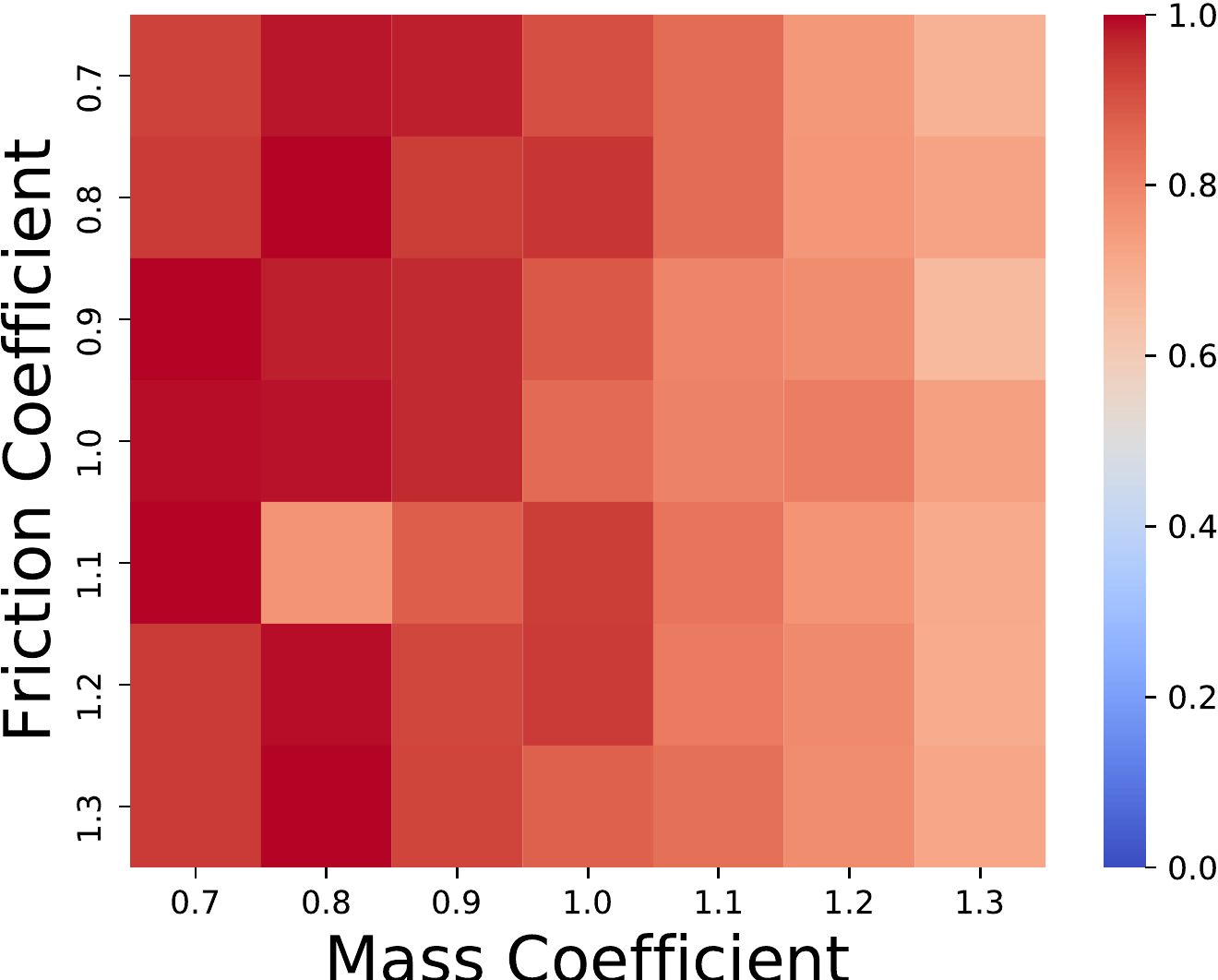}       \label{fig:ppo_halfcheetah_heatmap_percent_worst}
        }
         \vskip\baselineskip
     \centering
        \subfigure[Baseline (0 adv)]{
            \centering
    \includegraphics[width=0.23\textwidth]{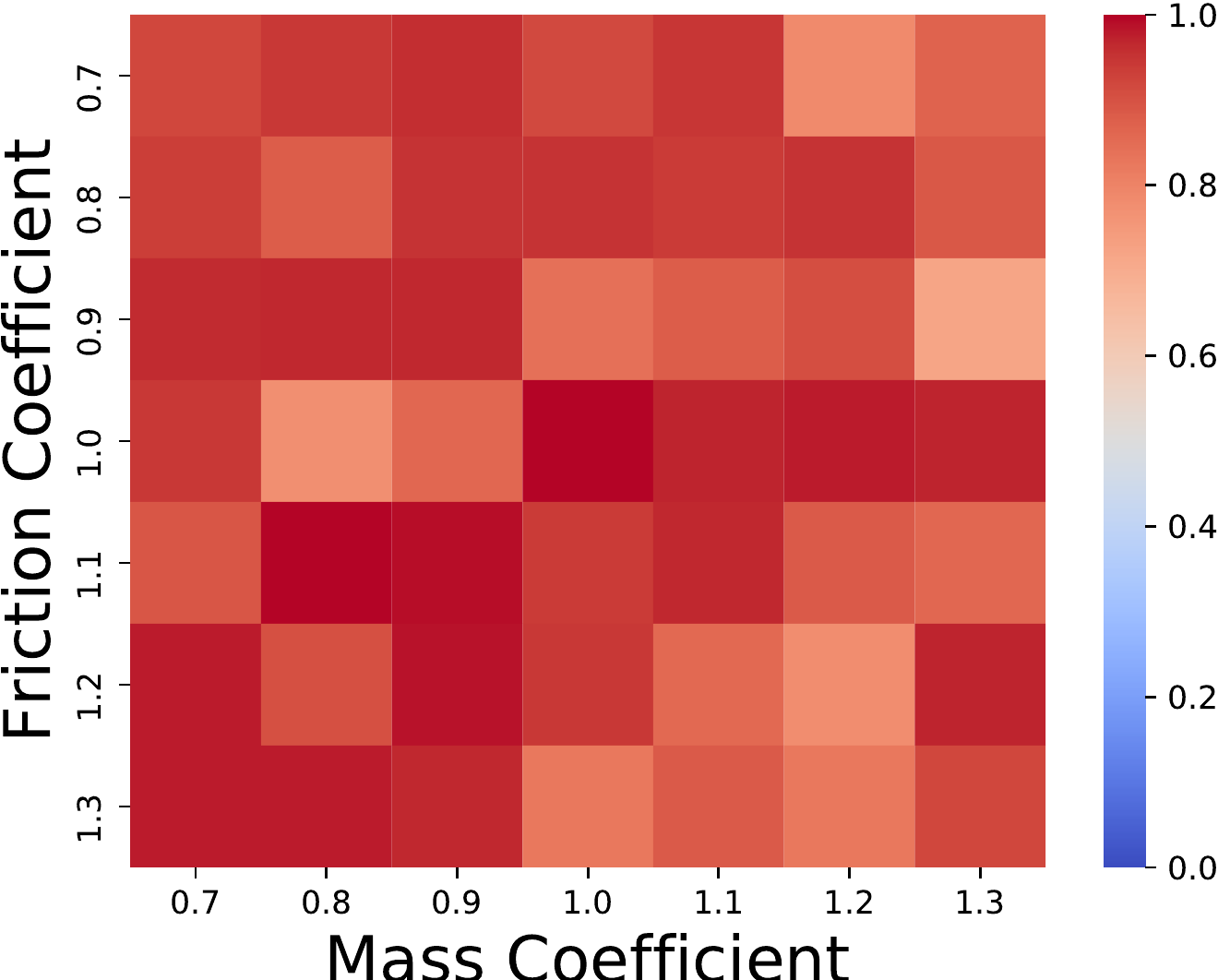}   
        \label{fig:ppo_walker2d_heatmap_zero}}
        \subfigure[RARL (1 adv)]{  
            \centering \includegraphics[width=0.23\textwidth]{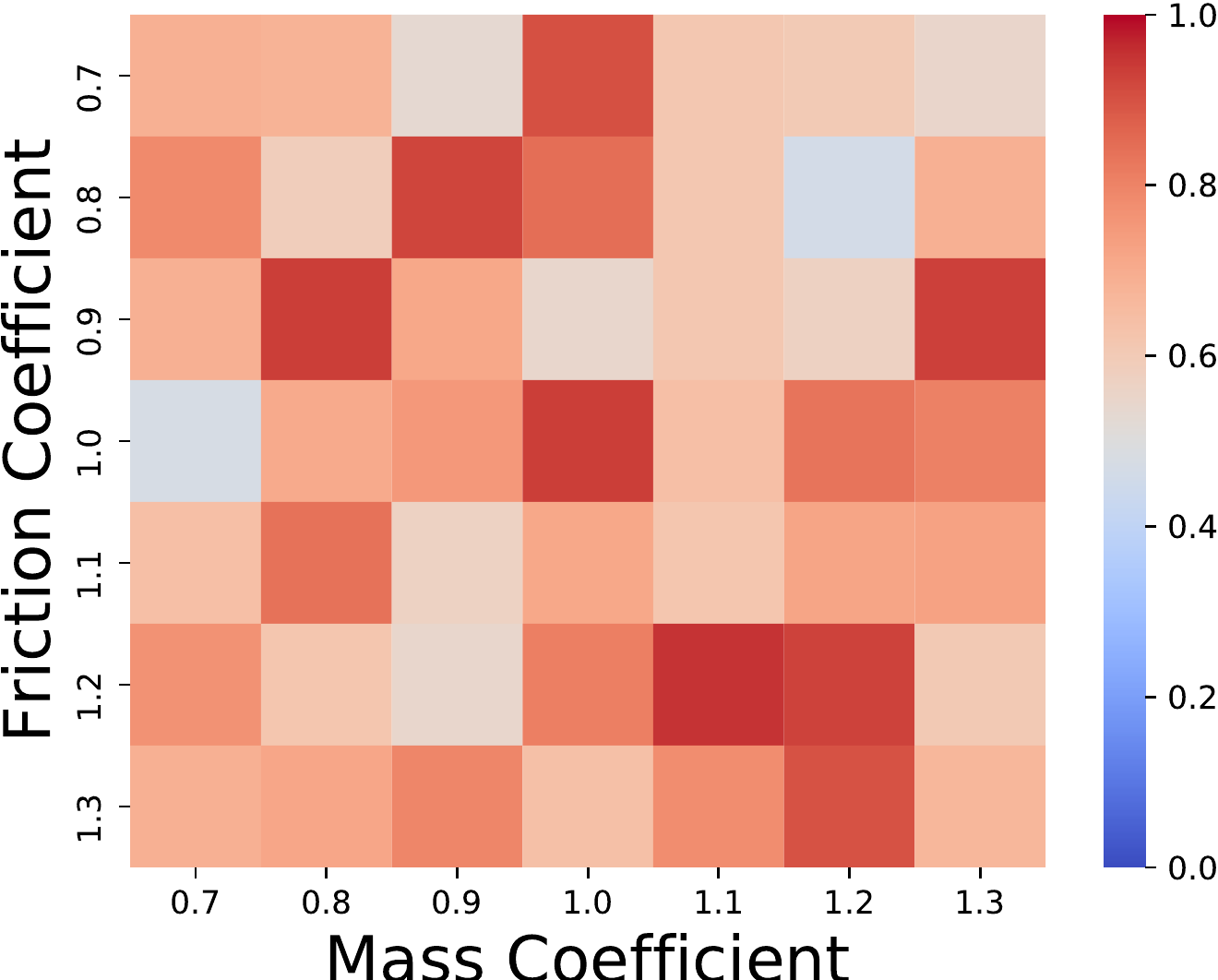}  
    \label{fig:ppo_walker2d_heatmap_single}}
        \subfigure[RAP (population)]{  
            \centering 
    \includegraphics[width=0.23\textwidth]{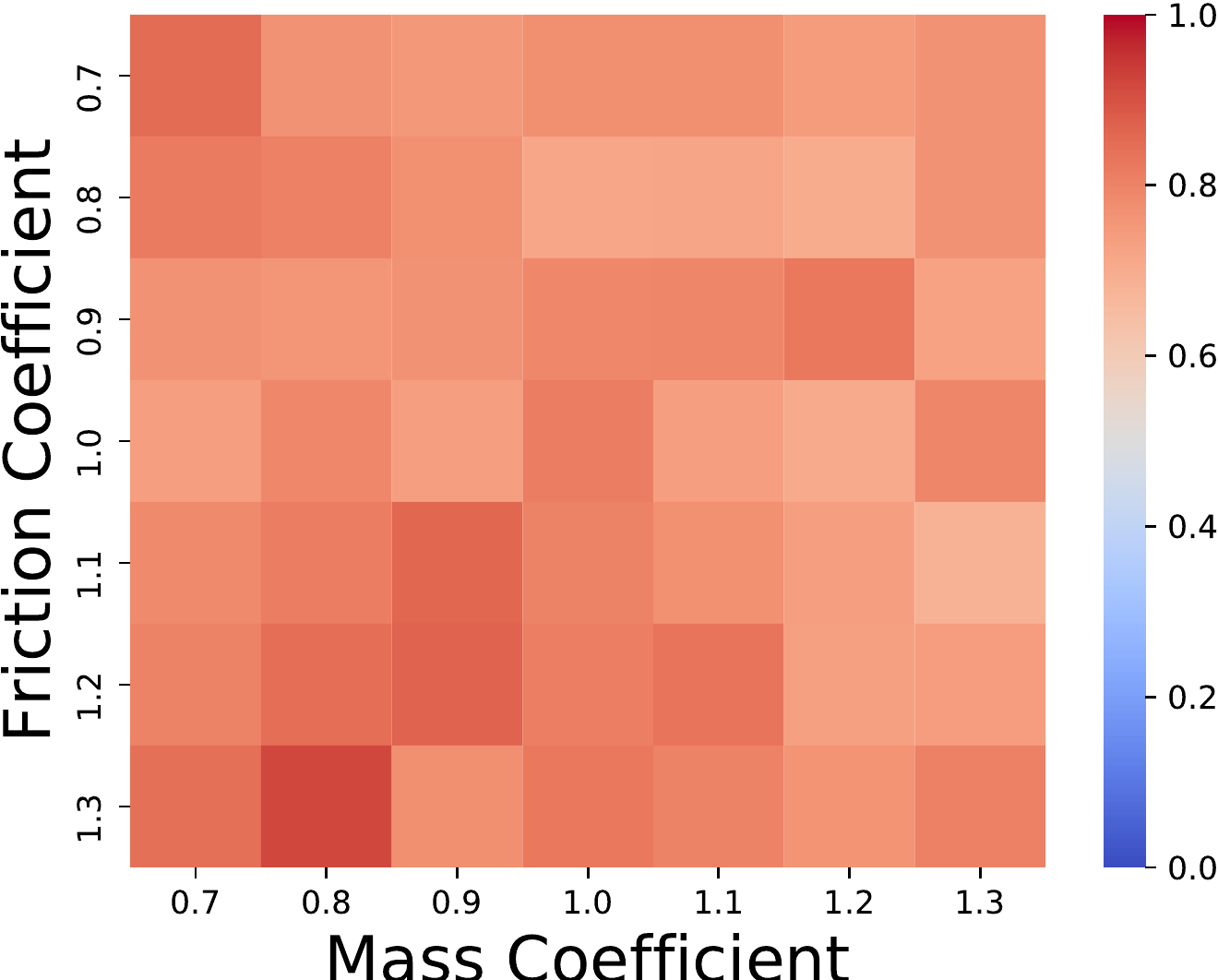}\label{fig:ppo_walker2d_heatmap_popu}
        }
        \subfigure[ROLAH]{ 
            \centering 
       \includegraphics[width=0.23\textwidth]{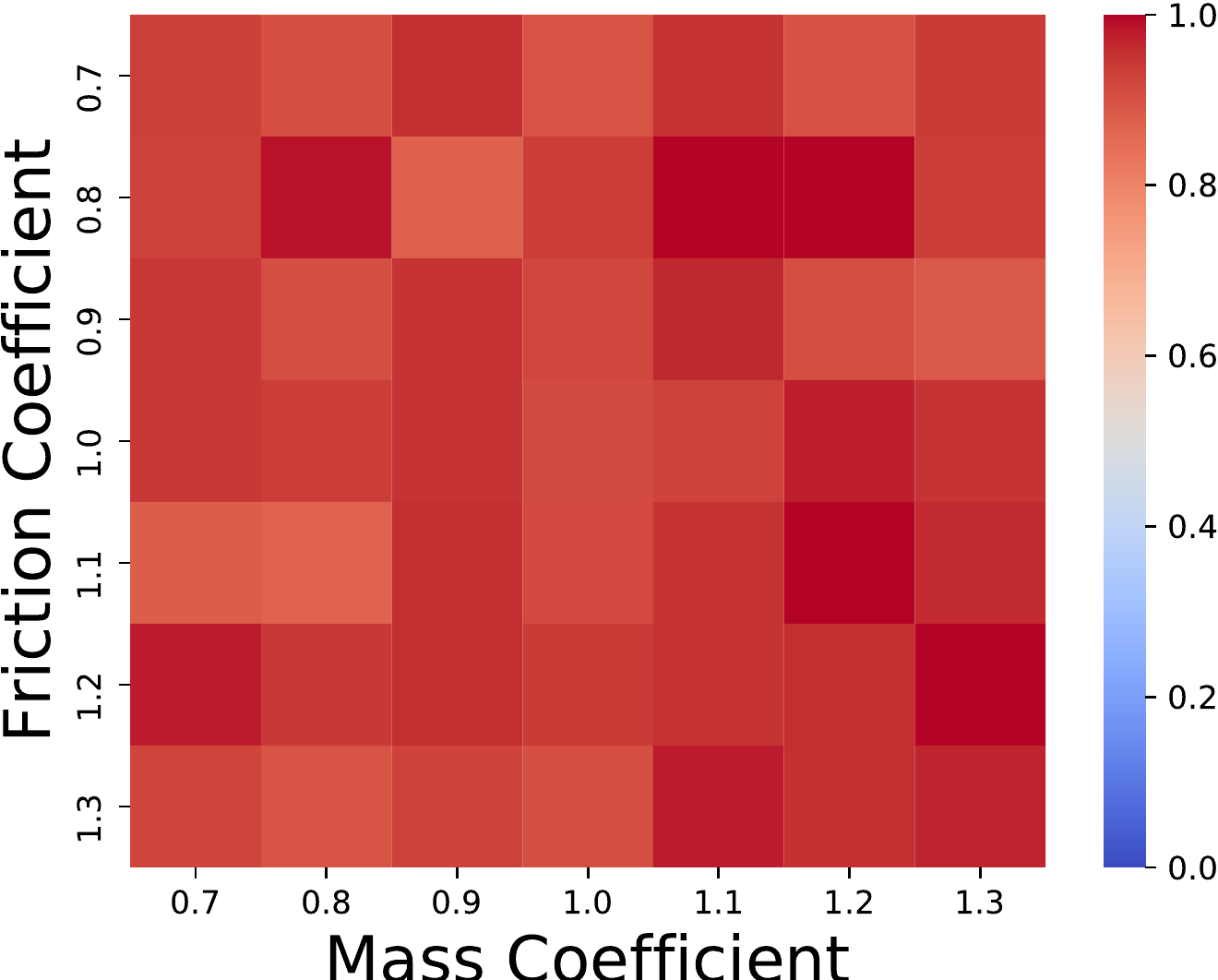}       \label{fig:ppo_walker2d_heatmap_percent_worst}
        }

         \caption {Average normalized return across 10 seeds tested via different mass coefficients for \textbf{PPO} on the x-axis and friction coefficients on the y-axis. High reward has red color; low reward has blue color. $1^{st}$ row: Hopper, $2^{nd}$ row: Half-Cheetah, $3^{rd}$: Walker2d} 
        \label{fig:heatmap_ppo}
    \end{figure*}

\begin{table*}[th!]
\caption{Performance of ROLAH and baselines under various disturbances for \textbf{PPO}.}
\begin{center}
\vspace{-10pt}
        \label{table:task_ppo}
    
        \resizebox{14cm}{!}{
        \begin{tabular}{l | l l l l }
            \toprule
            \textbf{Method} & \textbf{Baseline (0 adv)} & \textbf{RARL (1 adv)} & \textbf{RAP (population adv)} & \textbf{ROLAH} \\
            \midrule
            Hopper (No disturbance)        & 0.89$\pm$0.009          & \textbf{0.97$\pm$0.003}               & 0.87$\pm$0.003 & \textbf{0.97$\pm$0.33}  \\
            Hopper(Action noise)         & 0.72$\pm$0.07         & \textbf{0.94$\pm$0.002}      & 0.53$\pm$0.2     & 0.88$\pm$0.001      \\
            Hopper (Worst Adversary)       & 0.65$\pm$0.04           & \textbf{0.88$\pm$0.009}          & 0.68$\pm$0.17  & 0.87$\pm$0.24  \\
            \midrule
            Half-Cheetah (No disturbance)        & 0.89$\pm$0.04          & 0.91$\pm$0.04               & 0.89$\pm$0.08 & \textbf{0.92$\pm$0.08}  \\
            Half-Cheetah(Action noise)         & \textbf{0.91$\pm$0.03}         & 0.89$\pm$0.10      & 0.53$\pm$0.43     & \textbf{0.91$\pm$0.03}      \\
            Half-Cheetah (Worst Adversary)       & 0.21$\pm$0.24            & 0.24$\pm$0.04          & 0.28$\pm$0.39  & \textbf{0.51$\pm$0.43}  \\
            \midrule
            Walker2d (No disturbance)        & 0.94$\pm$0.32          & 0.91$\pm$0.33              & 0.90$\pm$0.10 & \textbf{0.98$\pm$0.09}  \\
            Walker2d (Action noise)         & 0.93$\pm$0.29         & 0.86$\pm$0.35      & \textbf{0.99$\pm$0.14}     & 0.98$\pm$0.03      \\
            Walker2d  (Worst Adversary)       & 0.30$\pm$0.13           & 0.51$\pm$0.16          & 0.53$\pm$0.24  & \textbf{0.71$\pm$0.37}  \\
            \bottomrule
        \end{tabular}}
        
   \end{center}
   
    \end{table*}
\section{More Experiments for PPO}
In Section \ref{sec:exp}, we adopt TRPO as our core baseline and consider different adversarial algorithms built on top of TRPO. We conduct the whole experiments using Proximal Policy Optimization (PPO) \cite{ppo} and show the robustness comparison with varying test conditions in Figure~\ref{fig:heatmap_ppo} and with various disturbances in Table \ref{table:task_ppo}. Our ROLAH performs better against other baselines using PPO, indicating that our approach is not limited to a specific RL policy optimization method.

\section{Hyperparameters}
In our experiments, $10$ adversarial candidates are considered in RAP and ROLAH and select the worst-$k$ adversaries for updating in each iteration with $k=3$. We implement our method as well as existing baselines using TRPO and PPO. Our implementation are partly based on the codes published by \emph{rllab}~\cite{rllab}. We list the hyperparameters we choose in Table \ref{table:hyper}. The clipping range for PPO is $0.2$. For those hyperparameters which are not listed, we adopt the default values in \emph{rllab}.

\begin{table}[htbp]
\centering
\caption{The hyperparameter used for experiments.}
\label{table:hyper} 
\begin{tabular}{ p{3.2cm} p{3cm}  }
  \toprule
    Hyperparameters           & Values                 \\ 
  \midrule
    N\underline {o} of layers &  3\\
    Neurons in each layer        & 256, 256, 256                      \\
    Batch Size    & 4000                          \\
    Discount Factor ($\gamma$)                    & 0.995                          \\
    GAE parameter ($\lambda$)           & 0.97                 \\
    N\underline {o} of iterations                                           & 500                       \\ 
  \bottomrule
\end{tabular} 
\end{table}

\end{document}